\pdfoutput=1
\documentclass[10pt,twocolumn,letterpaper]{article}

\usepackage[pagenumbers]{cvpr}

\usepackage[T1]{fontenc}
\usepackage{multirow}
\usepackage[expansion=false]{microtype}
\usepackage{tikz}
\usetikzlibrary{arrows.meta,positioning,calc}
\renewcommand{\paragraph}[1]{\vspace{.3em}\noindent\textbf{#1.}}
\newcommand{\lpips}{LPIPS\xspace}
\newcommand{\ds}{DreamSim\xspace}
\newcommand{\ir}{ImageReward\xspace}
\newcommand{\ps}{PickScore\xspace}
\newcommand{\fd}{FD-DINO\xspace}
\newcommand{\sh}{\sigma_h}
\newcommand{\xhat}{\hat{x}_0}
\newcommand{\logit}{\operatorname{logit}}
\DeclareMathOperator{\up}{up}
\DeclareMathOperator{\avgpool}{avgpool}

\usepackage{amsthm}
\theoremstyle{plain}
\newtheorem{proposition}{Proposition}

\usepackage{algorithm}
\usepackage[noend]{algpseudocode}
\usepackage{placeins}
\newcommand{\method}{DtR\xspace}
\AtEndPreamble{\crefname{appendix}{Supp.}{Supp.}\Crefname{appendix}{Supp.}{Supp.}}

\newcommand{\saveVsMrflow}{23}
\newcommand{\raluRatio}{1.5}
\newcommand{\cacheRatio}{1.2}

\providecommand{\saveVsMrflow}{??}\providecommand{\raluRatio}{??}\providecommand{\cacheRatio}{??}

\definecolor{cvprblue}{rgb}{0.21,0.49,0.74}
\usepackage[pagebackref,breaklinks,colorlinks,allcolors=cvprblue]{hyperref}
\title{Same Seed, Different Picture? \\ Deciding Early at Low Resolution for Seed-Faithful Few-Step Flow Sampling}

\author{Huiran Duan$^{1}$\qquad Weilun Feng$^{2,3}$\qquad Yuqi Li$^{1}$\qquad Yingli Tian$^{1,*}$\\
{\small $^{1}$The City University of New York}\\
{\small $^{2}$Institute of Computing Technology, Chinese Academy of Sciences} \quad {\small $^{3}$University of Chinese Academy of Sciences}
}

\begin{document}
\twocolumn[{%
\renewcommand\twocolumn[1][]{#1}%
\maketitle
\vspace{-2em}
\begin{center}
\captionsetup{type=figure}
\begin{tikzpicture}
\node[inner sep=0pt, anchor=south west] at (0,0) {\includegraphics[width=\textwidth]{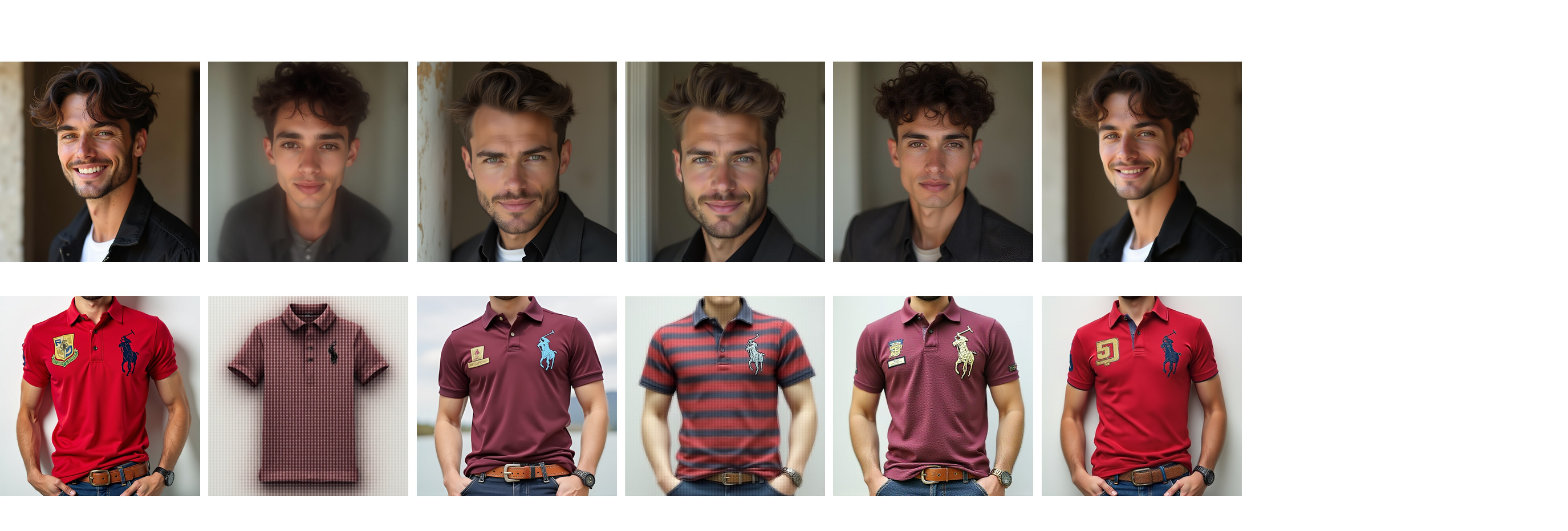}};
\node[inner sep=0pt, anchor=south west] at (0,0) {\includegraphics[width=\textwidth]{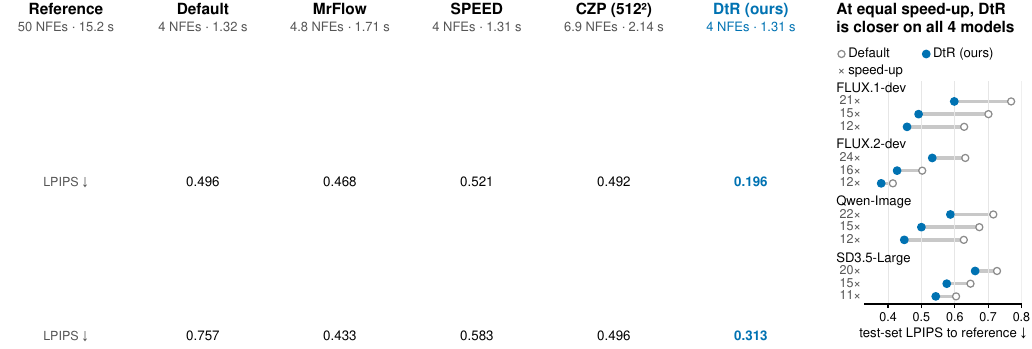}};
\end{tikzpicture}

\vspace{-1.2em}
\captionof{figure}{\textbf{From the same noise, few-step samplers generate a different image from the 50-step sampler, whereas \method stays closest to it.} Left: FLUX.1-dev samples at about 4 NFEs, with SPEED \cite{speed2026} at 4 NFEs and MrFlow \cite{mrflow2026} and CZP \cite{lowrespreview2026} at their cheapest settings above 4 NFEs. An NFE is one full-resolution evaluation, and low-resolution evaluations count by their measured cost. Headers give NFEs and seconds per image on one GPU, and numbers under the images give \lpips \cite{lpips2018} to the reference. Right: \method against Default, each model's official sampler with fewer steps, in every speed-up group of \cref{tab:main}.}
\label{fig:teaser}
\end{center}%
}]
{\renewcommand{\thefootnote}{}\footnotetext{$^{*}$Corresponding author: \texttt{ytian@ccny.cuny.edu}}}
\begin{abstract}
Few-step samplers make large text-to-image flow models fast, but from the same noise they no longer generate the image of the standard 50-step sampler. Subjects, attributes and layout change, not just fine detail. We trace this loss of seed fidelity to the start of sampling. The coarse part of the initial noise decides coarse structure first, while the noise level is still above 0.9. Few-step samplers cross this range in their first steps, which account for 25--52\% of their distance to the 50-step reference. Because this decision is coarse, a half-resolution run from the same coarse noise can make nearly the same decision at a third of the cost. We therefore propose Decide-then-Render (\method), a training-free sampler that decides the composition in a few low-resolution steps, rebuilds the full-resolution state with the seed's own noise and renders the details. On 4 large models, \method runs 11--25$\times$ faster than the reference sampler. At each speed-up, it is closer to the reference and rated higher by \ir than every compared sampler, including few-step schedules calibrated per budget and published multi-resolution samplers. On FLUX.1-dev, it remains the most faithful sampler down to about $8\times$, ahead of caching accelerators. A user study on this model found \method's samples closer to the reference in 86\% of pairwise trials. \method transfers to a held-out model and to video, and distilled models can serve as its renderers. Project link: {\small\url{https://duanhuiran.github.io/DtR}}
\end{abstract}

\vspace{-1em}
\section{Introduction}
\label{sec:intro}
Cutting the number of sampling steps of a large flow model not only blurs its images but also changes what they show. From the same noise, the official sampler of FLUX.1-dev \cite{flux2024} with 4 instead of 50 steps changes a person's identity and the colors and layout of an image (\cref{fig:teaser}). Published multi-resolution samplers, low-resolution previews and few-step schedules also move away from the 50-step image (\cref{tab:main}). This matters whenever a fast sample stands in for the full sampler's image. A preview is useful only if the full sampler would generate nearly the same image \cite{consolver2026,lowrespreview2026}, and a seed is reproducible across budgets only if its image changes little with the budget. We call this property \emph{seed fidelity} and measure it as the closeness of a sample to the \emph{reference}, which is the model's 50-step sample from the same noise.

Two lines of training-free work make sampling cheaper, but neither measures when along the trajectory the image of a seed is decided. Solvers and schedules \cite{dpmsolver2022,unipc2023,ays2024,sharpeuler2026} improve the integrator or the placement of the steps, but every step still runs at full resolution. Multi-resolution samplers \cite{postdiff2025,bottleneck2025,ralu2026,fresco2026,mrflow2026,speed2026} run the early steps at low resolution and must therefore choose when to switch to full resolution, and from which state and with which noise to continue. Existing samplers switch late, re-noise with fresh noise or carry the evolved low-resolution state across (\cref{fig:method}a), and each of these choices loses part of the seed's composition.

\begin{figure*}[t]
\centering
\input{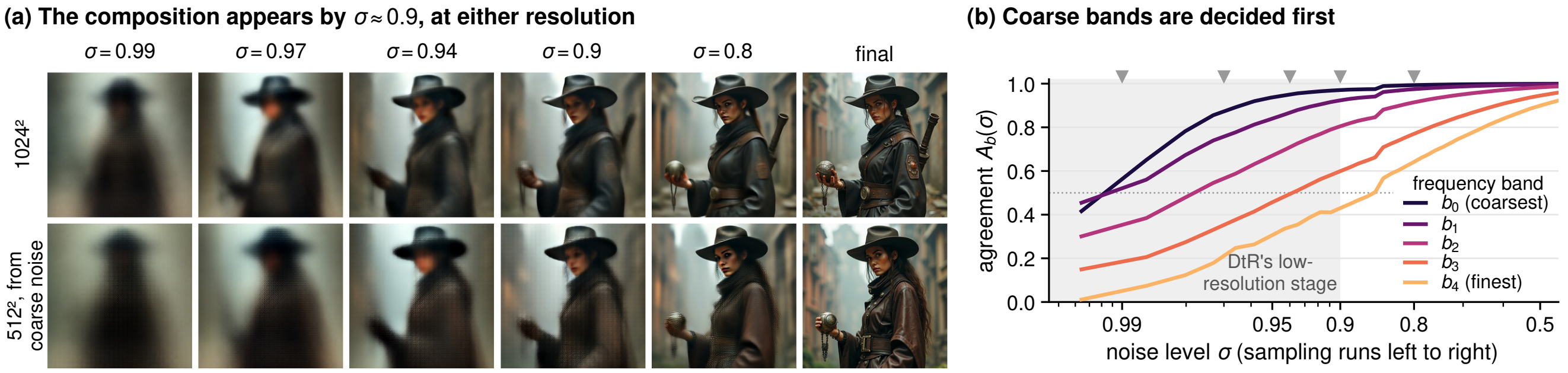}
\vspace{-1em}
\caption{\textbf{Samples are decided early, coarse structure first, and a half-resolution run makes nearly the same decision.} (a)~Clean estimates $\xhat(\sigma)$ along FLUX.1-dev's 50-step trajectory at $1024^2$ and, below, along a half-resolution trajectory started from the coarse part of the same noise. (b)~Band agreement $A_b$ of \cref{eq:agreement} on 36 FLUX.1-dev trajectories, with the $\sigma$ axis in log-odds and a dotted line at 0.5. Triangles mark the columns of (a), and shading marks $\sigma>0.9$, where \method runs at half resolution.}
\label{fig:mechanism}
\par\vspace{-0.5em}
\end{figure*}

We find that the content of a sample is decided early, coarse structure first. Sampling runs from pure noise at $\sigma=1$ to the image at $\sigma=0$, and the coarse frequency bands are decided while $\sigma$ is still above 0.9 (\cref{fig:mechanism}). The coarse part of the noise determines them. Few-step samplers cross this range in 1 or 2 large steps and lose a large part of their fidelity there. When a few-step schedule is given the reference trajectory's state near $\sigma=0.9$, its distance to the reference drops by 25--52\% (\cref{fig:decision}a).

Because this decision concerns only coarse structure, a lower resolution can make it. We derive a condition under which a half-resolution run started from the coarse part of the seed's noise follows the coarse part of the full-resolution flow (\cref{prop:resolution}). Without guidance this condition approximately holds, but classifier-free guidance \cite{cfg2022}, which large text-to-image models use directly or in distilled form, makes part of the decision differ between the resolutions. This difference can be measured in advance on a few trajectories, and it predicts how much a model gains from deciding at low resolution (\cref{fig:decision}c).

These findings tell a multi-resolution sampler when to switch and which noise to use, and lead to Decide-then-Render (\method). \method first runs a few cheap low-resolution steps that decide the composition (\cref{fig:method}b). It switches to full resolution while $\sigma$ is still at least 0.9, when coarse structure is decided but the finest details are not. At the switch it rebuilds the full-resolution state from the upsampled clean estimate and the seed's own noise. The remaining steps render the details, and no training is needed.

On FLUX.1-dev, FLUX.2-dev \cite{flux2dev2025}, Qwen-Image \cite{qwenimage2025} and SD3.5-Large \cite{esser2024sd3}, \method costs the equivalent of 2 to 4 full-resolution network function evaluations (NFEs) and runs 11--25$\times$ faster than the 50-step sampler. At every speed-up, it is closer to the reference and rated higher by \ir \cite{imagereward2023} than every compared sampler, including a few-step schedule calibrated per budget (\cref{tab:main}). On FLUX.1-dev it remains the most faithful sampler down to about $8\times$, ahead of caching accelerators (\cref{fig:cache}). In a user study on this model, participants chose its sample as the one closer to the reference in 86\% of the trials. At 2 and 3 NFEs, its few-step samples also pick better seeds than those of the few-step schedules.

\paragraph{Contributions}
\begin{itemize}[leftmargin=1.2em,itemsep=1pt,topsep=2pt]
\item \emph{When a sample is decided.} We find that coarse structure is decided first, while the noise level is still above 0.9, by the coarse part of the noise, and that few-step samplers lose a large part of their fidelity there (\cref{sec:spectrum,sec:noise,sec:firststep}).
\item \emph{Deciding at low resolution.} We derive when a low-resolution run makes the full-resolution decision, identify guidance as the main violation and show that its effect can be measured in advance (\cref{sec:lowres,sec:guidance}).
\item \emph{Decide-then-Render.} We propose \method, a training-free sampler that decides at low resolution and renders at full resolution with the seed's own noise, and is the most faithful of the compared samplers at 2 to 4 NFEs on 4 large models (\cref{sec:method,sec:experiments}).

\end{itemize}

\section{Related work}
\label{sec:related}
\paragraph{Few-step solvers and schedules}
DPM-Solver \cite{dpmsolver2022,dpmsolverpp2022}, UniPC \cite{unipc2023}, Align-Your-Steps \cite{ays2024}, SharpEuler \cite{sharpeuler2026} and other solvers and schedules \cite{amed2024,gits2024,betasampling2024,optstep2025,stork2025,tors2026,adasde2025,indis2026} improve the integrator or timesteps, and first-step corrections for guided models \cite{cfgzero2025,epcfg2024,apg2024,rescaledcfg2023} damp the oversized first update. All of them run every step at full resolution.

\paragraph{Caching accelerators}
TeaCache \cite{teacache2025}, TaylorSeer \cite{taylorseer2025} and DPCache \cite{dpcache2026} keep the long schedule and forecast transformer features between fully computed steps.

\paragraph{Multi-resolution samplers and previews}
Multi-resolution samplers make early evaluations cheap and build on three ideas (\cref{fig:method}a). First, generation runs coarse to fine, and SPEED \cite{speed2026} times its resolution growth by the spectrum. Second, the state is re-noised when the resolution changes. SDEdit-style hi-res passes \cite{sdedit2022}, PostDiff \cite{postdiff2025} and Bottleneck Sampling \cite{bottleneck2025} use fresh noise, and Fresco \cite{fresco2026} uses one coordinate-hashed noise field for all stages. Third, a lower resolution can agree with the full one when downsampling commutes with the flow, which is the condition behind commutator-zero previews (CZP) \cite{lowrespreview2026}. RALU \cite{ralu2026} upsamples edge regions first, MrFlow \cite{mrflow2026} refines the super-resolved decode of a low-resolution image, and other work repairs the transitions \cite{selflift2026,noiseshift2025} or learns them \cite{teng2024relay,jin2025pyramidal,starodubcev2025swd}.

\paragraph{How \method differs}
\method differs from these samplers in when it switches and which noise it uses. It switches while $\sigma$ is still at least 0.9, and both of its stages use the seed's own noise. The low-resolution stage starts from the noise's block averages, which after rescaling are again white noise \cite{chang2024warped}. SelfLift \cite{selflift2026} switches late in a step-distilled schedule, re-noises with fresh noise and targets image quality, not seed fidelity. ConsistencySolver \cite{consolver2026}, its FLUX.1-dev adaptation \cite{jia2026faithful} and step distillation \cite{add2024,ladd2024,hypersd2024,dmd2_2024} need training, unlike \method.

\paragraph{When diffusion models decide}
Coarse-to-fine generation follows from the forward process and the architecture \cite{ihd2023,spectralautoreg2024,dctdiff2025,reschroma2024}, and the low-frequency part of the initial noise carries the layout of a video \cite{freeinit2024}. Theory places decisions in critical windows \cite{li2024critical} or at a speciation time, when trajectories commit to one class of the data \cite{biroli2024,achilli2026speciation,raya2023}, and at high noise the score is close to a Gaussian one \cite{wang2024gaussian}. Guidance can harm quality and diversity at high noise \cite{guidanceinterval2024}, amplify estimation errors \cite{chidambaram2024guidance}, act during speciation \cite{pavasovic2025cfg} and change its role across sampling stages \cite{jin2025stage,jiang2026analytic,chung2025cfgpp}. Restart sampling \cite{restart2023} re-noises part of a trajectory to correct accumulated error, and models trained on the same data map the same noise to nearly the same image \cite{reproducibility2024}. We measure when the content of each seed is decided on large models.

\section{Where a flow sample is decided}
\label{sec:mechanism}
\paragraph{Setting}
We study the guidance-distilled FLUX.1-dev \cite{flux2024} and FLUX.2-dev \cite{flux2dev2025} and the classifier-free-guided (CFG) Qwen-Image \cite{qwenimage2025} and SD3.5-Large \cite{esser2024sd3}, all at $1024^2$ and at their native guidance. On the rectified-flow path from data $x_0$ to noise $\epsilon$, the model predicts a velocity $v_\theta$, which gives a clean estimate $\xhat$ at every noise level $\sigma$,
\begin{equation}
x_\sigma=(1-\sigma)x_0+\sigma\epsilon,\qquad\xhat(\sigma)=x_\sigma-\sigma v_\theta(x_\sigma,\sigma).
\label{eq:path}
\end{equation}
A higher $\sigma$ means an earlier point in sampling. To see when each spatial scale is decided, we split latents into 5 radial frequency bands with projections $\Pi_b$, from the coarsest band $b_0$ to the finest band $b_4$. The finest band holds the frequencies that a half-resolution latent cannot represent. To separate what the seed decides from what the prompt decides, we subtract the mean over seeds of the same prompt from a quantity $y$ and call $\delta y=y-\mathbb{E}_\epsilon[y]$ its \emph{seed-specific} part. In each band we compare the seed-specific part of the clean estimate with that of the final sample $x_0$,
\begin{equation}
A_b(\sigma)=\cos\big(\Pi_b\,\delta\xhat(\sigma),\;\Pi_b\,\delta x_0\big).
\label{eq:agreement}
\end{equation}
A band is \emph{decided} once its agreement $A_b$ is high and later sampling no longer lowers it. To compare bands, we note where $A_b$ reaches 0.5.

\subsection{Coarse structure is decided first and early}
\label{sec:spectrum}
The coarse bands are decided first, as \cref{fig:mechanism}b shows for FLUX.1-dev and Supp.~\ref{app:bands} for Qwen-Image and SD3.5-Large. The two coarsest bands reach an agreement of 0.5 above $\sigma=0.95$ and the next two above 0.9, whereas the finest band reaches it only below $\sigma=0.87$. The decoded clean estimates show this order (\cref{fig:mechanism}a). The woman's pose and hat are already set at $\sigma\approx0.97$, the orb in her hand appears by $\sigma\approx0.9$, and later steps add detail to a composition that no longer changes.

Why are the coarse bands decided first? They carry $10^2$ to $10^3$ times more seed-specific variance per coefficient than the finest band, and by the scaling argument behind the resolution-dependent time shift \cite{esser2024sd3,simplediffusion2023}, a band with more such variance is decided at a higher $\sigma$. The measured decision points of all bands follow this order across models and resolutions (Supp.~\ref{app:law}, \ref{app:lemma}).

\subsection{The coarse noise carries the decision}
\label{sec:noise}
Which part of the noise decides? We split the initial noise into two parts. The coarse part is its $2\times2$ block averages, and the fine part is the residual. We then redraw one part, keep the other and the prompt, and rerun the 50-step sampler. With a redrawn fine part, the seed-specific part of the clean estimates over the first steps keeps a cosine of 0.86--0.98 with that of the original run on all 4 models, so the composition stays. With a redrawn coarse part, this cosine is within 0.04 of zero, and the composition is lost (Supp.~\ref{app:noisesplit}).

\subsection{Few-step samplers lose fidelity early}
\label{sec:firststep}
\begin{figure*}[t]
\centering
\includegraphics[width=\textwidth]{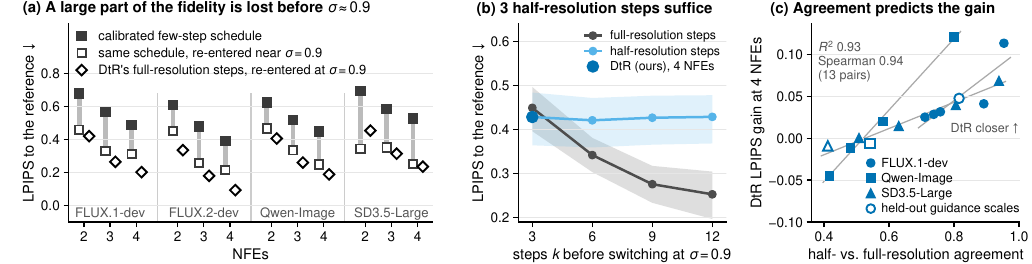}\par
\vspace{-0.3em}
\caption{\textbf{Few-step samplers lose a large part of their fidelity before $\sigma\approx0.9$, and on FLUX.1-dev half-resolution steps make the same decision at a third of the cost.} (a)~\lpips to the reference on at least 30 validation prompts per model. Gray bars mark the fidelity lost before $\sigma\approx0.9$, and diamonds show what \method would reach with the reference trajectory's state at $\sigma=0.9$. Re-entry serves only as a diagnostic. (b)~$k$ half- or full-resolution steps before the switch at $\sigma=0.9$ and the same 3 full-resolution steps after it, on 36 FLUX.1-dev trajectories with 95\% intervals. (c)~For 13 model--guidance pairs, the 4-NFE \lpips gain of \method over the calibrated schedule, or over SE + shrink (\cref{sec:protocol}) on SD3.5-Large, against the agreement of the two resolutions. Lines are per-model fits with a coefficient of determination ($R^2$) of 0.93, and open markers are held-out guidance scales.}
\label{fig:decision}
\par\vspace{-0.25em}
\end{figure*}

A few-step schedule makes this decision with 1 or 2 large full-resolution steps from pure noise. We measure how much fidelity it loses there with a re-entry diagnostic on a strong baseline, the calibrated schedule. For each budget, this schedule tunes the time shift of the default schedule and a first-step shrink, a constant factor that damps its first clean estimate (\cref{sec:protocol}). At the schedule's own noise level nearest $\sigma=0.9$, the diagnostic replaces the schedule's state with the reference trajectory's state at the same level, and the schedule finishes with its remaining steps. Its distance to the reference then drops by 25--52\% on all 4 models and budgets (open squares in \cref{fig:decision}a, Supp.~\ref{app:lead}). The rest of the distance comes from its few steps below $\sigma=0.9$. The loss is final, as a few-step sample returns to the reference only if it is re-noised with its original noise to above $\sigma=0.99$ and continued with many steps (Supp.~\ref{app:renoise}).

\begin{figure*}[t]
\centering
\input{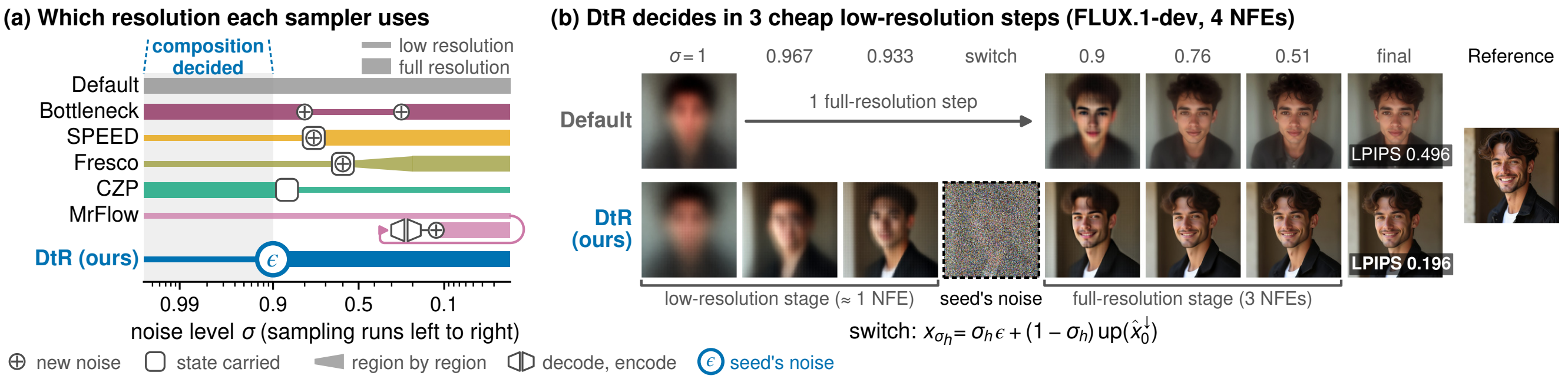}
\vspace{-1em}
\caption{\textbf{\method decides the composition at low resolution and renders it at full resolution from the seed's noise.} (a)~Which resolution each sampler uses along the noise level, based on the descriptions in \cite{bottleneck2025,speed2026,fresco2026,lowrespreview2026,mrflow2026}. The bars of Bottleneck Sampling show only the order of its stages, and a circle without a box means that the state is rebuilt from a clean estimate. MrFlow first finishes at low resolution and then restarts at full resolution from $\sigma=0.12$ (arrow). (b)~Clean estimates after every evaluation for the first example of \cref{fig:teaser}. The default 4-step schedule crosses $\sigma>0.9$ in 1 full-resolution step and generates a different person. \method takes 3 half-resolution steps there, rebuilds the state at the switch point $\sh=0.9$ with the seed's noise (dashed frame) and renders it with 3 full-resolution steps.}
\label{fig:method}
\par\vspace{-0.5em}
\end{figure*}

\subsection{Low resolution can make the same decision}
\label{sec:lowres}
The decision above $\sigma=0.9$ concerns only coarse structure, which a lower resolution can represent. Let $s$ be the ratio of the low to the full side length, so that ${s=1/2}$ takes $1024^2$ to $512^2$. Let $D$ map a latent to its coarse part at the low resolution, normalized so that $D\epsilon$ is again white noise. At ${s=1/2}$, $D=2\avgpool_2$ gives $2\times2$ block averages at unit variance. We write $\xhat^{\downarrow}$ for the clean estimate of the low-resolution run started from $D\epsilon$, the seed's coarse noise.
\begin{proposition}[one decision at two resolutions]
\label{prop:resolution}
Suppose (i) the low-resolution model's samples are distributed as $sDx_0$, the downsampled full-resolution samples, and (ii) the posterior mean of the data's coarse part depends on a noisy state only through the state's coarse part, $\mathbb{E}[Dx_0\mid x_\sigma]=\mathbb{E}[Dx_0\mid Dx_\sigma]$. Then, at every noise level $\sigma$,
\begin{equation}
\xhat^{\downarrow}(\sigma)=s\,D\,\xhat(\tilde\sigma),\qquad\logit\tilde\sigma=\logit\sigma-\log s.
\label{eq:prop}
\end{equation}
\end{proposition}
The low-resolution run thus computes the coarse part of the full-resolution clean estimate at the noisier level $\tilde\sigma$. Supp.~\ref{app:proof} gives the proof, and the condition is closely related to the commutator-zero condition of CZP \cite{lowrespreview2026}. Assumption (ii) holds only approximately, but best at high noise, where the state's fine part is almost all noise and where the low-resolution stage runs. \Cref{prop:resolution} is an explanatory model rather than a guarantee for real networks, and its value lies in its predictions. The more closely a model follows it, the more \method gains (\cref{fig:decision}c).

Without guidance, FLUX.1-dev (guidance embedding 1) and Qwen-Image nearly satisfy assumption (i). Their half-resolution seed-specific variance per coefficient is 0.26--0.27 of the full-resolution one, close to the predicted ${s^2=1/4}$, and the coarse compositions of both resolutions correlate at 0.88--0.97 (Supp.~\ref{app:crossres}).

On FLUX.1-dev, 3 half-resolution steps before the switch are as faithful as 3 full-resolution steps at a third of their cost (\cref{fig:decision}b). More half-resolution steps do not help, whereas more full-resolution steps do, because with guidance the half-resolution model decides part of the mid-frequency content differently.

\subsection{What guidance changes}
\label{sec:guidance}
Run alone, the conditional and unconditional branches each decide the same composition at both resolutions, but guidance amplifies their difference, explicitly as $v_g=v_c+(g-1)(v_c-v_u)$ at scale $g$ or implicitly through FLUX.1-dev's guidance embedding. At half resolution this difference is shorter than at full resolution and points in another direction. Guidance multiplies this mismatch by $g-1$, so the guided updates of the two resolutions diverge far more than either branch does (Supp.~\ref{app:guidance}, \ref{app:cfg}).

The size of this deviation can be measured in advance. We take the cosine between half- and full-resolution clean estimates at matching noise levels above 0.9, on a few trajectories, as the agreement of the two resolutions. Across 13 model--guidance pairs, this agreement and the 4-NFE \lpips gain of \method over a 4-step full-resolution schedule have a Spearman correlation of 0.94 (\cref{fig:decision}c). Two simpler remedies do not work. Withholding guidance early \cite{guidanceinterval2024} changes the target image itself, and rescaling guidance at low resolution does not improve fidelity (Supp.~\ref{app:inside}).

\section{Decide-then-Render}
\label{sec:method}
 \method decides the composition at low resolution and renders it at full resolution (\cref{fig:method}, and \cref{alg:method} in Supp.~\ref{app:sampler}). Starting from the coarse noise $D\epsilon$, it evaluates the model at $k$ low-resolution noise levels $\sigma_j=1-j(1-\sh)/k$, $j=0,\dots,k-1$, with an Euler step between consecutive levels. At the switch point $\sh$ it rebuilds the full-resolution state from the upsampled clean estimate of its last evaluation and the seed's own noise $\epsilon$,
\begin{equation}
x_{\sh}=\sh\,\epsilon+(1-\sh)\,\up\big(\xhat^{\downarrow}(\sigma_{k-1})\big),
\label{eq:switch}
\end{equation}
where $\up$ is bicubic latent upsampling. The switch needs no evaluation of its own, and the remaining full-resolution steps render the image. Three design choices, each backed by a measurement above, define \method.
\begin{itemize}[leftmargin=1.1em,itemsep=1pt,topsep=2pt]
\item \emph{When to switch.} The low-resolution stage runs to $\sh=0.9$ or $0.95$, when the bands $b_0$ to $b_3$, which the low resolution represents, are decided and the finest band $b_4$ is not (\cref{sec:spectrum}). SPEED switches at $\sigma\approx0.76$ on FLUX.1-dev.
\item \emph{Which noise.} Both stages use the seed's noise, the low-resolution stage its coarse part and the switch all of it, because the coarse noise carries the decision (\cref{sec:noise}). At $\sh$ the upsampled estimate carries only the decided coarse bands, and the full-resolution steps render the fine band from the seed's noise. Fresh noise at the switch, as in PostDiff and Bottleneck Sampling, degrades fidelity about as much as redrawing only the coarse part (\cref{tab:ablation}).
\item \emph{How many low-resolution steps.} The stage needs only 3 half-resolution steps, since more do not help (\cref{sec:lowres}).
\end{itemize}

\paragraph{Model-specific settings}
FLUX.2-dev uses FLUX.1-dev's configuration unchanged, and 4 settings depend on the model (\cref{tab:config} in Supp.~\ref{app:sampler}). SD3.5-Large runs the stage at ${s=3/4}$, Qwen-Image shifts its noise levels as in \cref{eq:prop}, and the CFG models Qwen-Image and SD3.5-Large damp their first clean estimate by a first-step shrink in the spirit of CFG-Zero* \cite{cfgzero2025}. From 4 NFEs on Qwen-Image and 5 on SD3.5-Large, the switch keeps the seed's fine noise but takes its coarse noise from the stage's own noise estimate, because re-inserting the seed's coarse noise rotates their guidance direction (Supp.~\ref{app:noise}). Diagnostics on a few trajectories set the resolution and the switch noise, and validation prompts set the shrink and the noise-level map.

\paragraph{Switch point}
A short full-resolution stage starts with a large step. With 1 to 3 full-resolution steps, \method therefore mostly switches at the lower $\sh=0.9$, so that the low-resolution stage decides more of the image, and with longer stages at 0.95 (Supp.~\ref{app:config}). For a new model, the procedure of Supp.~\ref{app:switch} sets $\sh$ in under a GPU-hour by balancing, on a few reference trajectories, the error at the switch against that of the full-resolution stage. Lowering $\sh$ reduces the second and raises the first. On the held-out FLUX.2-klein-base \cite{flux2klein2026} it selects 0.95 at every budget, as a grid search does.

\begin{table*}[t]
\centering\scriptsize
\caption{\textbf{Fidelity and preference by speed-up over the default 50-step sampler.} Test sets, each sample from the same noise as its reference. Seconds (sec) are per $1024^2$ image on one GPU, and $\times$ is the speed-up. \method, in bold, has the best value of every metric in every group, and $^*$ marks \method ahead of all others in its group by a paired 95\% interval excluding zero. CZP is slower than its group. \Cref{sec:protocol} defines the samplers, and Supp.~\ref{app:full} gives the $8\times$ group, \ps, \fd and all intervals.}
\label{tab:main}
\vspace{-0.3em}
\begin{minipage}[t]{0.497\textwidth}\centering{\setlength{\tabcolsep}{2.1pt}\begin{tabular}[t]{@{}c l r r r r r r r@{}}
\toprule
& sampler & NFE & sec & $\times$ & LPIPS$\downarrow$ & DS$\downarrow$ & DINO$\uparrow$ & IR$\uparrow$ \\
\midrule
\multicolumn{9}{@{}l}{\textit{(a) FLUX.1-dev, 1000 prompts}, default 50 steps at 15.2\,sec, IR 0.94} \\
\midrule
$\approx$21$\times$ & default & 2 & 0.71 & 21.4 & 0.769 & 0.635 & 0.203 & -1.90 \\
 & calibrated & 2 & 0.71 & 21.4 & 0.697 & 0.514 & 0.330 & -1.36 \\
 & \textbf{\method} & \textbf{1.95} & \textbf{0.71} & \textbf{21.5} & \textbf{0.598}$^*$ & \textbf{0.325}$^*$ & \textbf{0.572}$^*$ & \textbf{-0.24}$^*$ \\
\midrule
$\approx$15$\times$ & default & 3 & 1.02 & 15.0 & 0.701 & 0.539 & 0.357 & -1.08 \\
 & calibrated & 3 & 1.01 & 15.0 & 0.576 & 0.299 & 0.588 & 0.20 \\
 & SPEED-4 & 2.64 & 0.91 & 16.7 & 0.626 & 0.333 & 0.600 & 0.20 \\
 & MrFlow 4+1 & 2.27 & 0.95 & 16.0 & 0.622 & 0.375 & 0.525 & 0.10 \\
 & MrFlow 6+1 & 2.91 & 1.14 & 13.4 & 0.543 & 0.270 & 0.658 & 0.57 \\
 & \textbf{\method} & \textbf{2.95} & \textbf{1.01} & \textbf{15.0} & \textbf{0.491}$^*$ & \textbf{0.192}$^*$ & \textbf{0.726}$^*$ & \textbf{0.63}$^*$ \\
\midrule
$\approx$12$\times$ & default & 4 & 1.32 & 11.6 & 0.628 & 0.381 & 0.530 & -0.08 \\
 & SE + shrink & 4 & 1.32 & 11.6 & 0.540 & 0.262 & 0.651 & 0.53 \\
 & calibrated & 4 & 1.32 & 11.6 & 0.491 & 0.207 & 0.706 & 0.45 \\
 & SPEED-6 & 3.95 & 1.31 & 11.6 & 0.521 & 0.225 & 0.706 & 0.68 \\
 & CZP ($512^2$) & 6.86 & 2.14 & 7.1 & 0.559 & 0.253 & 0.678 & 0.59 \\
 & \textbf{\method} & \textbf{3.95} & \textbf{1.31} & \textbf{11.6} & \textbf{0.457}$^*$ & \textbf{0.166}$^*$ & \textbf{0.764}$^*$ & \textbf{0.82}$^*$ \\
\midrule
\multicolumn{9}{@{}l}{\textit{(b) FLUX.2-dev, 500 prompts}, default 50 steps at 51.0\,sec, IR 1.19} \\
\midrule
$\approx$24$\times$ & default & 2 & 2.15 & 23.8 & 0.631 & 0.406 & 0.507 & -0.60 \\
 & calibrated & 2 & 2.15 & 23.8 & 0.622 & 0.392 & 0.514 & -0.68 \\
 & \textbf{\method} & \textbf{1.91} & \textbf{2.07} & \textbf{24.6} & \textbf{0.532}$^*$ & \textbf{0.255}$^*$ & \textbf{0.676}$^*$ & \textbf{0.36}$^*$ \\
\midrule
$\approx$16$\times$ & default & 3 & 3.16 & 16.1 & 0.502 & 0.243 & 0.706 & 0.40 \\
 & calibrated & 3 & 3.16 & 16.1 & 0.481 & 0.234 & 0.718 & 0.43 \\
 & \textbf{\method} & \textbf{2.91} & \textbf{3.09} & \textbf{16.5} & \textbf{0.427}$^*$ & \textbf{0.172}$^*$ & \textbf{0.780}$^*$ & \textbf{0.91}$^*$ \\
\midrule
$\approx$12$\times$ & default & 4 & 4.18 & 12.2 & 0.414 & 0.172 & 0.789 & 0.77 \\
 & SE + shrink & 4 & 4.18 & 12.2 & 0.382 & 0.161 & 0.795 & 0.77 \\
 & calibrated & 4 & 4.18 & 12.2 & 0.399 & 0.161 & 0.799 & 0.79 \\
 & \textbf{\method} & \textbf{3.91} & \textbf{4.11} & \textbf{12.4} & \textbf{0.379} & \textbf{0.144}$^*$ & \textbf{0.821}$^*$ & \textbf{1.06}$^*$ \\
\bottomrule
\end{tabular}
}\end{minipage}\hfill
\begin{minipage}[t]{0.497\textwidth}\centering{\setlength{\tabcolsep}{2.1pt}\begin{tabular}[t]{@{}c l r r r r r r r@{}}
\toprule
& sampler & NFE & sec & $\times$ & LPIPS$\downarrow$ & DS$\downarrow$ & DINO$\uparrow$ & IR$\uparrow$ \\
\midrule
\multicolumn{9}{@{}l}{\textit{(c) Qwen-Image, 500 prompts}, default 50 steps at 29.9\,sec, IR 1.22} \\
\midrule
$\approx$22$\times$ & default & 2 & 1.31 & 22.8 & 0.716 & 0.597 & 0.227 & -1.64 \\
 & calibrated & 2 & 1.31 & 22.8 & 0.651 & 0.523 & 0.237 & -1.34 \\
 & \textbf{\method} & \textbf{2.01} & \textbf{1.35} & \textbf{22.2} & \textbf{0.587}$^*$ & \textbf{0.328}$^*$ & \textbf{0.536}$^*$ & \textbf{0.02}$^*$ \\
\midrule
$\approx$15$\times$ & default & 3 & 1.91 & 15.7 & 0.674 & 0.483 & 0.427 & -0.90 \\
 & calibrated & 3 & 1.91 & 15.7 & 0.543 & 0.285 & 0.603 & 0.17 \\
 & MrFlow 4+1 & 2.35 & 1.66 & 18.0 & 0.643 & 0.449 & 0.467 & -0.56 \\
 & MrFlow 6+1 & 3.02 & 2.06 & 14.5 & 0.581 & 0.319 & 0.640 & 0.35 \\
 & \textbf{\method} & \textbf{3.01} & \textbf{1.94} & \textbf{15.4} & \textbf{0.500}$^*$ & \textbf{0.195}$^*$ & \textbf{0.754}$^*$ & \textbf{0.81}$^*$ \\
\midrule
$\approx$12$\times$ & default & 4 & 2.50 & 12.0 & 0.627 & 0.399 & 0.552 & -0.28 \\
 & SE + shrink & 4 & 2.50 & 12.0 & 0.529 & 0.275 & 0.679 & 0.43 \\
 & calibrated & 4 & 2.50 & 12.0 & 0.456 & 0.188 & 0.743 & 0.73 \\
 & \textbf{\method} & \textbf{4.01} & \textbf{2.54} & \textbf{11.8} & \textbf{0.449} & \textbf{0.159}$^*$ & \textbf{0.803}$^*$ & \textbf{0.98}$^*$ \\
\midrule
\multicolumn{9}{@{}l}{\textit{(d) SD3.5-Large, 500 prompts}, default 50 steps at 14.3\,sec, IR 1.05} \\
\midrule
$\approx$20$\times$ & default & 2 & 0.67 & 21.3 & 0.727 & 0.516 & 0.331 & -1.30 \\
 & calibrated & 2 & 0.67 & 21.3 & 0.693 & 0.477 & 0.384 & -1.15 \\
 & \textbf{\method} & \textbf{2.08} & \textbf{0.70} & \textbf{20.4} & \textbf{0.661}$^*$ & \textbf{0.371}$^*$ & \textbf{0.490}$^*$ & \textbf{-0.66}$^*$ \\
\midrule
$\approx$15$\times$ & default & 3 & 0.95 & 15.0 & 0.647 & 0.384 & 0.527 & -0.32 \\
 & calibrated & 3 & 0.95 & 15.0 & 0.607 & 0.329 & 0.586 & -0.16 \\
 & \textbf{\method} & \textbf{3.08} & \textbf{0.98} & \textbf{14.5} & \textbf{0.576}$^*$ & \textbf{0.270}$^*$ & \textbf{0.675}$^*$ & \textbf{0.20}$^*$ \\
\midrule
$\approx$11$\times$ & default & 4 & 1.24 & 11.5 & 0.604 & 0.327 & 0.612 & 0.11 \\
 & SE + shrink & 4 & 1.24 & 11.5 & 0.556 & 0.281 & 0.676 & 0.29 \\
 & calibrated & 4 & 1.24 & 11.5 & 0.553 & 0.261 & 0.677 & 0.35 \\
 & CZP ($512^2$) & 6.57 & 1.89 & 7.6 & 0.663 & 0.273 & 0.666 & 0.49 \\
 & \textbf{\method} & \textbf{4.08} & \textbf{1.27} & \textbf{11.3} & \textbf{0.543}$^*$ & \textbf{0.238}$^*$ & \textbf{0.716}$^*$ & \textbf{0.51} \\
\bottomrule
\end{tabular}
}\end{minipage}\par
\vspace{-0.5em}
\end{table*}
\paragraph{Cost}
One NFE is one full-resolution step, and with classifier-free guidance the two branches of a step count as one NFE. A low-resolution evaluation counts as its measured fraction of a full-resolution one, about a third at half resolution. The cost of a configuration is thus the number of its full-resolution steps plus its low-resolution evaluations weighted by this fraction. A 4-NFE budget on FLUX.1-dev consists of 3 half-resolution steps, which cost about 1 NFE, and 3 full-resolution steps at $\sigma=0.9$, 0.76 and 0.51.

\vfill
\section{Experiments}
\label{sec:experiments}
\vspace{-0.5em}
\subsection{Setup}
\label{sec:protocol}
\paragraph{Models and prompts}
We evaluate the 4 models at $1024^2$ on prompts of MJHQ-30K \cite{mjhq2024}, with development prompts for design, validation prompts for configurations and baseline coefficients, and test prompts for the final evaluation, one seed per prompt. FLUX.1-dev uses 1000 test prompts, and the other 3 models use the first 500 of them.

\paragraph{Reference and metrics}
The default sampler with 50 steps provides the reference of every fidelity metric and the unit $1\times$ of every speed-up. We measure seed fidelity by \lpips \cite{lpips2018} and \ds (DS) \cite{dreamsim2023} to the reference, and semantic similarity (DINO) by the cosine between the DINOv2 class tokens \cite{dinov2_2023} of sample and reference. \ir (IR) \cite{imagereward2023} and \ps \cite{pickscore2023} rate the samples, GenEval \cite{geneval2023} measures text alignment, and \fd \cite{fddino2023} compares distributions. For scale, on the first 200 FLUX.1-dev test prompts, \lpips to the reference is 0.58 for a sample from another seed and 0.20 for a 100-step sample from the same noise. The default 4-step sample has 0.62 and is thus farther from the reference than a sample from another seed, whereas \method at 4 NFEs has 0.45. Every paired difference in \cref{tab:main} keeps its sign and significance with a 100-step reference (Supp.~\ref{app:robust}).

\paragraph{Baselines}
Default is each model's official sampler with fewer steps. The calibrated schedule keeps the form of the default schedule and sets its time shift and first-step shrink per budget on validation prompts. SharpEuler \cite{sharpeuler2026} places steps by the model's measured acceleration, and we add a first-step shrink to it (SE + shrink, Supp.~\ref{app:schedules}). Its grids equal the default ones at 2 and 3 NFEs, so it appears only at 4 NFEs. We also run the official code of MrFlow \cite{mrflow2026}, SPEED \cite{speed2026}, RALU \cite{ralu2026}, DPCache \cite{dpcache2026} and TaylorSeer \cite{taylorseer2025}. SPEED-$n$ takes $n$ steps, and MrFlow $K_L$+1 takes $K_L$ low-resolution steps and a full-resolution refinement. We start MrFlow and SPEED from noise derived from the reference's seed, which brings their samples closer to the reference (Supp.~\ref{app:multires}). CZP \cite{lowrespreview2026} has no released code, so we reimplement it with 8 steps, the last 5 at $512^2$. Its switch adds 1 evaluation at $1024^2$ and 4 at $512^2$ to choose its subsampling pattern, and we compare its previews with the downsampled reference (Supp.~\ref{app:czp}). Seconds are per image on one RTX PRO 6000 GPU, from noise draw to decoded image, with one protocol for all methods (Supp.~\ref{app:timing}).

\subsection{Main results}
\label{sec:frontier}
\begin{figure*}[t]
\centering
\input{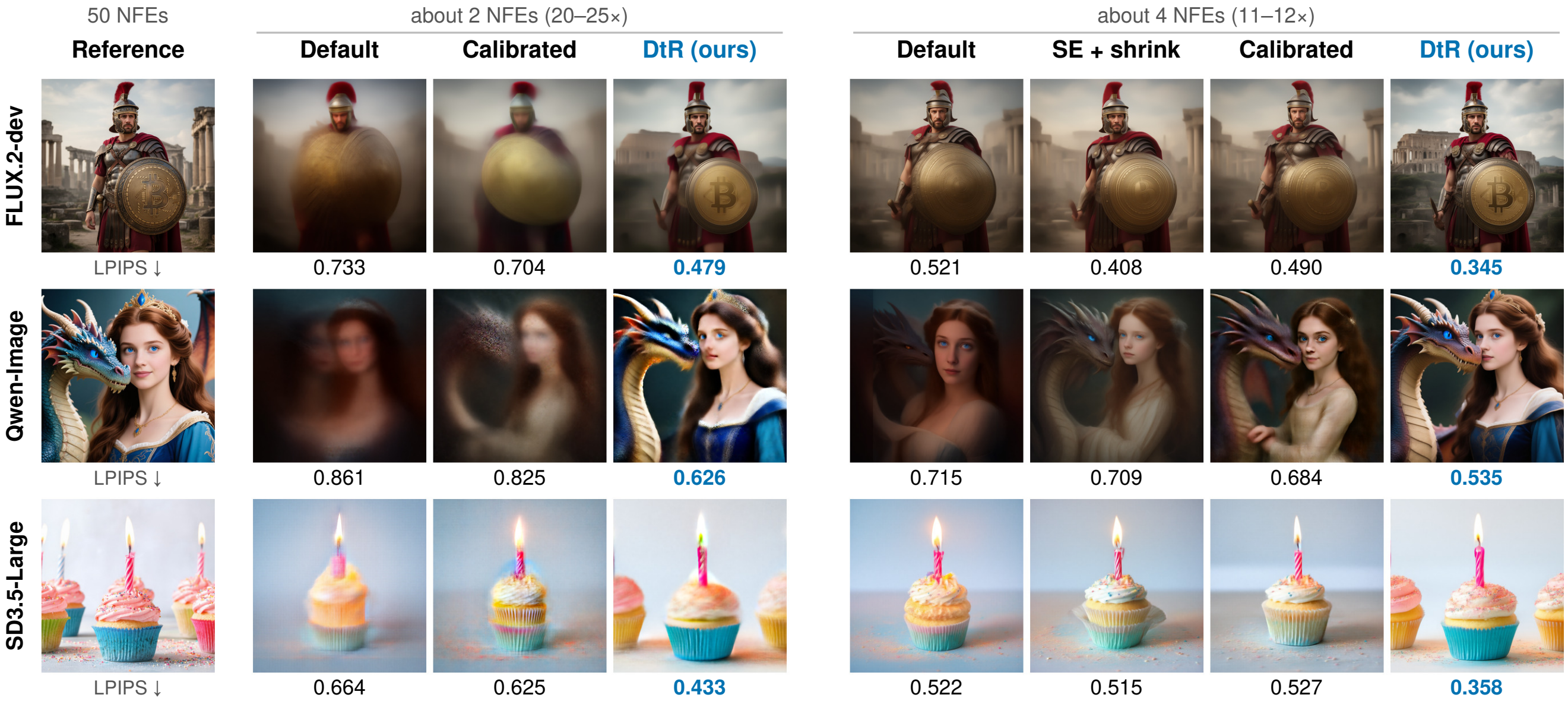}
\vspace{-1em}
\caption{\textbf{On FLUX.2-dev, Qwen-Image and SD3.5-Large, \method stays closest to the reference at both budgets.} Test prompts, each sample from the same noise as its 50-step reference, with \lpips under each image and the best value at each budget in bold.}
\label{fig:models_main}
\par\vspace{-0.5em}
\end{figure*}
\paragraph{Fidelity}
In every speed-up group of \cref{tab:main} from $11\times$ to $25\times$, on all 4 models, \method is the closest to the reference in \lpips and DINO and the best rated by \ir. The gap is largest at the highest speed-ups, where few-step schedules produce blurred drafts while \method already shows the subject and layout of the reference (\cref{fig:models_main}). On FLUX.2-dev it leads clearly at $24\times$ and $16\times$. At $12\times$ its \lpips is not significantly different from that of SE + shrink, while it leads on the other metrics and by 0.06 in GenEval.

\paragraph{Layout without blur}
Downsampled to $128^2$, where blur matters less, the 2-NFE FLUX.1-dev samples of \method are still closer to the reference than a sample from another seed, whereas the calibrated and default ones are farther away. DINOv2 patch tokens and the layout of the objects that a COCO detector finds on the GenEval prompts give the same order (Supp.~\ref{app:layout}).

\paragraph{Against the calibrated schedule}
The calibrated schedule narrows the gap but does not close it. At 2 and 3 NFEs \method leads it on every metric and model, by 0.03--0.10 in \lpips, and is up to 1.35$\times$ faster at equal \lpips (Supp.~\ref{app:full}). At 4 NFEs it leads by 0.01--0.03 in \lpips on 3 models and is not significantly different on Qwen-Image, where DINO, \ir and \ps still favor it. As a larger budget of \method reuses the low-resolution stage of a smaller one, the 3- and 4-NFE samples of a seed differ by 0.07--0.11 \lpips on the 4 models, against 0.20--0.30 with the calibrated schedule (Supp.~\ref{app:consistency}).

\paragraph{Text alignment}
Fidelity does not cost text alignment. The 4-NFE GenEval score of \method is within 0.045 of that of the 50-step sampler on SD3.5-Large and not significantly different on the other 3 models (Supp.~\ref{app:geneval}).

\paragraph{User study}
\phantomsection\label{sec:userstudy}%
On FLUX.1-dev, 10 participants compared the samples of \method with those of the calibrated schedule, SE + shrink, SPEED, MrFlow and CZP in blind two-alternative forced choices \cite{lpips2018,dreamsim2023} between samples from the same noise. With the 50-step reference shown between the two samples, they chose \method as the closer one in 86\% of 1040 trials, with a 95\% interval of 81--91\%. With only the prompt shown, they chose it as the better image in 86\% of 600 trials, with an interval of 80--91\%. Every participant chose \method in most trials of both tasks, and the share per comparison ranges from 76\% to 97\% in the first task and from 68\% to 100\% in the second (Supp.~\ref{app:userstudy}).

\subsection{Multi-resolution and caching samplers}
\label{sec:baselines}
\begin{figure}[t]
\centering
\includegraphics[width=\linewidth]{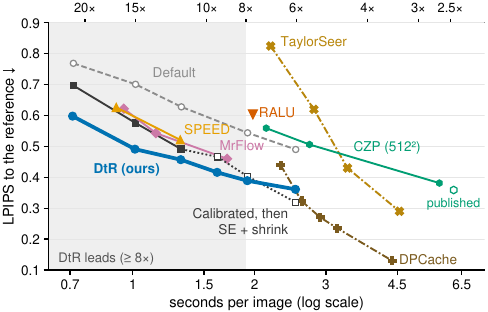}\par
\vspace{-0.3em}
\caption{\textbf{\method is the most faithful sampler on FLUX.1-dev up to about 2 seconds per image, a speed-up of $8\times$.} FLUX.1-dev test set. The full-resolution baseline is the calibrated schedule up to 4 NFEs, drawn solid, and SE + shrink from 5 NFEs, drawn dotted with open squares. CZP's published setting is the open marker, and CZP without its correction (Supp.~\ref{app:czp}) is the line.}
\label{fig:cache}
\par\vspace{-0.5em}
\end{figure}
Around $15\times$ on FLUX.1-dev, where SPEED and MrFlow both operate, \method is closer to the reference than either and rated higher by \ir (\cref{tab:main}). At $12\times$ it is ahead of SPEED on every metric at the same cost, mostly through its higher switch point and step placement and partly through its rebuilt state (\cref{tab:ablation}, Supp.~\ref{app:multires}). \method reaches the fidelity of MrFlow 12+1 in \saveVsMrflow\% less time on FLUX.1-dev and in 22\% less on Qwen-Image. RALU is less faithful even though it takes \raluRatio$\times$ as long (\cref{fig:cache}). CZP runs the two resolutions in the opposite order, full resolution first. Its $512^2$ previews take twice as long as \method and are less faithful than \method at 3 and 4 NFEs (\cref{fig:teaser}, Supp.~\ref{app:previews}).

Caching accelerators keep the 50-step trajectory and forecast features between full evaluations, of which they need at least 6 on FLUX.1-dev, so their speed-up stays below about $7\times$ (\cref{fig:cache}). With 6 full evaluations, DPCache is less faithful than \method and \cacheRatio$\times$ slower. With 8 or more full evaluations, at about $5\times$ or less, DPCache is closer to the 50-step image than any few-step sampler.

\paragraph{Where the lead narrows}
\label{sec:where}
The lead of \method over a full-resolution schedule is the difference of two fidelity costs (Supp.~\ref{app:lead}). The schedule pays a decision cost by crossing $\sigma>0.9$ in 1 or 2 large steps, and \method pays a switch cost when it moves to full resolution. Beyond 4 NFEs the schedule's first step and its decision cost shrink, whereas the switch cost does not, so the lead is largest at 2 to 4 NFEs. It lasts to 6 NFEs on FLUX.1-dev and closes at 5 to 6 NFEs on the CFG models, whose switch costs more, and on FLUX.2-dev the two costs nearly meet already at 4 NFEs.

\subsection{Other models and applications}
\label{sec:transfer}
\paragraph{Held-out model}
FLUX.2-klein-base \cite{flux2klein2026}, a 4B CFG model, played no part in designing the sampler. With the stage at ${s=3/4}$ that \cref{prop:resolution} predicted for it and the switch point from Supp.~\ref{app:switch}, \method is 0.08--0.09 \lpips closer to the reference than the default sampler at 2 and 3 NFEs and 0.03 closer than SharpEuler at 4 NFEs, and it scores higher on \ir and GenEval at every budget (Supp.~\ref{app:heldout}).

\paragraph{Distilled renderers}
The distilled FLUX.1-schnell, Hyper-FLUX \cite{hypersd2024} and FLUX.2-klein-4B can render what the low-resolution stage of \method decides. Rendered by schnell, the 2-NFE samples reach an \lpips of 0.465 in 0.68\,s, against 0.589 in 0.64\,s for schnell's own 2 steps (Supp.~\ref{app:distill}).

\paragraph{Video}
On Wan2.1-T2V-1.3B \cite{wan2025}, \method with a stage at ${s=3/4}$ is ahead of the default UniPC sampler \cite{unipc2023} in \lpips, \ds, \ir and \ps, and at equal \lpips UniPC needs about twice the time. \method is also more faithful than SPEED's video schedule at matched seconds. On VBench \cite{vbench2024} it scores higher on imaging and aesthetic quality and stays within 0.022 on the consistency dimensions, which favor static videos (Supp.~\ref{app:video}).

\paragraph{Choosing seeds from previews}
\phantomsection\label{sec:select}%
At 2 and 3 NFEs, the few-step samples of \method pick better seeds than those of the calibrated schedule and the default sampler. With its 3-NFE samples, picking among 4 FLUX.1-dev seeds by \ir recovers 32\% of the gain that picking by the 50-step images achieves over a random pick, against 20\% for the calibrated schedule and 2\% for the default sampler. The slower previews of SPEED-6, MrFlow, RALU, CZP and Hyper-FLUX recover less (Supp.~\ref{app:select}).

\begin{table}[t]
\centering\footnotesize
\caption{\textbf{Ablations.} $\Delta$\lpips is positive when the variant is less faithful, and $^*$ marks a 95\% interval excluding zero. The 0.95 row's long stage has 5 full-resolution steps, and dashes mark variants not run on Qwen (Supp.~\ref{app:protocols}).}
\label{tab:ablation}
\vspace{-0.3em}
{\setlength{\tabcolsep}{2.4pt}\begin{tabular}{@{}l r r@{}}
\toprule
variant & FLUX.1 & Qwen \\
\midrule
\multicolumn{3}{@{}l}{\textit{noise at the switch}} \\
fresh noise & $+0.084^*$ & $+0.060^*$ \\
redraw coarse part, keep fine & $+0.081^*$ & $+0.048^*$ \\
redraw fine part, keep coarse & $+0.017^*$ & $+0.012^*$ \\
keep the evolved state, as in SPEED & $+0.019^*$ & -- \\
\midrule
\multicolumn{3}{@{}l}{\textit{switch point and grid}} \\
switch at 0.95 (vs.\ 0.9), long stage & $-0.049^*$ & $-0.034^*$ \\
late switch and step placement, $\sigma\approx0.76$ & $+0.051^*$ & -- \\
\bottomrule
\end{tabular}
}\par
\vspace{-0.5em}
\end{table}

\subsection{Ablations}
\label{sec:ablation}
\Cref{tab:ablation} changes one part of \method at a time. The noise at the switch matters most. Fresh noise or a redrawn coarse part costs several times more fidelity than a redrawn fine part, and keeping the evolved low-resolution state, as SPEED does, is also worse. Switching at $\sh=0.95$ pays off once the full-resolution stage has 5 or more steps, and the late switch of SPEED costs fidelity. In the low-resolution stage, more steps, a second-order integrator and a lower resolution at equal cost do not help (Supp.~\ref{app:inside}).

\vspace{-0.39em}
\section{Discussion and conclusion}
\label{sec:conclusion}
\paragraph{Discussion}
While the premise of \method, that a lower resolution makes the same decision as the full resolution, holds closely without guidance, it holds least for lettering and small details like a reflection (\cref{fig:supp_more}). Though guidance makes part of the decision depend on the resolution, its effect is measurable in advance. In practice, a few reference trajectories of a new model show its gain and set its switch point (Supp.~\ref{app:newmodel}). A natural direction is to train models whose guided decisions agree across resolutions, and to extend the decide-then-render split to editing and long videos.

\paragraph{Conclusion}
A flow sample is decided early, coarse structure first, by the coarse part of the initial noise, and a lower-resolution run from the same noise can make nearly the same decision. Deciding at low resolution and rendering at full resolution with the seed's noise yields training-free few-step samples that stay closer to the 50-step reference than those of the compared samplers at 2 to 4 NFEs.

\par
{
    \small
    \bibliographystyle{ieeenat_fullname}
    \bibliography{main}
}

\clearpage
\appendix
\setlength{\dbltextfloatsep}{20pt plus 2pt minus 4pt}
\widowpenalty=150
\crefalias{section}{appendix}
\crefalias{subsection}{subappendix}
\counterwithin{figure}{section}\counterwithin{table}{section}
\renewcommand{\thefigure}{\thesection\arabic{figure}}\renewcommand{\thetable}{\thesection\arabic{table}}
\crefname{subappendix}{Supp.}{Supp.}\Crefname{subappendix}{Supp.}{Supp.}
\newtheorem{lemma}{Lemma}
\twocolumn[{\centering{\Large\textbf{\thetitle}\\\vspace{0.5em}Supplementary Material\par}\vspace{1.0em}
\begin{minipage}{\textwidth}\small
\noindent The supplement follows the order of the main text. \Cref{app:measure,app:theory} support the analysis of \cref{sec:mechanism}, \cref{app:sampler} details the sampler of \cref{sec:method}, and \cref{app:setup,app:results,app:budgets,app:transfer,app:ablation} support the experiments of \cref{sec:experiments}. The main results are in \cref{tab:mainfull}, with the paired intervals in \cref{tab:diff_a,tab:diff_b}. \Cref{app:userstudy} gives the user study, and \cref{app:heldout} gives the held-out model. The table below shows where each part of the main text is supported.
\par\vspace{0.3em}
{\centering\footnotesize\setlength{\tabcolsep}{6pt}
\begin{tabular}{@{}l l l@{}}
\toprule
main text & what the supplement adds & Supp. \\
\midrule
\cref{sec:spectrum} & Bands are decided coarse to fine, at noise levels set by their seed-specific variance & \ref{app:bands}, \ref{app:law}, \ref{app:lemma} \\
\cref{sec:noise} & The coarse part of the noise carries the decision & \ref{app:noisesplit} \\
\cref{sec:firststep} & Few-step samplers lose fidelity above $\sigma\approx0.9$, and the loss is final & \ref{app:lead}, \ref{app:renoise} \\
\cref{sec:lowres} & The proof of \cref{prop:resolution}, and how closely the two resolutions agree & \ref{app:proof}, \ref{app:crossres} \\
\cref{sec:guidance} & Guidance delays the decision and separates the resolutions & \ref{app:readout}, \ref{app:guidance}, \ref{app:cfg} \\
\cref{sec:method} & The algorithm, configurations, cost, switch point, switch noise and the procedure for a new model & \ref{app:config} to \ref{app:newmodel}, \ref{app:linear} \\
\cref{sec:protocol} & Prompts, metrics, timing and baselines & \ref{app:prompts} to \ref{app:cachesetup} \\
\cref{sec:frontier} & All metrics and intervals, GenEval, layout, robustness, refinement and the user study & \ref{app:full} to \ref{app:userstudy} \\
\cref{sec:baselines} & Multi-resolution, caching and preview samplers, and why the lead narrows beyond 4 NFEs & \ref{app:multires} to \ref{app:cachesetup}, \ref{app:beyond} to \ref{app:previews} \\
\cref{sec:transfer} & The held-out model, distilled renderers, video and choosing seeds from previews & \ref{app:heldout} to \ref{app:video}, \ref{app:select} \\
\cref{sec:ablation} & Protocols and further ablations & \ref{app:protocols} to \ref{app:shrink}, \ref{app:steps} \\
\cref{sec:conclusion} & Where \method falls behind, and configuring a new model & \ref{app:qual}, \ref{app:newmodel} \\
\bottomrule
\end{tabular}\par}
\vspace{0.1em}
\paragraph{Conventions}
\begin{itemize}[leftmargin=1.1em,itemsep=0pt,topsep=1pt]
\item \textit{Noise level.} Sampling runs from $\sigma=1$ to $\sigma=0$, so a higher $\sigma$ comes earlier, and a switch at a higher $\sh$ is an earlier switch.
\item \textit{Reference.} The reference of a sample is the 50-step sample of the default sampler from the same noise. Closer and more faithful mean lower \lpips or \ds, or higher DINO, to the reference.
\item \textit{Cost.} One NFE is one evaluation at full resolution, the two branches of classifier-free guidance count as one, and a low-resolution evaluation counts as its measured fraction of one (\cref{app:cost}).
\item \textit{Differences.} A difference is \method minus the other sampler, so negative \lpips and \ds and positive DINO, \ir, \ps and GenEval favor \method. Ablations give the variant minus \method, so a positive $\Delta$\lpips means a less faithful variant. The lead or gain of \method is the other sampler's \lpips minus its own.
\item \textit{Intervals.} Intervals are 95\% prompt-cluster bootstraps, which resample prompts with all their samples. In tables, $^*$ marks a paired interval that excludes zero. In the text, a difference stated with a direction, such as closer or ahead, has an interval that excludes zero, and on par means that the interval includes zero.
\item \textit{Names.} DS, IR and PS abbreviate \ds, \ir and \ps. The full-resolution schedules are the default one, the calibrated one and SE + shrink (SharpEuler with the first-step shrink, \cref{app:schedules}). SPEED-$n$ runs $n$ steps, and MrFlow $K_L{+}1$ runs $K_L$ low-resolution steps and 1 full-resolution step (\cref{app:multires}). Among the 4 models, the CFG models are Qwen-Image and SD3.5-Large.
\end{itemize}
\par\noindent\textbf{Notation.}\par\nobreak
{\footnotesize\setlength{\tabcolsep}{4pt}\noindent\begin{tabular}{@{}l l l l@{}}
$s$, $k$ & low-to-full side ratio, low-resolution steps & $\epsilon$, $\hat\epsilon$ & the seed's noise, the low-resolution stage's noise estimate \\
$\sh$, $\sigma_e$ & switch point, last noise level of the low-resolution stage & $A_b$, $V_b$ & agreement of band $b$, its seed-specific variance per coefficient \\
$D$, $P$ & coarse part at low resolution with $D\epsilon$ white, $P=D^\top D$ & $R$, $\chi$ & cross-resolution agreement of trajectories, of final images \\
$\xhat$, $\xhat^{\downarrow}$ & clean estimates at full and at low resolution & $E_{\mathrm{re}}$ & error after re-entry at the reference trajectory's state \\
$\up$ & bicubic latent upsampling & $C_{\mathrm{sw}}$, $C_{\mathrm{dec}}$ & switch cost of \method, decision cost of a few-step schedule \\
\end{tabular}\par}
\end{minipage}\vspace{0.4em}}]

\section{Measurements of the decision}
\label{app:measure}
\noindent This section measures where a sample is decided on FLUX.1-dev, Qwen-Image and SD3.5-Large, with the default 50-step sampler at $1024^2$ and the native guidance, 3.5 on FLUX.1-dev and SD3.5-Large and 4 on the others, unless a subsection states otherwise. The prompt sets are those of \cref{tab:prompts}. \Cref{app:bands,app:noisesplit,app:law} locate the decision, \cref{app:readout,app:crossres,app:guidance} cover guidance and the two resolutions, and \cref{app:renoise} shows that the decision is not undone.

\subsection{Band agreement}
\label{app:bands}
On all 3 models the two coarsest bands reach an agreement of 0.5 above $\sigma=0.95$ and the next two above 0.9, whereas the finest band, which a half-resolution latent cannot represent, reaches it only below $\sigma=0.87$ (\cref{tab:bands}). The agreement $A_b$ of \cref{eq:agreement} is averaged over the development trajectories, 12 prompts, or 8 on Qwen-Image, with 3 seeds each, after the mean over the seeds of each prompt is subtracted. The 5 bands are octaves of the latent's radial frequency, and the finest one lies above the Nyquist frequency of a half-resolution latent. We compare bands by the noise level at which $A_b$ reaches 0.5.
\begin{table}[t]
\centering\footnotesize
\caption{\textbf{Where each band is decided.} The noise level $\sigma$ at which the band agreement $A_b$ reaches 0.5, on the development trajectories. Frequencies are in cycles per image, and 32 is the Nyquist frequency of a half-resolution latent. The low-resolution stage of \method runs above $\sigma=0.9$.}
\label{tab:bands}
\setlength{\tabcolsep}{4pt}
\begin{tabular}{@{}l l c c c@{}}
\toprule
band & frequency & FLUX.1-dev & Qwen-Image & SD3.5-Large \\
\midrule
$b_0$ & below 4 & 0.987 & 0.958 & $>0.99$ \\
$b_1$ & 4 to 8 & 0.987 & 0.979 & 0.979 \\
$b_2$ & 8 to 16 & 0.973 & 0.958 & 0.964 \\
$b_3$ & 16 to 32 & 0.935 & 0.911 & 0.922 \\
$b_4$ & above 32 & 0.860 & 0.820 & 0.850 \\
\bottomrule
\end{tabular}

\end{table}

\subsection{The coarse noise carries the decision}
\label{app:noisesplit}
Replacing the fine part of the initial noise keeps the seed's composition, and replacing the coarse part erases it. We split the noise $\epsilon$ into a coarse part $P\epsilon$ and a fine part $(I-P)\epsilon$, redraw one of them, keep the other and the prompt, and rerun the 50-step sampler. Here $P$ is the $2\times2$ block mean on FLUX.1-dev, FLUX.2-dev and Qwen-Image and a Fourier low-pass at the resolution of SD3.5-Large's $768^2$ low-resolution stage (\cref{app:crossres}). With a redrawn fine part, the seed-specific part of the clean estimates at noise levels from 0.99 to 0.90 keeps a cosine of 0.86--0.98 with that of the original run on all 4 models. With a redrawn coarse part, this cosine is within 0.04 of zero. The decision therefore lies in the part of the noise that the low resolution can represent, and \method starts its low-resolution stage from exactly this part.

\subsection{When a band is decided}
\label{app:law}
\begin{figure*}[t]
\centering
\includegraphics[width=\textwidth]{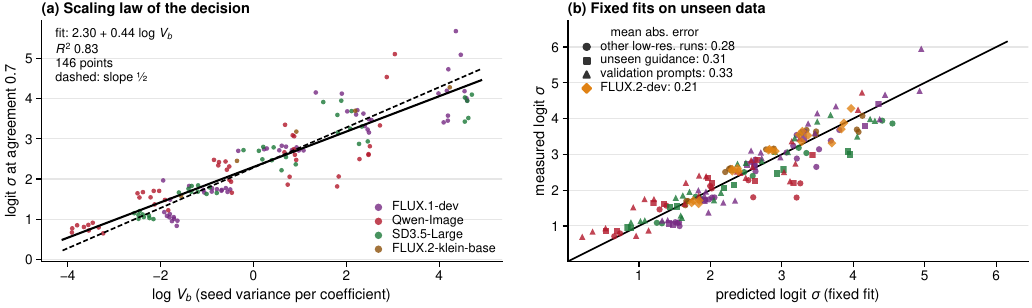}
\caption{\textbf{The scaling law of the decision.} (a)~The $\sigma$ where a band reaches an agreement of 0.7 against its per-coefficient seed-specific variance $V_b$, on FLUX.1-dev, Qwen-Image, SD3.5-Large and FLUX.2-klein-base, over 5 bands, at full and at low resolution and for 14 model--guidance pairs, with the scaling lemma's slope $\tfrac12$ dashed. (b)~Predicted against measured decision points not used in fitting, by \cref{eq:law} for other low-resolution runs and FLUX.2-dev, and by a second fit with a guidance term, $2.79+0.46\log V_b-0.49\log g$, for unseen guidance scales and validation prompts. The legend gives mean absolute errors in log-odds.}
\label{fig:law}
\end{figure*}
The noise level at which a band is decided follows from the seed-specific variance it carries. By the scaling lemma (\cref{lem:scaling} in \cref{app:lemma}), scaling the data by $\sqrt V$ shifts every event defined by a cosine, such as reaching a given band agreement, by exactly $\tfrac12\log V$ in $\logit\sigma$. Applied to each band as its own flow, with $V_b$ the per-coefficient variance of the seed-specific final sample in band $b$, it predicts that the decision points fall on a line of slope $\tfrac12$ in $\logit\sigma$ against $\log V_b$. We measure 146 decision points on FLUX.1-dev, Qwen-Image, SD3.5-Large and FLUX.2-klein-base, over 5 bands, at full and at low resolution and for 14 model--guidance pairs. They follow
\begin{equation}
\logit\sigma^*_{0.7}(b)=2.30+0.44\log V_b
\label{eq:law}
\end{equation}
with a coefficient of determination of 0.83 (\cref{fig:law}a), where $\sigma^*_{0.7}(b)$ is the noise level at which band $b$ reaches an agreement of 0.7, with natural logarithms. The law uses 0.7 rather than the 0.5 of \cref{app:bands}, because the coarsest bands reach 0.5 almost at once. The fitted slope 0.44 is close to the predicted $\tfrac12$. Coarse bands carry $10^2$ to $10^3$ times more seed-specific variance per coefficient than the finest band, so they are decided first. The fitted intercept matches the speciation criterion of Biroli \etal \cite{biroli2024}, a signal-to-noise ratio of about 1 along the few directions that carry the seed-specific content.

The fit of \cref{eq:law} also predicts decision points that it never saw (\cref{fig:law}b). On other low-resolution runs of these models, from $256^2$ to $768^2$, its mean absolute error is 0.28 in $\logit\sigma$, and on FLUX.2-dev it is 0.21. A second fit with a guidance term, $2.79+0.46\log V_b-0.49\log g$, predicts unseen guidance scales and validation prompts with mean absolute errors of 0.31 and 0.33.

\subsection{A linear read-out, and what guidance changes}
\label{app:readout}
\begin{figure*}[t]
\centering
\includegraphics[width=\textwidth]{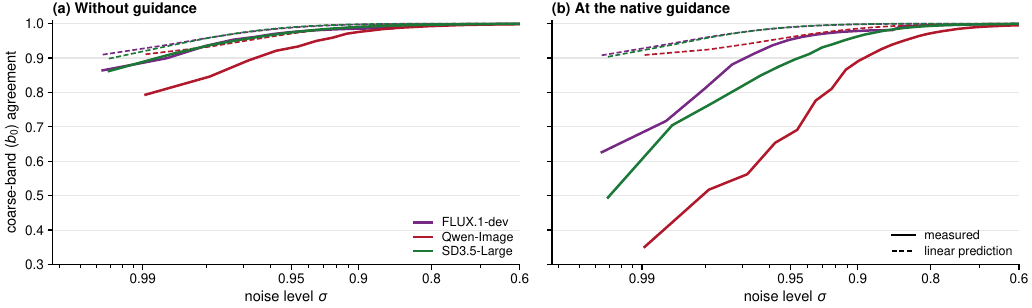}
\caption{\textbf{The coarse band follows a linear read-out of the noise without guidance and lags behind it with guidance.} Agreement of the coarse band $b_0$ along the trajectory, measured as solid lines and predicted by the linear-Gaussian model as dashed lines, for 4 prompts $\times$ 64 seeds per model. The prediction has no fitted parameter and uses only the final samples' seed covariance. (a)~Without guidance. (b)~At the native guidance.}
\label{fig:readout}
\end{figure*}
\begin{table}[t]
\centering\footnotesize
\caption{\textbf{Agreement of the coarse band at the first step.} Per model, 4 prompts with 64 seeds each. The linear prediction has no fitted parameter and uses only the final samples' seed covariance.}
\label{tab:readout}
\setlength{\tabcolsep}{4pt}
\begin{tabular}{@{}l c c c@{}}
\toprule
 & FLUX.1-dev & Qwen-Image & SD3.5-Large \\
\midrule
linear prediction & \multicolumn{3}{c}{0.90--0.91} \\
measured, no guidance & 0.86 & 0.79 & 0.86 \\
measured, native guidance & 0.63 & 0.35 & 0.49 \\
\bottomrule
\end{tabular}

\end{table}
Without guidance the coarse decision is a linear read-out of the noise, and a prediction with no fitted parameter reproduces it. With guidance the models decide later. For Gaussian data the clean estimate is the final sample filtered along each principal direction (\cref{eq:gauss}), so the band agreement follows from the seed covariance of the final samples alone (\cref{app:linear}). We draw 64 seeds for each of 4 prompts per model. The prediction uses only their final samples and never sees the trajectories it is compared with. On the guidance-distilled FLUX.1-dev the guidance scale $g$ is its guidance embedding, and $g=1$ means no guidance.

\paragraph{Without guidance} The coarse band follows the prediction along the whole trajectory (\cref{tab:readout}, \cref{fig:readout}a). With $\Sigma$ the seed covariance, the model's first clean estimate is aligned with its linear response $\Sigma\epsilon$ to the noise at a cosine of 0.73--0.80. A Gaussian fitted to the final samples of 48 other seeds gives a seed's state at $\sigma\approx0.97$ in closed form, and continuing the sampler from this state reproduces the seed's reference composition on FLUX.1-dev, Qwen-Image and SD3.5-Large. The prediction also holds on 4 validation prompts.

\paragraph{At the native guidance} The coarse band's agreement at the first step falls to 0.35--0.63, although the linear prediction from the guided samples' own covariance is 0.90--0.91, as without guidance (\cref{fig:readout}b). The first guided step, $gv_c-(g-1)v_u$, subtracts $g-1$ times the unconditional model's linear response to the same noise, which obscures the composition, and the guided states then decide later. Across 4 guidance scales per model this delay grows with the scale.

\subsection{Agreement between resolutions}
\label{app:crossres}
\begin{table}[t]
\centering\footnotesize
\caption{\textbf{Final images at two resolutions.} Correlation $\chi$ of the two resolutions' coarse final coordinates, and the low resolution's seed variance per coefficient relative to the full resolution's without guidance, which assumption (i) of \cref{prop:resolution} predicts to be $s^2$. Measured on 4 prompts with 64 seeds each.}
\label{tab:chi}
\setlength{\tabcolsep}{3.5pt}
\begin{tabular}{@{}l c c c@{}}
\toprule
 & FLUX.1-dev & Qwen-Image & SD3.5-Large \\
\midrule
low resolution & $512^2$ & $512^2$ & $768^2$ \\
$\chi$, no guidance & 0.97 & 0.88 & 0.98 \\
$\chi$, native guidance & 0.87 & 0.63 & 0.86 \\
seed variance ratio & 0.26 & 0.27 & 0.75 \\
$s^2$ & 0.25 & 0.25 & 0.56 \\
\bottomrule
\end{tabular}

\end{table}
\begin{figure*}[t]
\centering
\includegraphics[width=\textwidth]{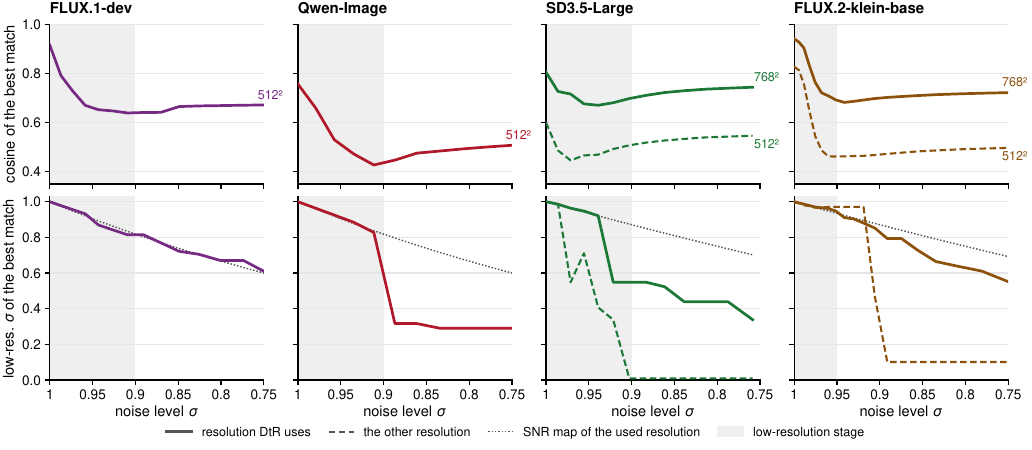}
\caption{\textbf{Where a low-resolution run follows the full-resolution one.} Development trajectories. Top, how well the best match agrees, as the cosine between the full-resolution clean estimate at $\sigma$ and the best-matching low-resolution clean estimate from $D\epsilon$. Bottom, where the best match lies, as its low-resolution noise level, with the SNR map of the used resolution dotted. In the bottom row the solid lines of the resolution that \method uses keep falling longer than the dashed lines of the other resolution, which stall or turn back early. Gray marks the low-resolution stage.}
\label{fig:ridges}
\end{figure*}
\begin{figure}[t]
\centering
\includegraphics[width=\columnwidth]{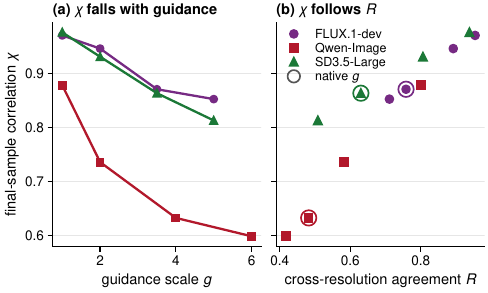}
\caption{\textbf{Final images at two resolutions.} (a)~The correlation $\chi$ of the two resolutions' coarse final coordinates falls with the guidance scale. (b)~$\chi$ against the cross-resolution agreement $R$ over 12 model--guidance pairs, native guidance ringed.}
\label{fig:crossres}
\end{figure}
A low-resolution run follows the full-resolution run at shifted noise levels, and how closely it follows predicts the gain of \method.

\paragraph{Best match} Take a full-resolution trajectory and a low-resolution trajectory started from the seed's coarse noise $D\epsilon$. For each noise level $\sigma$ of the full-resolution trajectory, the best match is the low-resolution clean estimate with the highest cosine to the full-resolution clean estimate at $\sigma$. Cosines compare the seed-specific parts of the two estimates on a common $16\times16$ grid. At high noise the noise level of the best match follows the resolution-dependent SNR map of \cref{eq:prop} \cite{simplediffusion2023,esser2024sd3} (\cref{fig:ridges}). A low resolution is usable when the noise level of its best match keeps falling as $\sigma$ falls, down to $\sigma=0.9$, so that the low-resolution run follows the full-resolution one through the whole stage. At $512^2$ this holds on FLUX.1-dev and Qwen-Image. On SD3.5-Large and FLUX.2-klein-base the noise level of the best match stalls or turns back inside the low-resolution stage at $512^2$ and keeps falling much longer at $768^2$, which is why these 2 models use ${s=3/4}$.

\paragraph{Cross-resolution agreement $R$} $R$ is the mean cosine of the best match over $\sigma\in[0.90,0.99]$, where the low-resolution stage runs (top row of \cref{fig:ridges}). It grows with the low resolution, on FLUX.1-dev from 0.48 at $256^2$ to 0.77 at $768^2$, so a higher low resolution agrees better but costs more per step. $R$ predicts the gain of \method. With the per-model lines of \cref{fig:decision}c fixed, $R$ predicts the 4-NFE gain at 3 held-out guidance scales to within 0.015 \lpips. Within each model $R$ orders the guidance scales exactly as their gains, and a line fitted on the other 2 models, leaving each model out in turn, predicts its gains with a mean absolute error of 0.016--0.028 \lpips and the correct sign. The delay of the guided decision in \cref{app:readout} ranks opposite to $R$, with a Spearman correlation of $-0.95$ over the 12 model--guidance pairs of \cref{app:readout}.

\paragraph{Final images} The final images of the two resolutions agree as well (\cref{tab:chi}). For 4 prompts with 64 seeds each, we run the default sampler at the low resolution from each seed's coarse noise. In the coarse band below 4 cycles per image, we correlate the final coordinates of the two resolutions along the 8 leading principal directions of the full-resolution seed residuals, estimated on half of the seeds and evaluated on the other half. Without guidance this correlation $\chi$ is 0.88--0.98, and at $512^2$ the low resolution's seed variance per coefficient is close to the $s^2$ that assumption (i) of \cref{prop:resolution} predicts. Run alone, the unconditional branch agrees as well. With guidance $\chi$ falls, monotonically in $g$ on every model (\cref{fig:crossres}a), and over the same 12 model--guidance pairs it ranks with $R$ at a Spearman correlation of 0.98 (\cref{fig:crossres}b).

\subsection{Why guidance separates the resolutions}
\label{app:guidance}
\begin{figure}[t]
\centering
\includegraphics[width=\columnwidth]{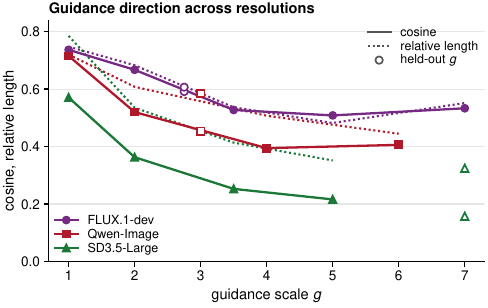}
\caption{\textbf{The guidance direction across resolutions.} The low-resolution guidance direction $v_c-v_u$ compared with its full-resolution value at the states of each guidance scale, at $512^2$, or $768^2$ on SD3.5-Large. Solid lines give the cosine and dotted lines the relative length along the full-resolution direction. Open markers are held-out guidance scales.}
\label{fig:guidance}
\end{figure}
\begin{figure*}[t]
\centering
\includegraphics[width=\textwidth]{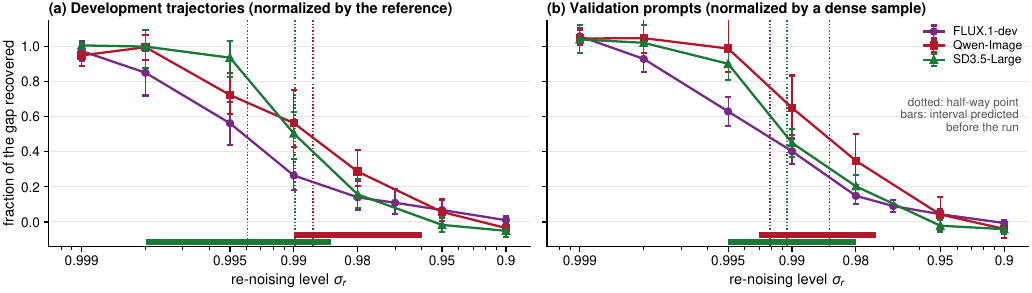}
\caption{\textbf{Re-noising.} Fraction of the fidelity gap recovered when a 3-NFE sample is re-noised to $\sigma_r$ with its original noise and continued with a dense sampler, (a)~on development trajectories, normalized by the reference, and (b)~on validation prompts, normalized by a dense sample from scratch, with 95\% intervals. Measured half-way points are dotted. The bars for Qwen-Image and SD3.5-Large are the intervals predicted beforehand from the band-agreement curves.}
\label{fig:supp_reentry}
\end{figure*}
\begin{table}[t]
\centering\footnotesize
\caption{\textbf{How much of the guidance direction survives.} Cosine of the guidance direction with its full-resolution value, given by the low-resolution model at $512^2$ and at $768^2$ for the same content and noise, and after the switch's noise replacement. On FLUX.2-klein-base the cosine after the replacement is 0.64.}
\label{tab:guidance}
\setlength{\tabcolsep}{3.5pt}
\begin{tabular}{@{}l c c c@{}}
\toprule
 & FLUX.1-dev & Qwen-Image & SD3.5-Large \\
\midrule
low-resolution model, $512^2$ & 0.78 & 0.64 & 0.08 \\
low-resolution model, $768^2$ & 0.94 & 0.85 & 0.33 \\
after the noise replacement & 0.70 & 0.33 & 0.29 \\
\bottomrule
\end{tabular}

\end{table}
Each guidance branch agrees across resolutions. Their difference, the guidance direction $v_c-v_u$, does not, and guidance amplifies it. On the guidance-distilled FLUX.1-dev, the conditional prediction is the one at embedding 1, and the guidance direction is the prediction at the native embedding minus it. The two branches' resolution errors, each branch's low-resolution prediction minus its full-resolution one, are correlated at 0.60--0.97 across FLUX.1-dev, Qwen-Image and SD3.5-Large, so most of the error cancels in the guidance direction. The remainder is still large relative to this direction, which is a small difference of two similar vectors. At the low resolution of each model its length along the full-resolution direction is only 32--79\% of the full-resolution length, and its cosine with the full-resolution direction falls from 0.57--0.74 at $g=1$ to 0.25--0.53 at the native guidance (\cref{fig:guidance}).

The deflection of the guidance direction does not come from the detail that the low resolution lacks, since band-limiting the full-resolution state leaves the direction unchanged, at a cosine of 0.99. Two small changes cause it. One is the low-resolution model's own response to the same content, and the other is the switch's noise replacement (\cref{tab:guidance}). Because the guidance direction is the difference of two similar vectors, both changes are 4 to 18 times larger in it than in the branches. Classifier-free guidance then multiplies the direction by $g-1$, and a guided model amplifies the resulting deviation instead of damping it (\cref{app:cfg}). The deflection shrinks as the low resolution grows, which is why SD3.5-Large runs its stage at ${s=3/4}$.

In the low-resolution stage we replace the conditional prediction, the guidance direction or both by their full-resolution values and measure how much of the switch error each removes. On Qwen-Image the guidance direction alone removes 83\% of what both remove. On FLUX.1-dev and SD3.5-Large the conditional prediction alone removes most of it, 55\% and 61\%. Two simpler remedies do not work. Withholding guidance while the composition is decided \cite{guidanceinterval2024} changes the target image itself, by 0.42--0.51 \lpips, and rescaling the guidance at low resolution does not improve fidelity (\cref{app:inside}).

\subsection{Decisions are not undone}
\label{app:renoise}
Re-noising a few-step sample recovers the reference only from above $\sigma=0.99$. We re-noise a 3-NFE sample to a level $\sigma_r$ with its original noise and continue it with a dense full-resolution sampler (\cref{fig:supp_reentry}). The continued sample closes its gap to the reference only if $\sigma_r$ is above 0.99. The half-way points fall inside the intervals that the band-agreement curves predicted beforehand. With fresh noise, re-noising returns to neither the reference nor the sample.

\FloatBarrier
\section{Theory}
\label{app:theory}
\noindent The theory has three levels. The scaling lemma and \cref{prop:resolution} are exact statements about probability flows under stated assumptions. A linear-Gaussian model gives the explicit gains of the switch in \cref{prop:handover,prop:floor}. Measurements on the real models test both, at the end of each subsection and in \cref{app:measure}.

\subsection{The scaling lemma}
\label{app:lemma}
Scaling the data shifts every event of the flow to another noise level.
\begin{lemma}[scaling]
\label{lem:scaling}
Let $x_0=a\,y_0$ for a scalar $a>0$, and let $\zeta(\sigma)=(1-\sigma)a+\sigma$ and $\logit\tilde\sigma=\logit\sigma-\log a$. If $y(\tilde\sigma)$ solves the probability-flow ODE of the rectified-flow path of $y_0$, then $x(\sigma)=\zeta(\sigma)\,y(\tilde\sigma(\sigma))$ solves that of $x_0$, for any distribution of $y_0$.
\end{lemma}
\noindent\textit{Proof.} On the path, $x_\sigma=(1-\sigma)a\,y_0+\sigma\epsilon$. Dividing by $\zeta$ gives $x_\sigma/\zeta=(1-\tilde\sigma)y_0+\tilde\sigma\epsilon$ with $\tilde\sigma=\sigma/\zeta$, which is the stated relation of the logits. Let $u(y,\tilde\sigma)=\mathbb E[\epsilon-y_0\mid y_{\tilde\sigma}=y]$ be the velocity of the probability flow of $y_0$. With $\mathrm d\zeta/\mathrm d\sigma=1-a$ and $\mathrm d\tilde\sigma/\mathrm d\sigma=a/\zeta^2$, substituting these derivatives shows that $\mathrm dx/\mathrm d\sigma=(\mathrm d\zeta/\mathrm d\sigma)\,x/\zeta+\zeta\,(\mathrm d\tilde\sigma/\mathrm d\sigma)\,u(x/\zeta,\tilde\sigma)$ equals $\mathbb E[\epsilon-x_0\mid x_\sigma=x]$, the velocity of the probability flow of $x_0$. \hfill$\square$

\noindent Every cosine-based quantity is invariant to the scalar $\zeta$, in particular the band agreement of the clean estimate with the final sample. The band-agreement curve of data scaled by $a$ is therefore the curve of $y_0$ shifted by $\log a=\tfrac12\log V$ in $\logit\sigma$, with $V=a^2$ the variance ratio. This is the argument behind the resolution-dependent time shift \cite{simplediffusion2023,esser2024sd3}. Applied to each band as its own flow, with its seed-specific variance, it predicts the slope $\tfrac12$ of \cref{eq:law}, and the fitted slope is 0.44 (\cref{app:law}).

\subsection{Proof of Proposition 1}
\label{app:proof}
\Cref{prop:resolution} assumes (i) that the low-resolution model's samples are distributed as the downsampled full-resolution samples $sDx_0$, and (ii) that the posterior mean of the data's coarse part depends on a noisy state only through the state's coarse part. In the proof, $D$ keeps the orthonormal Fourier coordinates below the Nyquist frequency of the low resolution, indexed by cycles per image, so $DD^\top=I$ and $D\epsilon$ is white noise. The sampler uses scaled $2\times2$ block means, $D=2\avgpool_2$, instead, which carry the same coarse noise at a cosine of 0.996 below 4 cycles per image (\cref{app:atswitch}). In these coordinates, assumption (i) says that the coarse coordinates of the low-resolution samples are those of the full-resolution samples multiplied by $s$, because downsampling a smooth image by $s$ multiplies its orthonormal coarse coordinates by $s$. Assumption (ii) holds, for example, if the coarse part of the data is independent of the fine part of $x_\sigma$ given the coarse part of $x_\sigma$.

\noindent\textit{Proof.} Write $y_\sigma=Dx_\sigma$ for the coarse coordinates of the full-resolution flow. Under (ii) their velocity $\mathbb E[D\epsilon-Dx_0\mid x_\sigma]=(y_\sigma-\mathbb E[Dx_0\mid x_\sigma])/\sigma$ depends on $y_\sigma$ alone, so they follow the probability flow of the data $Dx_0$ with the white noise $D\epsilon$ as a closed system. Under (i) the low resolution follows the probability flow of $sDx_0$ from the same noise. By the scaling lemma with $a=s$, the low-resolution state is $x^{\downarrow}(\sigma)=\zeta\,y(\tilde\sigma)$ with $\logit\tilde\sigma=\logit\sigma-\log s$ and $\zeta=(1-\sigma)s+\sigma$, and its clean estimate is $s\,\mathbb E[Dx_0\mid y_{\tilde\sigma}]=s\,D\,\xhat(\tilde\sigma)$, again by (ii). This is \cref{eq:prop}. \hfill$\square$

\noindent The low resolution then decides what the full resolution decides, linear and nonlinear parts alike, and the linear-Gaussian model below is needed only for the explicit gains of \cref{prop:handover,prop:floor}. The same argument applies to each guidance branch, so classifier-free guidance would preserve the agreement if both branches satisfied (i) and (ii) everywhere. \Cref{app:guidance} locates the loss in the guidance direction. Assumption (ii) holds only approximately, best at high noise, where the state's fine part is almost all noise.

\subsection{A linear-Gaussian model}
\label{app:linear}
\paragraph{Flow and clean estimate}
Fix a prompt and a band in orthonormal Fourier coordinates, in which the noise has unit variance per coordinate, and model the band's final latent across seeds as $\mathcal N(\mu,\Sigma)$ with $\Sigma=\sum_i\lambda_ie_ie_i^\top$. On the rectified-flow path $x_\sigma=(1-\sigma)x_0+\sigma\epsilon$ the marginals are $\mathcal N((1-\sigma)\mu,\;\Sigma_\sigma)$ with $\Sigma_\sigma=(1-\sigma)^2\Sigma+\sigma^2I$. These covariances commute, so the probability-flow ODE transports the noise $\epsilon$ along each $e_i$ independently and linearly,
\begin{equation}
x_\sigma=(1-\sigma)\mu+\sum_i\big((1-\sigma)^2\lambda_i+\sigma^2\big)^{1/2}\langle\epsilon,e_i\rangle e_i .
\end{equation}
The final sample is $x_0=\mu+\sum_i\lambda_i^{1/2}\langle\epsilon,e_i\rangle e_i$, so its seed residual is $\delta x_0=x_0-\mu$ with $\langle\delta x_0,e_i\rangle=\lambda_i^{1/2}\langle\epsilon,e_i\rangle$. The posterior mean along the trajectory is
\begin{align}
\xhat(\sigma)-\mu&=(1-\sigma)\Sigma\,\Sigma_\sigma^{-1}\big(x_\sigma-(1-\sigma)\mu\big)\nonumber\\
&=\sum_i w_i(\sigma)^{1/2}\langle\delta x_0,e_i\rangle e_i,\qquad w_i=\frac{\rho\lambda_i}{1+\rho\lambda_i},
\label{eq:gauss}
\end{align}
with $\rho=((1-\sigma)/\sigma)^2$ the signal-to-noise ratio. The clean estimate is the final sample with each principal coordinate shrunk by the gain $w_i^{1/2}$.

\paragraph{Band agreement}
The per-seed band agreement is therefore a gain-weighted cosine,
\begin{equation}
A(\sigma)=\frac{\sum_i w_i^{1/2}p_i}{\big(\sum_i w_ip_i\,\sum_ip_i\big)^{1/2}},\qquad p_i=\langle\delta x_0,e_i\rangle^2 .
\end{equation}
It equals 1 for isotropic $\Sigma$, as a deterministic linear flow points at its final sample from the start, and it falls below 1 only because directions with larger $\lambda_i$ reach their final gain earlier. A band therefore counts as decided once its agreement is high and re-noising no longer undoes it (\cref{app:renoise}). With $U$ the eigenvectors of the Gram matrix of the 64 seed residuals and $\lambda_i$ the eigenvalues of their covariance, with 63 degrees of freedom, $p_{ji}=63\,\lambda_iU_{ji}^2$ for seed $j$, so $A(\sigma)$ is a prediction from the final samples alone, with no free parameter to fit.

\paragraph{The low-resolution stage and its switch}
The same model gives the error of the switch in closed form. Let the low resolution have $s$ times the full side length. In orthonormal coordinates indexed by cycles per image, the coarse noise $D\epsilon$ has the same coarse coordinates as $\epsilon$, and a smooth image downsampled by $s$ has its coarse coordinates multiplied by $s$. Assume that the low resolution's final samples are the full-resolution ones seen at that resolution, with the same principal directions and eigenvalues $s^2\lambda_i$, and that both flows are the linear Gaussian flows above. The low-resolution clean estimate at the stage's last noise level $\sigma_e=\sigma_{k-1}$, upsampled, is then $\mu+\sum_i(w'_i\lambda_i)^{1/2}\epsilon_ie_i$ with $\epsilon_i=\langle\epsilon,e_i\rangle$, the signal-to-noise ratio $\xi_e=\rho(\sigma_e)\lambda_i$ along $e_i$ and $w'_i=s^2\xi_e/(1+s^2\xi_e)$.
\begin{proposition}[switch gain]
\label{prop:handover}
Rebuilding the state at $\sh$ with the seed's noise gives, along each $e_i$, the seed-specific part of the full-resolution flow's state multiplied by
\begin{equation}
\gamma_i=\frac{1+q_i}{(1+\xi_h)^{1/2}},\qquad \xi_h=\rho(\sh)\lambda_i ,
\end{equation}
with $q_i=s\,(\xi_h\xi_e)^{1/2}(1+s^2\xi_e)^{-1/2}$, and the flow carries it unchanged to the final sample.
\end{proposition}
\noindent\textit{Proof.} Along $e_i$ and about its mean, the full-resolution flow's state at $\sh$ is $(\sh^2+(1-\sh)^2\lambda_i)^{1/2}\epsilon_i$, and the rebuilt state is $\sh\epsilon_i+(1-\sh)(w'_i\lambda_i)^{1/2}\epsilon_i$. Their ratio is $\gamma_i$, and the flow, linear along each $e_i$, carries this factor to the final sample. \hfill$\square$

\noindent In the model, $\gamma_i$ stays close to 1 over all $\lambda_i$ (\cref{tab:gamma}), where 1 means that the rebuilt state equals the full-resolution flow's state, and every seed's coarse composition is reproduced with a cosine of at least 0.997.

\noindent\textit{Why the noise levels stay unshifted.} With a full-resolution stage in place of the low-resolution one, the same rebuild would overshoot more, with $\gamma_i$ up to 1.25, because the seed's noise is not the noise that the flow predicts at $\sh$. A low-resolution estimate is less decided than a full-resolution one at the same $\sigma$, which offsets part of this overshoot, and shifting its noise levels by the SNR map would remove the offset. This is why the noise levels of the low-resolution stage stay unshifted on FLUX.1-dev and SD3.5-Large.

\noindent Fresh noise correlates with the reference along $e_i$ only by $q_i/(1+q_i^2)^{1/2}$, which vanishes for undecided directions, so it redraws the composition in proportion to the undecided share, the ordering measured in \cref{app:atswitch}.

\noindent\textit{Test on the models.} The ordering of the three ranges in \cref{tab:gamma} matches a direct comparison on the ablation prompts. With the same 3 steps, switch and full-resolution stage, the \lpips of the low-resolution stage minus that of a full-resolution one is $-0.021$ on FLUX.1-dev, $+0.003$ on SD3.5-Large and $+0.126$ on Qwen-Image, whose SNR map removes the offset.
\begin{table}[t]
\centering\footnotesize
\caption{\textbf{Gain of the switch in the linear-Gaussian model.} Range of $\gamma_i$ over all $\lambda_i$ for the stages of the 2- to 4-NFE configurations, which switch at $\sh=0.9$. $\sigma_e$ is the stage's last noise level, after the SNR map on Qwen-Image.}
\label{tab:gamma}
\setlength{\tabcolsep}{5pt}
\begin{tabular}{@{}l c c c@{}}
\toprule
model & $s$ & $\sigma_e$ & range of $\gamma_i$ \\
\midrule
FLUX.1-dev & 1/2 & 0.933 & $[0.92, 1.05]$ \\
Qwen-Image, SNR map & 1/2 & 0.875 & $[1.00, 1.16]$ \\
SD3.5-Large & 3/4 & 0.95 & $[0.94, 1.06]$ \\
\bottomrule
\end{tabular}

\end{table}

\paragraph{Transfer of a switch deviation} A deviation in the state at the switch reaches the final sample with a gain set by how far its direction is decided.

\par\smallskip\noindent\begin{minipage}{\linewidth}
\begin{proposition}[transfer of a switch deviation]
\label{prop:floor}
In the linear Gaussian model, a deviation $\Delta_i$ in the switched clean estimate along $e_i$, moving the state at $\sh$ by $(1-\sh)\Delta_i$, reaches the final sample multiplied by
\begin{equation}
G_i=\frac{(1-\sh)\lambda_i^{1/2}}{\big((1-\sh)^2\lambda_i+\sh^2\big)^{1/2}}=\Big(\frac{\xi_h}{1+\xi_h}\Big)^{1/2}.
\end{equation}
\end{proposition}
\end{minipage}
\noindent\textit{Proof.} The flow maps the state at $\sh$ to the final sample linearly, with the ratio of the standard deviations $\lambda_i^{1/2}/((1-\sh)^2\lambda_i+\sh^2)^{1/2}$. Hence $G_i^2=w_i(\sh)$, the gain of \cref{eq:gauss} at the switch. \hfill$\square$

\noindent The gain depends on how far a direction is decided at the switch and not on how the full-resolution stage is discretized. A deviation along a decided direction, $\xi_h\gg1$, survives unchanged. One along an undecided direction, $\xi_h\ll1$, is decided again by the full-resolution stage from the seed's noise. The error of \method thus splits into two parts. One is resolution-specific and is carried with gain $\approx1$ along the directions decided before $\sh$, regardless of the full-resolution stage. The other is that stage's discretization error, which falls with its length.

\subsection{Guidance amplifies a cross-resolution deviation}
\label{app:cfg}
\begin{table}[t]
\centering\footnotesize
\caption{\textbf{Share of a cross-resolution deviation that reaches the clean estimate} at native-guidance states. The guided estimate at scale $g'$ is measured in band $b_2$, and $g^*$ is the predicted crossing of 1.}
\label{tab:kappa}
\setlength{\tabcolsep}{3.5pt}
\begin{tabular}{@{}l c c c c@{}}
\toprule
model & each branch & $g'$ & guided estimate & $g^*$ \\
\midrule
Qwen-Image & 0.12--0.52 & 1, 2, 4, 6 & 0.38, 0.65, 1.11, 1.44 & 3.7 \\
SD3.5-Large & 0.50--0.77 & 1, 2, 3.5, 5 & 0.66, 0.89, 1.36, 1.86 & 3.0 \\
\bottomrule
\end{tabular}

\end{table}
For data $\mathcal N(\mu,\Sigma)$, the Jacobian of the posterior mean with respect to the state is $(1-\sigma)\Sigma((1-\sigma)^2\Sigma+\sigma^2I)^{-1}$. Writing a content deviation $\eta$ into the state as $(1-\sigma)\eta$ therefore moves the clean estimate by
\begin{equation}
J\eta=\rho\Sigma\,(\rho\Sigma+I)^{-1}\eta,
\end{equation}
where $J$ is the Jacobian of the clean estimate with respect to the content deviation. Its eigenvalues are the gains $w_i$ of \cref{eq:gauss}, all below one, so each branch \emph{contracts} a deviation of its own content. Classifier-free guidance combines the Jacobians $J_c$ and $J_u$ of the conditional and the unconditional branch into $J_g=J_u+g(J_c-J_u)=J_c+(g-1)J_\Delta$, with $J_\Delta=J_c-J_u$. Let $\kappa_c=\|J_c\eta\|/\|\eta\|$ and $\kappa_\Delta=\|J_\Delta\eta\|/\|\eta\|$. When $J_c\eta$ and $J_\Delta\eta$ are nearly orthogonal, $\|J_g\eta\|\approx(\kappa_c^2+(g-1)^2\kappa_\Delta^2)^{1/2}\|\eta\|$, which exceeds $\|\eta\|$ beyond
\begin{equation}
g^*\approx1+\frac{(1-\kappa_c^2)^{1/2}}{\kappa_\Delta}.
\end{equation}

\noindent\textit{Test on the models.} The actual cross-resolution deviation is the low-resolution clean estimate minus the full-resolution one. We insert it into native-guidance states as the switch does, rebuilt with the seed's noise, and then evaluate each branch and their combination at guidance scales $g'$ (\cref{tab:kappa}). Each branch passes only part of the deviation, whereas beyond the crossing $g^*$ that the branch gains predict, the guided estimate amplifies it, with a gain above 1. The guidance-distilled FLUX.1-dev shows the same rise, from 0.62 at $g'=1$ to 1.57 at $g'=5$. Random deviations of the same norm pass with a gain of only 0.02--0.34 through every combination. The amplification thus acts on content that the model finds plausible, which is what a switch from another resolution supplies.

\FloatBarrier
\section{The sampler in detail}
\label{app:sampler}
\begin{algorithm}[t]
\caption{\method}
\label{alg:method}
\small
\begin{algorithmic}[1]
\Require seed noise $\epsilon$, prompt $c$, coarse-noise map $D$;\newline low-resolution steps $k$, switch point $\sh$;\newline full-resolution noise levels $\sh=\tau_0>\dots>\tau_T=0$
\State $x\gets D\epsilon$ \Comment{coarse part of the seed's noise}
\State $\sigma_j\gets1-j(1-\sh)/k$ for $j=0,\dots,k-1$
\For{$j=0,\dots,k-1$} \Comment{low-resolution stage}
\State $v\gets v_\theta(x,\sigma_j,c)$;\; $\xhat^{\downarrow}\gets x-\sigma_jv$;\; $\hat\epsilon\gets\xhat^{\downarrow}+v$
\State \textbf{if} $j<k-1$ \textbf{then} $x\gets x+(\sigma_{j+1}-\sigma_j)\,v$
\EndFor
\State $\epsilon_h\gets\epsilon$ \Comment{or $(I-D^\top D)\epsilon+D^\top\hat\epsilon$, \cref{eq:switchnoise}}
\State $x\gets\sh\,\epsilon_h+(1-\sh)\,\up\big(\xhat^{\downarrow}\big)$ \Comment{switch, \cref{eq:switch}}
\For{$i=0,\dots,T-1$} \Comment{full-resolution stage}
\State $x\gets x+(\tau_{i+1}-\tau_i)\,v_\theta(x,\tau_i,c)$
\EndFor
\State \Return $\mathrm{decode}(x)$
\end{algorithmic}
\end{algorithm}

\subsection{Configurations}
\label{app:config}
\begin{table*}[t]
\centering\footnotesize
\caption{\textbf{Configurations at 2 to 4 NFEs.} $k$ low-resolution steps at scale $s$, the switch point $\sh$, the full-resolution noise levels from $\sh$ down, the first-step shrink, the map of the low-resolution noise levels, and the coarse switch noise. On FLUX.1-dev, Qwen-Image and SD3.5-Large the 4-NFE noise levels below $\sh$ are those of SharpEuler's 4-step grid, and FLUX.2-dev reuses FLUX.1-dev's.}
\label{tab:config}
\setlength{\tabcolsep}{4pt}
\begin{tabular}{@{}l c c c l l l c l l@{}}
\toprule
 & & & & \multicolumn{3}{c}{full-resolution noise levels} & & & \\
\cmidrule(lr){5-7}
model & $s$ & $k$ & $\sh$ & 2 NFEs & 3 NFEs & 4 NFEs & shrink & noise-level map & coarse switch noise \\
\midrule
FLUX.1-dev & 1/2 & 3 & 0.9 & $[0.9]$ & $[0.9, 0.7]$ & $[0.9, 0.76, 0.51]$ & -- & identity & seed \\
FLUX.2-dev & 1/2 & 3 & 0.9 & $[0.9]$ & $[0.9, 0.7]$ & $[0.9, 0.76, 0.51]$ & -- & identity & seed \\
Qwen-Image & 1/2 & 3 & 0.9 & $[0.9]$ & $[0.9, 0.7]$ & $[0.9, 0.67, 0.4]$ & 0.50 & SNR map & seed, stage's estimate at 4 NFEs \\
SD3.5-Large & 3/4 & 2 & 0.9 & $[0.9]$ & $[0.9, 0.6]$ & $[0.9, 0.75, 0.5]$ & 0.70 & identity & seed \\
FLUX.2-klein-base & 3/4 & 2, 2, 4 & 0.95 & $[0.95]$ & $[0.95, 0.767]$ & $[0.95, 0.767]$ & -- & identity & seed \\
\bottomrule
\end{tabular}

\end{table*}
\begin{table*}[t]
\centering\footnotesize
\caption{\textbf{Configurations beyond 4 NFEs.} The extra budget goes to a full-resolution stage on SharpEuler's grid. Qwen-Image and SD3.5-Large take the coarse switch noise from the stage's own noise estimate, and Qwen-Image keeps the SNR map at every budget.}
\label{tab:config_more}
\setlength{\tabcolsep}{5pt}
\begin{tabular}{@{}l c l c c l@{}}
\toprule
model & NFEs & low-resolution stage & shrink & $\sh$ & full-resolution noise levels \\
\midrule
FLUX.1-dev & 5 & $3\times512^2$ & -- & 0.95 & $[0.95, 0.837, 0.612, 0.387]$ \\
 & 6 & $3\times512^2$ & -- & 0.95 & $[0.95, 0.851, 0.735, 0.513, 0.311]$ \\
 & 8 & $3\times512^2$ & -- & 0.95 & $[0.95, 0.913, 0.858, 0.785, 0.644, 0.441, 0.260]$ \\
\midrule
Qwen-Image & 5 & $3\times512^2$ & 0.50 & 0.90 & $[0.90, 0.796, 0.572, 0.333]$ \\
 & 6 & $3\times512^2$ & -- & 0.95 & $[0.95, 0.864, 0.708, 0.500, 0.286]$ \\
 & 8 & $3\times512^2$ & -- & 0.95 & $[0.95, 0.879, 0.796, 0.681, 0.511, 0.364, 0.200]$ \\
\midrule
SD3.5-Large & 5 & $2\times768^2$ & 0.70 & 0.90 & $[0.90, 0.831, 0.600, 0.375]$ \\
 & 6 & $4\times768^2$ & -- & 0.95 & $[0.95, 0.831, 0.600, 0.375]$ \\
 & 8 & $4\times768^2$ & -- & 0.95 & $[0.95, 0.915, 0.849, 0.732, 0.500, 0.300]$ \\
\bottomrule
\end{tabular}

\end{table*}
One sampler serves all models. Four settings depend on the model, and each follows from a measurement (\cref{tab:config,tab:newmodel}). The low-resolution stage takes $k$ noise levels spaced uniformly from 1 towards $\sh$ as in \cref{alg:method}, 3 at $512^2$ or 2 to 4 at $768^2$. Its last evaluation gives the clean estimate for the switch, and the stage takes no Euler step after it. The switch upsamples this estimate bicubically and rebuilds the state with the seed's noise at $\sh$ (\cref{eq:switch}), and the full-resolution stage takes Euler steps on the listed noise levels. The four settings are the following.
\begin{itemize}[leftmargin=1.1em,itemsep=1pt,topsep=2pt]
\item \textit{Resolution.} The low resolution is ${s=1/2}$. SD3.5-Large and FLUX.2-klein-base use ${s=3/4}$, because their half-resolution runs agree poorly with the full resolution inside the stage (\cref{app:crossres,app:guidance}).
\item \textit{Noise levels.} The noise levels of the low-resolution stage, line 2 of \cref{alg:method}, stay unshifted except on Qwen-Image, which shifts them as in \cref{eq:prop}, by the SNR map $\sigma\mapsto\sigma/(2-\sigma)$, and \cref{app:linear} explains why the others keep them unshifted.
\item \textit{First-step shrink.} Up to 5 NFEs, the CFG models damp the first low-resolution clean estimate, line 4 of \cref{alg:method}, by the first-step shrink of SE + shrink, 0.50 on Qwen-Image and 0.70 on SD3.5-Large (\cref{tab:calib}).
\item \textit{Switch noise.} The switch takes the seed's noise, line 6 of \cref{alg:method}. From 4 NFEs on Qwen-Image and 5 on SD3.5-Large, its coarse part comes from the stage's own noise estimate instead (\cref{app:noise}).
\end{itemize}
Beyond 4 NFEs the extra budget lengthens the full-resolution stage on SharpEuler's grid, and the switch moves to $\sh=0.95$ as that stage grows (\cref{tab:config_more}).

\paragraph{Transfer of a fixed configuration} FLUX.2-dev uses FLUX.1-dev's configuration unchanged, with no setting tuned for it. FLUX.1-dev's configuration also carries over to Qwen-Image. Applied unchanged, it comes within 0.006 \lpips of Qwen-Image's own configuration at 2 and 3 NFEs and within 0.024 at 4 NFEs, where it is still 0.15 \lpips closer to the reference than the default sampler. On Qwen-Image the model-specific settings thus refine the gain rather than create it.

\subsection{Cost}
\label{app:cost}
\begin{table*}[t]
\centering\footnotesize
\caption{\textbf{Cost of \method.} The cost of one low-resolution evaluation relative to a full-resolution one, in measured time on an idle GPU and in FLOPs, and the NFE count of each configuration. Dashes mark models without a FLOP count.}
\label{tab:cost}
\setlength{\tabcolsep}{6pt}
\begin{tabular}{@{}l c c c c c c@{}}
\toprule
 & & \multicolumn{2}{c}{cost of a low-resolution evaluation} & \multicolumn{3}{c}{NFEs in total} \\
\cmidrule(lr){3-4}\cmidrule(l){5-7}
model & low resolution & time & FLOPs & budget 2 & budget 3 & budget 4 \\
\midrule
FLUX.1-dev & $512^2$ & 0.318 & 0.29 & 1.95 & 2.95 & 3.95 \\
FLUX.2-dev & $512^2$ & 0.304 & -- & 1.91 & 2.91 & 3.91 \\
Qwen-Image & $512^2$ & 0.337 & 0.22 & 2.01 & 3.01 & 4.01 \\
SD3.5-Large & $768^2$ & 0.541 & 0.54 & 2.08 & 3.08 & 4.08 \\
FLUX.2-klein-base & $768^2$ & 0.568 & -- & 2.14 & 3.14 & 4.27 \\
\bottomrule
\end{tabular}

\end{table*}
At half resolution a low-resolution step costs about a third of a full-resolution one (\cref{tab:cost}), and measured seconds follow the NFE count within 4\%. One NFE is one evaluation at full resolution, and the two branches of a classifier-free-guidance step count as one. A low-resolution evaluation counts as its measured time relative to a full-resolution one on an idle GPU, so a configuration with $k$ low-resolution and $T$ full-resolution evaluations costs $T+k\,r$ NFEs, with $r$ the cost ratio of \cref{tab:cost}. FLUX.1-dev and FLUX.2-dev evaluate one branch per step and the CFG models two, so the same NFE count costs different seconds on different models, and every comparison reports seconds. At 4 NFEs the time of \method is within 2.4\% of that of SE + shrink on FLUX.1-dev, Qwen-Image and SD3.5-Large. The first-step shrink adds no evaluation, and peak memory is unchanged.

Counted with PyTorch's FLOP counter on shapes alone, every guidance branch included, the 2- to 4-NFE images of \method on FLUX.1-dev, Qwen-Image and SD3.5-Large need 12--29$\times$ fewer FLOPs than the default 50-step sampler. Read from the GPU's energy counter, they need 11--24$\times$ less energy.

\subsection{Setting the switch point}
\label{app:switch}
\begin{figure*}[t]
\centering
\includegraphics[width=\textwidth]{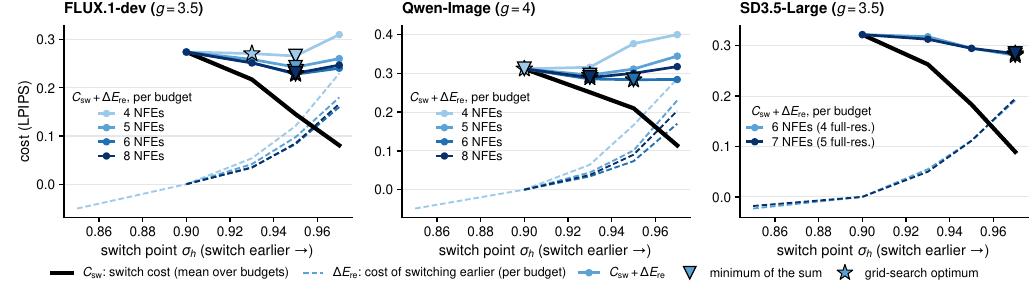}
\caption{\textbf{The switch point is the balance of two measured costs.} 30 validation prompts at the native guidance. The black curve is the switch cost $C_{\mathrm{sw}}$ averaged over the swept budgets, over which it changes little. Dashed curves give the cost of switching earlier, $\Delta E_{\mathrm{re}}$, per budget, and solid curves give the sum $C_{\mathrm{sw}}+\Delta E_{\mathrm{re}}$, which is the error of \method up to a constant. Its minimum $\blacktriangledown$ is the grid search's switch point $\bigstar$ at every budget. SD3.5-Large has full-resolution stages of 4 and 5 evaluations.}
\label{fig:curves}
\end{figure*}
\begin{figure*}[t]
\centering
\includegraphics[width=\textwidth]{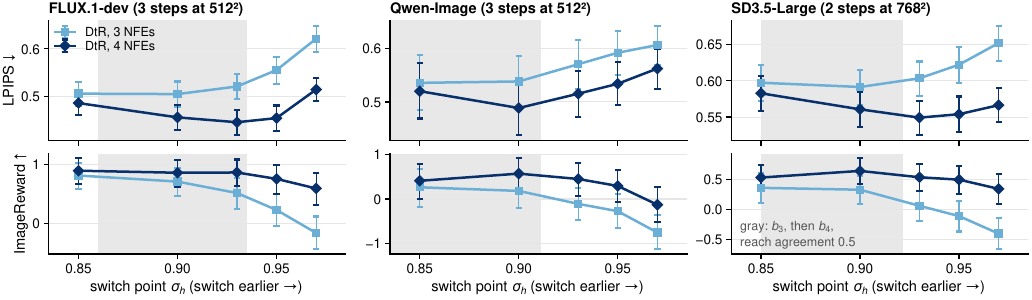}
\caption{\textbf{Switch-point sweep.} \lpips above and \ir below against $\sh$ on validation prompts, 3 NFEs light and 4 dark, with the full-resolution noise levels fixed and 95\% intervals. Gray spans the noise levels from where $b_3$ to where $b_4$ reaches an agreement of 0.5.}
\label{fig:supp_sigmah}
\end{figure*}
The switch point balances two costs measured on a few stored trajectories, and the switch cost measured at one budget serves every budget. All errors in this subsection are \lpips to the reference. Let $E_{\mathrm{DtR}}(\sh)$ be the error of \method with switch point $\sh$, and $E_{\mathrm{re}}(\sh)$ the error that its full-resolution stage reaches when it starts from the reference trajectory's own state at $\sh$. This re-entry is a measurement, not a sampler. It shows the error that would remain if the state at the switch were exactly the reference's. Two costs follow from it.
\begin{itemize}[leftmargin=1.1em,itemsep=1pt,topsep=2pt]
\item The switch cost $C_{\mathrm{sw}}(\sh)=E_{\mathrm{DtR}}(\sh)-E_{\mathrm{re}}(\sh)$ is what \method loses by deciding at low resolution and switching.
\item The cost of switching earlier, at a higher $\sh$, is $\Delta E_{\mathrm{re}}(\sh)=E_{\mathrm{re}}(\sh)-E_{\mathrm{re}}(0.90)$, the extra error of a full-resolution stage that starts at $\sh$ rather than at 0.90.
\end{itemize}
Their sum is $E_{\mathrm{DtR}}(\sh)$ minus the constant $E_{\mathrm{re}}(0.90)$, so its minimum is the best switch point (\cref{fig:curves}).

The two costs move in opposite directions. The switch cost falls as $\sh$ rises, by nearly the same amount at every budget, and at each switch point its standard deviation across the budgets of \cref{fig:curves} is at most 0.02. The cost of switching earlier rises with $\sh$ and falls with the budget. The minimum of their sum lies within one grid step of the searched optimum in all 26 searched settings, on FLUX.1-dev and Qwen-Image at 3 guidance scales and 4 budgets each and on SD3.5-Large at 2 budgets. Because the switch cost changes little across budgets, measuring it at one budget predicts the switch point at the others. The prediction is within one grid step for 53 of the 56 pairs of budgets, and the 3 misses lie where the summed curve is flat.

On FLUX.1-dev and Qwen-Image the searched optimum is 0.90--0.93 at 4 NFEs. On SD3.5-Large at 6 and 7 NFEs the error is flat between 0.95 and 0.97.

Choosing $\sh$ per prompt does not help. An oracle that picks it per prompt on one seed gains nothing on another seed in any of 8 model--budget settings, so one switch point per model and budget suffices.

A direct sweep confirms the rule (\cref{fig:supp_sigmah}). With the full-resolution noise levels fixed, the optimum lies in the gray interval of \cref{fig:supp_sigmah}, between the noise levels at which $b_3$ and $b_4$ reach an agreement of 0.5, or at its edge. At 3 NFEs it is 0.85--0.90 on FLUX.1-dev, Qwen-Image and SD3.5-Large, and at 4 it moves to 0.90--0.95. A shorter full-resolution stage starts with a larger step, so \method then switches lower and lets the low-resolution stage decide more. As the full-resolution stage grows, the optimum moves to $\sh=0.95$, from 5 NFEs on FLUX.1-dev and from 6 on Qwen-Image and SD3.5-Large (\cref{tab:config_more}). With 5 to 11 full-resolution steps, switching there instead of at 0.9 gains 0.026--0.057 \lpips on these models.

\subsection{Which noise at the switch}
\label{app:noise}
\begin{table}[t]
\centering\footnotesize
\caption{\textbf{The stage's own noise estimate at the switch.} Paired \lpips change when the coarse switch noise is the stage's own noise estimate instead of the seed's, in the configurations of \cref{tab:config,tab:config_more}. With the estimate, \ir is on par and \ps is higher. Every listed change has an interval that excludes zero. At 2 and 3 NFEs the change is below 0.01, and the seed's noise is kept.}
\label{tab:noise}
\setlength{\tabcolsep}{5pt}
\begin{tabular}{@{}l c c c c@{}}
\toprule
model & 4 NFEs & 5 NFEs & 6 NFEs & 8 NFEs \\
\midrule
Qwen-Image & $-0.018$ & $-0.016$ & $-0.032$ & $-0.028$ \\
SD3.5-Large & -- & $-0.012$ & $-0.021$ & $-0.025$ \\
\bottomrule
\end{tabular}

\end{table}
\begin{figure}[t]
\centering
\includegraphics[width=\columnwidth]{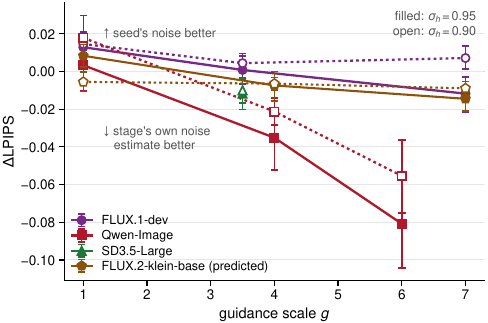}
\caption{\textbf{Which noise at the switch.} $\Delta$\lpips of the stage's own noise estimate minus the seed's noise as the coarse switch noise, against the guidance scale, filled at $\sh=0.95$ and open at $0.90$. Below zero the stage's own noise estimate is better. Pentagons mark FLUX.2-klein-base, whose preference was predicted before its test run (\cref{app:heldout}).}
\label{fig:noise}
\end{figure}
The fine noise always comes from the seed. The coarse noise comes from the seed on FLUX.1-dev, FLUX.2-dev and FLUX.2-klein-base. On Qwen-Image from 4 NFEs on and on SD3.5-Large from 5 NFEs on, it comes from the stage's own noise estimate $\hat\epsilon$. With that estimate the switch noise is
\begin{equation}
\epsilon_h=(I-P)\,\epsilon+D^\top\hat\epsilon,\qquad\hat\epsilon=\xhat^{\downarrow}+v,
\label{eq:switchnoise}
\end{equation}
where $v$ and $\xhat^{\downarrow}$ are the velocity and the clean estimate of the last low-resolution evaluation and $P=D^\top D$. On the rectified-flow path the velocity is the noise minus the clean sample, so $\hat\epsilon$ is the noise that $v$ implies. For a $512^2$ stage $D=2\avgpool_2$, so $D^\top$ is half of nearest-neighbor upsampling and $P$ is the $2\times2$ block mean. For the $768^2$ stage of SD3.5-Large, $P$ keeps the band below that resolution's Nyquist frequency and is taken with a Fourier projector. Both noises cost the same evaluations and seconds.

The seed's noise has two opposite effects. It corrects the stage's estimation drift, but re-inserting it rotates the guidance direction at the first full-resolution step. On the guidance-distilled FLUX.1-dev the drift matters more up to the native scale, so the seed's noise is better. On Qwen-Image and SD3.5-Large the guidance direction keeps a cosine of only 0.33 and 0.29 under the replacement, against 0.70 on FLUX.1-dev (\cref{tab:guidance}), so the stage's own noise estimate is better at larger budgets. On every model the preference moves towards the stage's own noise estimate as the guidance scale grows, on Qwen-Image from $+0.003$ \lpips at $g=1$ to $-0.081$ at $g=6$ (\cref{fig:noise}). \Cref{tab:noise} gives the change in the configurations.

When the two coarse noises are mixed, \lpips changes monotonically with the mixing weight on Qwen-Image and stays flat on FLUX.1-dev, so the choice of one of the two is optimal and adds no parameter. The fine bands must come from the seed. With no fine noise at the switch, \lpips rises by 0.240 on FLUX.1-dev and by 0.091 on Qwen-Image. \Cref{tab:ablation} and \cref{app:atswitch} give the ablations of the switch noise, and \cref{app:guidance,app:linear} the analysis of its two effects.

\subsection{How many low-resolution steps, at which \texorpdfstring{\mbox{resolution}}{resolution}}
\label{app:steps}
\begin{figure*}[t]
\centering
\includegraphics[width=\textwidth]{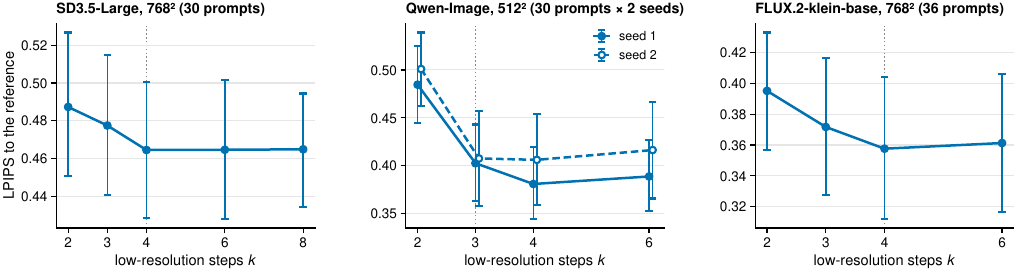}
\caption{\textbf{Number of low-resolution steps on the other models.} The same switch at $\sh=0.95$ and the full-resolution stage of the 6-NFE configuration at every $k$, with 95\% intervals. SD3.5-Large on 30 validation prompts and FLUX.2-klein-base on its 36 verification prompts saturate with 4 steps at $768^2$, and Qwen-Image on 2 seeds of 30 validation prompts with 3 to 4 steps at $512^2$. Dotted lines mark the adopted $k$.}
\label{fig:supp_k3}
\end{figure*}
\begin{table*}[t]
\centering\footnotesize
\caption{\textbf{From measurements to settings.} Measurements are the cost of one low-resolution evaluation at the chosen resolution (\cref{tab:cost}), whether the cross-resolution match holds through the stage at $512^2$ (\cref{fig:ridges}), and the cosine of the guidance direction after the switch's noise replacement (\cref{tab:guidance}). Where the match breaks at $512^2$, it holds much longer at $768^2$. The coarse switch noise is the seed's or the stage's own noise estimate. FLUX.2-dev reuses the settings of FLUX.1-dev.}
\label{tab:newmodel}
\setlength{\tabcolsep}{4pt}
\begin{tabular}{@{}l c c c c l l l@{}}
\toprule
 & \multicolumn{3}{c}{measurements} & \multicolumn{4}{c}{settings} \\
\cmidrule(lr){2-4}\cmidrule(l){5-8}
model & step cost & match at $512^2$ & guidance cosine & $s$ & $k$ & $\sh$ & coarse switch noise \\
\midrule
FLUX.1-dev & 0.318 & holds & 0.70 & 1/2 & 3 & 0.9, 0.95 from 5 NFEs & seed \\
FLUX.2-dev & 0.304 & \multicolumn{2}{c}{as FLUX.1-dev} & 1/2 & 3 & 0.9 & seed \\
Qwen-Image & 0.337 & holds & 0.33 & 1/2 & 3 & 0.9, 0.95 from 6 NFEs & estimate from 4 NFEs \\
SD3.5-Large & 0.541 & breaks & 0.29 & 3/4 & 2, 4 from 6 NFEs & 0.9, 0.95 from 6 NFEs & estimate from 5 NFEs \\
FLUX.2-klein-base & 0.568 & breaks & 0.64 & 3/4 & 2, 4 from 4 NFEs & 0.95 & seed \\
\bottomrule
\end{tabular}

\end{table*}
A few low-resolution steps suffice, and more do not help. On FLUX.1-dev the low-resolution stage saturates at 3 steps (\cref{fig:decision}b). With the full-resolution stage of the 4-NFE configuration, 24 full-resolution steps before the switch reach 0.256 \lpips. More half-resolution steps do not get there, as 24 or 48 of them are no more faithful than 3 (\cref{tab:ablation_lowres}). The $512^2$ model decides part of the mid-frequency content differently (\cref{app:guidance}). The other models saturate as well, SD3.5-Large and FLUX.2-klein-base with 4 steps at $768^2$ and Qwen-Image with 3 to 4 steps at $512^2$ (\cref{fig:supp_k3}). \Cref{app:inside} gives the further ablations inside the stage.

A lower resolution of equal cost is less faithful. A $384^2$ stage of 4 steps is 0.048 \lpips farther from the reference than the $512^2$ stage on the development trajectories and 0.030 farther on the ablation prompts. On Qwen-Image, 3 steps at $512^2$ are on par with 2 at $768^2$ in \lpips and ahead in \ds and \ir at matched cost.

\subsection{Configuring \method for a new model}
\label{app:newmodel}
\begin{algorithm}[t]
\caption{Configuring \method for a new model}
\label{alg:newmodel}
\small
\begin{algorithmic}[1]
\Require 50-step reference trajectories of a few prompts and seeds, a target budget
\State \textbf{Low resolution.} Take ${s=1/2}$ if an evaluation there is cheap enough to buy 3 low-resolution steps and the cross-resolution match holds down to $\sigma=0.9$, and ${s=3/4}$ otherwise (\cref{app:crossres}).
\State \textbf{Steps.} Add low-resolution steps until fidelity stops improving (\cref{app:steps}).
\State \textbf{Switch point.} Measure $C_{\mathrm{sw}}$ at one budget and $\Delta E_{\mathrm{re}}$ at the target budget, and take the minimum of their sum (\cref{app:switch}).
\State \textbf{Switch noise.} Keep the seed's noise if the guidance direction keeps most of itself under the noise replacement. Otherwise compare it with the stage's own noise estimate at the target budget (\cref{app:noise}).
\end{algorithmic}
\end{algorithm}
The low resolution, the switch point and the switch noise follow from diagnostics on a few dozen stored trajectories. \Cref{alg:newmodel} lists the procedure, and \cref{tab:newmodel} shows the measurements behind each model's settings. Setting the switch point at a target budget needs 36 reference trajectories of 50 steps, from 12 prompts with 3 seeds each, and runs of \method and of the re-entry at 5 switch points. This takes about 25 GPU-minutes on FLUX.2-klein-base, against 80 minutes for a full grid search over 5 switch points and 5 budgets. Estimated from evaluation counts, it takes 17--58 minutes on the other models. The first-step shrink and the noise-level map are set on validation prompts. \Cref{app:heldout} tests the procedure as a prediction on a held-out model.

\section{Experimental setup}
\label{app:setup}

\subsection{Prompts and seeds}
\label{app:prompts}
\begin{table*}[t]
\centering\footnotesize
\caption{\textbf{Prompt sets.} Each set's size, source and use. Test prompts are disjoint from all other sets.}
\label{tab:prompts}
\setlength{\tabcolsep}{5pt}
\begin{tabular}{@{}l p{0.27\textwidth} l p{0.25\textwidth}@{}}
\toprule
set & prompts $\times$ seeds & source & used for \\
\midrule
development & $12\times3$ per model, $8\times3$ on Qwen-Image, 50-step trajectories stored & MJHQ-30K & Supp.~\ref{app:measure}, \cref{fig:decision}b, noise rows of \cref{tab:ablation} \\
validation & $60\times2$, $30\times2$ on Qwen-Image, $60\times1$ on FLUX.2-dev & MJHQ-30K & configurations, calibrated schedule \\
diagnostics & first 30 validation prompts $\times\,1$, all 60 on FLUX.2-dev, 50-step trajectories stored & MJHQ-30K & re-entry, \cref{fig:decision}a, Supp.~\ref{app:switch}, \ref{app:lead} \\
ablation & $100\times1$, ten per category & MJHQ-30K & ablations marked as such \\
test & $1000\times1$ on FLUX.1-dev, 100 per category, and the first 500 on the other models, 50 per category & MJHQ-30K & main results \\
seed selection & $200\times8$ & MJHQ-30K & Supp.~\ref{app:select}, \cref{tab:select} \\
linear read-out & $4\times64$ per model, and 4 validation prompts & MJHQ-30K & Supp.~\ref{app:readout}, \ref{app:crossres} \\
GenEval & $553\times4$, the first seed for the 50-step sampler & GenEval & Supp.~\ref{app:geneval}, \ref{app:heldout}, \ref{app:distill} \\
held-out model, development & $12\times3$ & GenEval, hard prompts & Supp.~\ref{app:heldout} \\
held-out model, verification & $36\times1$ & GenEval, other hard prompts & Supp.~\ref{app:heldout} \\
video, tuning & $16\times1$ & written for this study & Supp.~\ref{app:video} \\
video, held-out & $100\times1$ & written for this study & Supp.~\ref{app:video} \\
VBench & 200 prompts & VBench's 6 quality dimensions & Supp.~\ref{app:video} \\
\bottomrule
\end{tabular}

\end{table*}
The test prompts are disjoint from every set used for design or configuration (\cref{tab:prompts}). The image prompts come from MJHQ-30K \cite{mjhq2024}, except the GenEval prompts and the development and verification prompts of the held-out model. The video tuning and held-out prompts were written for this study. The 1000 test prompts take 100 from each of the 10 MJHQ categories with one seed each. FLUX.1-dev uses all of them, and the other models use the first 500, 50 per category, with the same seeds.

\subsection{Metrics}
\label{app:metrics}
\lpips uses AlexNet features on images resized to $512^2$. \ds, \ir and \ps use their official checkpoints at their native resolutions. DINO is the cosine between the DINOv2 ViT-B/14 \cite{dinov2_2023} class tokens of a sample and its reference, resized to 256 and center-cropped to 224 pixels. For scale, DINO between the references of two different prompts is about 0.05. \fd is the Fr\'echet distance of the same features to the set of references. GenEval follows the official protocol, 553 prompts with 4 seeds each and object-detector scoring. The default 50-step sampler runs the first seed only, so comparisons with it are paired on those 553 images.

\subsection{Timing}
\label{app:timing}
\begin{table*}[t]
\centering\footnotesize
\caption{\textbf{End-to-end speed-ups} of \method against the default 50-step sampler and against the pipeline's default step count, 28 steps on FLUX.1-dev and SD3.5-Large and 50 on Qwen-Image and FLUX.2-dev.}
\label{tab:e2efull}
{\setlength{\tabcolsep}{3pt}\begin{tabular}{@{}l r r r r r@{}}
\toprule
model & default, 50 steps (s) & pipeline default (s) & \method, 2 / 3 / 4 NFEs (s) & speed-up over 50 steps & over pipeline default \\
\midrule
FLUX.1-dev & 15.23 & 8.58 (28 steps) & 0.71 / 1.01 / 1.31 & 21.5 / 15.0 / 11.6$\times$ & 12.1 / 8.5 / 6.5$\times$ \\
FLUX.2-dev & 51.01 & 51.01 (50 steps) & 2.07 / 3.09 / 4.11 & 24.6 / 16.5 / 12.4$\times$ & 24.6 / 16.5 / 12.4$\times$ \\
Qwen-Image & 29.94 & 29.94 (50 steps) & 1.35 / 1.94 / 2.54 & 22.2 / 15.4 / 11.8$\times$ & 22.2 / 15.4 / 11.8$\times$ \\
SD3.5-Large & 14.29 & 8.05 (28 steps) & 0.70 / 0.98 / 1.27 & 20.4 / 14.5 / 11.3$\times$ & 11.5 / 8.2 / 6.4$\times$ \\
\bottomrule
\end{tabular}
}
\end{table*}
Every image method is timed with the same protocol on the same GPU, and every speed-up divides the seconds of the default 50-step sampler by those of a configuration (\cref{tab:e2efull}). Seconds are measured on one RTX PRO 6000 Blackwell GPU in bf16 with batch size 1, from the seeded noise draw, including the low-resolution noise $D\epsilon$, to the decoded image. Text embeddings are precomputed for every method alike, and encoding would add 15--49\,ms per image. Per model, the default 50-step sampler and the samplers built on the default pipeline are timed in one run, interleaved across configurations. Each configuration is timed on 20 prompts after 2 warm-up images, and we report the median. Methods with their own code bases are timed in separate runs with the same protocol and the same 50-step anchor. Configurations that run the same evaluations, such as the two choices of switch noise, are reported with the same seconds.

\subsection{Full-resolution schedules and the calibrated schedule}
\label{app:schedules}
\begin{table*}[t]
\centering\footnotesize
\caption{\textbf{Time shift and first-step shrink of the full-resolution schedules.} The time shift is the scheduler parameter $\mu$ on FLUX.1-dev, FLUX.2-dev and Qwen-Image and the static shift factor on SD3.5-Large. The calibrated schedule of FLUX.2-dev keeps the default shift at 4 NFEs.}
\label{tab:calib}
\setlength{\tabcolsep}{6pt}
\begin{tabular}{@{}l c c c c c c c c@{}}
\toprule
 & default & \multicolumn{6}{c}{calibrated schedule} & SE + shrink \\
\cmidrule(lr){3-8}
 & & \multicolumn{2}{c}{2 NFEs} & \multicolumn{2}{c}{3 NFEs} & \multicolumn{2}{c}{4 NFEs} & \\
\cmidrule(lr){3-4}\cmidrule(lr){5-6}\cmidrule(lr){7-8}
model & shift & shift & shrink & shift & shrink & shift & shrink & shrink \\
\midrule
FLUX.1-dev & 1.15 & 2.5 & 0.69 & 2.0 & 0.50 & 2.5 & 0.69 & 0.69 \\
FLUX.2-dev & $\approx$2.3 & 3.0 & 0.7 & 2.0 & 0.5 & default & 0.7 & 0.7 \\
Qwen-Image & 0.69 & 2.5 & 0.50 & 2.0 & 0.50 & 2.0 & 0.50 & 0.50 \\
SD3.5-Large & 3 & 5 & 0.70 & 6 & 0.70 & 6 & 0.85 & 0.70 \\
\bottomrule
\end{tabular}

\end{table*}
The calibrated schedule tests whether the gain of \method is explained by the two adjustments that a practitioner would make to a full-resolution schedule at a given budget. The default sampler uses each pipeline's resolution-dependent time shift. The calibrated schedule keeps this form and sets the time shift and the first-step shrink per budget on the validation prompts (\cref{tab:calib}). It tunes these two coefficients rather than searching over noise levels. Every coefficient was set on the validation prompts before the test sets were generated. On SD3.5-Large at 4 NFEs, SE + shrink is within 0.003 \lpips of the calibrated schedule. From 5 NFEs on, no time shift beats SE + shrink, which is then the full-resolution baseline.

SharpEuler's released code \cite{sharpeuler2026} keeps the first level and the last two interior levels of the default grid and places the levels between them by the model's measured acceleration profile. Its 2- and 3-step grids therefore equal the default ones. The first-step shrink multiplies the first clean estimate inside the scheduler by a constant and needs no extra evaluation. It is a one-coefficient correction in the spirit of CFG-Zero* \cite{cfgzero2025}, which damps the first steps of guided flow sampling. We add it to SharpEuler on every model, so that this baseline has the correction that \method uses in its first step on Qwen-Image and SD3.5-Large. At 2 and 3 steps SE + shrink is therefore the default grid with the shrink. \Cref{app:shrink} gives the results of \method without it.

A second-order multistep solver does not change these comparisons. DPM-Solver++(2M) \cite{dpmsolverpp2022} on the default and calibrated grids coincides with Euler at 2 and 3 NFEs within 0.001 \lpips. At 4 NFEs it changes \lpips by $-0.013$ to $+0.010$ on 30 validation prompts.

\subsection{Multi-resolution samplers}
\label{app:multires}
\begin{table}[t]
\centering\footnotesize
\caption{\textbf{NFE counts of the multi-resolution samplers.} MrFlow also runs a super-resolution network and a VAE re-encoding. RALU's 3 resolution stages have no NFE count, and \cref{tab:mainfull} gives its seconds.}
\label{tab:nfe}
\setlength{\tabcolsep}{4pt}
\begin{tabular}{@{}l l c c@{}}
\toprule
 & & \multicolumn{2}{c}{NFEs} \\
\cmidrule(l){3-4}
sampler & evaluations & FLUX.1-dev & Qwen-Image \\
\midrule
SPEED-4 & mixed $512^2$ and $1024^2$ & 2.64 & -- \\
SPEED-6 & mixed $512^2$ and $1024^2$ & 3.95 & -- \\
SPEED-8 & mixed $512^2$ and $1024^2$ & 5.95 & -- \\
MrFlow 4+1 & $4\times512^2+1\times1024^2$ & 2.27 & 2.35 \\
MrFlow 6+1 & $6\times512^2+1\times1024^2$ & 2.91 & 3.02 \\
MrFlow 12+1 & $12\times512^2+1\times1024^2$ & 4.82 & 5.04 \\
MrFlow 20+1 & $20\times512^2+1\times1024^2$ & 7.36 & 7.74 \\
RALU & 3 resolution stages & -- & -- \\
\bottomrule
\end{tabular}

\end{table}
\begin{table}[t]
\centering\footnotesize
\caption{\textbf{Where and how SPEED switches.} FLUX.1-dev, 4 NFEs, 1000 test prompts, \lpips to the reference and \ds in brackets. Rows are SPEED's own schedule, which switches at $\sigma\approx0.76$, or the switch point and step placement of \method. Columns are SPEED's spectral expansion, which keeps the evolved state and fills the new bands with the seed's high-frequency noise block, or the switch of \method, which rebuilds the state from the clean estimate and the seed's full noise.}
\label{tab:speed2x2}
\setlength{\tabcolsep}{4pt}
\begin{tabular}{@{}l c c@{}}
\toprule
schedule & SPEED's expansion & \method's switch \\
\midrule
SPEED (switch at $\sigma\approx0.76$) & 0.521 (0.225) & 0.508 (0.213) \\
\method ($\sh=0.9$, its steps) & 0.476 (0.182) & \textbf{0.457} (\textbf{0.166}) \\
\bottomrule
\end{tabular}
\end{table}
We run the official code of MrFlow, SPEED and RALU from noise tied to the reference, which brings their samples closer to it. MrFlow's official code \cite{mrflow2026} generates at $512^2$ for $K_L$ steps, upsamples the decoded image with Real-ESRGAN \cite{realesrgan2021}, re-encodes it and refines it with a single $1024^2$ step from $\sigma=0.12$. We run $K_L=4,6,12$ and 20, also on Qwen-Image with its official script. SPEED's official training-free FLUX implementation \cite{speed2026} runs its spectral schedule with 4, 6 and 8 steps. RALU's official code \cite{ralu2026} runs 2, 2 and 3 steps in its 3 stages at its default level 7 from the same noise. \Cref{tab:nfe} gives their NFE counts.

\paragraph{Noise coupling} MrFlow starts from the seed's coarse noise $D\epsilon$, and SPEED takes the low DCT block of $\epsilon$ in its first stage and the high block in its expansion. This coupling lowers their \lpips by 0.04--0.17 without changing their ratings. With their own noise they produce a different image, 0.59--0.69 \lpips from the reference, no closer than a sample from another seed.

\paragraph{Where and how SPEED switches} SPEED's implementation also lets us isolate the two choices that our analysis makes (\cref{tab:speed2x2}). The switch point and step placement are worth 0.045--0.051 \lpips at 4 NFEs, and what to switch with is worth 0.013--0.019. The two are nearly additive and together give the gap of 0.064 between SPEED's 6 steps and \method. Even with the switch point and step placement of \method, SPEED's spectral expansion stays behind \method on \lpips, \ds, \ir and \ps, by 0.019 \lpips at 4 NFEs and 0.038 at 3.

\subsection{CZP}
\label{app:czp}
Commutator-zero previews (CZP) \cite{lowrespreview2026} have no released code, so we reimplement the algorithm from the paper. CZP follows the default schedule of $N$ steps and evaluates the model at $1024^2$ for its first steps and once more at the switch. It then subsamples the state by 2 with whichever of 4 subsampling patterns minimizes the commutator of subsampling and the model's velocity, using 4 evaluations at $512^2$. It applies $m$ correction steps towards a zero commutator and finishes the remaining steps at $512^2$. The published setting is $N=30$ with 10 steps at $1024^2$ and $m=5$. Its correction strength is 0.04 on FLUX.1-dev and 0.01 on SD3.5-Large. We also run $N=8$ and 12 with 3 and 4 steps at $1024^2$. These and a setting with $N=30$ run without the correction, as in the authors' ablation, because at $N=8$ and 12 the correction does not help.

The CZP paper defines the velocity as pointing towards the data, while the Euler step of its Algorithm 1 uses the network output, which points towards the noise. We take the sign of the correction from the definition, with which the correction lowers the commutator and brings the published setting closer to the reference on both models. The authors measure consistency against a 30-step sample. For the first 200 prompts on FLUX.1-dev, the published setting is 0.070 \ds from that sample, and the authors report 0.068.

A $1024^2$ evaluation counts as one NFE and a $512^2$ evaluation at its measured cost, 0.32 on FLUX.1-dev and 0.29 on SD3.5-Large. The published setting then costs 20.2 NFEs on FLUX.1-dev and 19.3 on SD3.5-Large. This agrees with the authors' report of 65\% of the FLOPs of 30 full-resolution steps. The previews are $512^2$ images. As the authors do, we compare them with the reference downsampled to $512^2$, the resolution at which our \lpips is computed, and \ds, DINOv2 and the two rating models resize their input. References, prompts and seeds are those of the other baselines.

\subsection{Caching accelerators and trained preview solvers}
\label{app:cachesetup}
\paragraph{Caching accelerators} We run TaylorSeer \cite{taylorseer2025} at official commit 704ee98 and DPCache \cite{dpcache2026} at official commit 5d9dcad, with DPCache's released FLUX cost tensor, over the default 50-step schedule from the same noise, embeddings and reference as every other sampler. TaylorSeer computes the first 3 steps in full and refreshes at a fixed step interval with a second-order Taylor forecast, the setting of the DPCache paper, which is 0.013 \lpips better than the released first-order default at an interval of 5. DPCache computes $K$ key steps chosen by its dynamic program, always including the first 3, and forecasts the rest to second order. With every step computed, both reproduce the reference bit-exactly, which verifies our integration. Their 50 outer steps cost 0.49\,s per image for DPCache and 0.97\,s for TaylorSeer beyond the full evaluations, and the seconds we report include them.

\paragraph{Trained preview solvers} ConsistencySolver \cite{consolver2026} and its adaptation to FLUX.1-dev \cite{jia2026faithful} train per model for the same preview goal. The released ConsistencySolver weights are for SD1.5, and the adaptation trains it for 4-step FLUX.1-dev with reinforcement learning against 28-step references at $512^2$. Both differ from our protocol in resolution and reference, so their numbers are not comparable with ours.

\FloatBarrier
\section{Results on the 4 models}
\label{app:results}
\noindent In every speed-up group from $11\times$ to $25\times$, on all 4 models, \method has the lowest \lpips and \fd, the highest DINO and the highest \ir (\cref{app:full}). Fidelity does not cost text alignment (\cref{app:geneval}), the comparisons are robust (\cref{app:robust}), and refining a preview keeps the image (\cref{app:consistency}). On FLUX.1-dev, \method keeps the seed's layout (\cref{app:layout}), participants judge its samples closer to the reference and better overall (\cref{app:userstudy}), and at 2 and 3 NFEs its previews pick better seeds (\cref{app:select}).

\subsection{All metrics and paired differences}
\label{app:full}
\begin{table*}[t]
\centering\scriptsize
\caption{\textbf{Test sets, all metrics, by speed-up over the default 50-step sampler.} Means. Seconds per image on one GPU, timed as described in \cref{app:timing}. Groups and baselines as in \cref{tab:main}, with the $8\times$ group of FLUX.1-dev. Bold is the best value of each group and metric, and $^*$ marks \method ahead of every other sampler of its group by a paired 95\% interval excluding zero. \fd uses the set of references.}
\label{tab:mainfull}
{\setlength{\tabcolsep}{3.6pt}\begin{tabular}{@{}c l r r r r r r r r r@{}}
\toprule
& sampler & NFE & s/img & speed-up ($\times$) & LPIPS$\downarrow$ & DreamSim$\downarrow$ & DINO$\uparrow$ & ImageReward$\uparrow$ & PickScore$\uparrow$ & FD-DINO$\downarrow$ \\
\midrule
\multicolumn{11}{@{}l}{\textit{FLUX.1-dev}, default 50 steps at 15.2\,s, \ir 0.94, \ps 21.81} \\
\midrule
$\approx$21$\times$ & default & 2 & 0.71 & 21.4 & 0.769 & 0.635 & 0.203 & -1.90 & 18.40 & 1604 \\
 & calibrated & 2 & 0.71 & 21.4 & 0.697 & 0.514 & 0.330 & -1.36 & 19.17 & 1317 \\
 & \textbf{\method} & \textbf{1.95} & \textbf{0.71} & \textbf{21.5} & \textbf{0.598}$^*$ & \textbf{0.325}$^*$ & \textbf{0.572}$^*$ & \textbf{-0.24}$^*$ & \textbf{19.95}$^*$ & \textbf{727} \\
\midrule
$\approx$15$\times$ & default & 3 & 1.02 & 15.0 & 0.701 & 0.539 & 0.357 & -1.08 & 19.32 & 1117 \\
 & calibrated & 3 & 1.01 & 15.0 & 0.576 & 0.299 & 0.588 & 0.20 & 20.49 & 617 \\
 & SPEED-4 & 2.64 & 0.91 & 16.7 & 0.626 & 0.333 & 0.600 & 0.20 & 20.56 & 587 \\
 & MrFlow 4+1 & 2.27 & 0.95 & 16.0 & 0.622 & 0.375 & 0.525 & 0.10 & 20.43 & 650 \\
 & MrFlow 6+1 & 2.91 & 1.14 & 13.4 & 0.543 & 0.270 & 0.658 & 0.57 & \textbf{21.27} & 424 \\
 & \textbf{\method} & \textbf{2.95} & \textbf{1.01} & \textbf{15.0} & \textbf{0.491}$^*$ & \textbf{0.192}$^*$ & \textbf{0.726}$^*$ & \textbf{0.63}$^*$ & 21.01 & \textbf{388} \\
\midrule
$\approx$12$\times$ & default & 4 & 1.32 & 11.6 & 0.628 & 0.381 & 0.530 & -0.08 & 20.29 & 659 \\
 & SE + shrink & 4 & 1.32 & 11.6 & 0.540 & 0.262 & 0.651 & 0.53 & 20.99 & 479 \\
 & calibrated & 4 & 1.32 & 11.6 & 0.491 & 0.207 & 0.706 & 0.45 & 20.77 & 442 \\
 & SPEED-6 & 3.95 & 1.31 & 11.6 & 0.521 & 0.225 & 0.706 & 0.68 & 21.30 & 386 \\
 & CZP ($512^2$) & 6.86 & 2.14 & 7.1 & 0.559 & 0.253 & 0.678 & 0.59 & 21.09 & 409 \\
 & \textbf{\method} & \textbf{3.95} & \textbf{1.31} & \textbf{11.6} & \textbf{0.457}$^*$ & \textbf{0.166}$^*$ & \textbf{0.764}$^*$ & \textbf{0.82}$^*$ & \textbf{21.44}$^*$ & \textbf{309} \\
\midrule
$\approx$8$\times$ & default & 6 & 1.92 & 7.9 & 0.544 & 0.279 & 0.659 & 0.48 & 21.04 & 436 \\
 & SE + shrink & 6 & 1.92 & 7.9 & 0.403 & 0.147 & 0.788 & \textbf{0.89} & 21.65 & 279 \\
 & MrFlow 12+1 & 4.82 & 1.71 & 8.9 & 0.460 & 0.187 & 0.749 & 0.80 & 21.64 & 293 \\
 & RALU & -- & 1.98 & 7.7 & 0.604 & 0.270 & 0.654 & 0.69 & 21.31 & 427 \\
 & DPCache & 6 & 2.32 & 6.6 & 0.439 & 0.183 & 0.749 & 0.86 & 21.03 & 375 \\
 & TaylorSeer & 6 & 2.80 & 5.4 & 0.620 & 0.342 & 0.606 & 0.40 & 20.35 & 596 \\
 & \textbf{\method} & \textbf{5.95} & \textbf{1.92} & \textbf{7.9} & \textbf{0.389}$^*$ & \textbf{0.137}$^*$ & \textbf{0.800}$^*$ & \textbf{0.89} & \textbf{21.73}$^*$ & \textbf{260} \\
\midrule
\multicolumn{11}{@{}l}{\textit{FLUX.2-dev}, default 50 steps at 51.0\,s, \ir 1.19, \ps 22.09} \\
\midrule
$\approx$24$\times$ & default & 2 & 2.15 & 23.8 & 0.631 & 0.406 & 0.507 & -0.60 & 19.48 & 1161 \\
 & calibrated & 2 & 2.15 & 23.8 & 0.622 & 0.392 & 0.514 & -0.68 & 19.36 & 1166 \\
 & \textbf{\method} & \textbf{1.91} & \textbf{2.07} & \textbf{24.6} & \textbf{0.532}$^*$ & \textbf{0.255}$^*$ & \textbf{0.676}$^*$ & \textbf{0.36}$^*$ & \textbf{20.32}$^*$ & \textbf{819} \\
\midrule
$\approx$16$\times$ & default & 3 & 3.16 & 16.1 & 0.502 & 0.243 & 0.706 & 0.40 & 20.55 & 730 \\
 & calibrated & 3 & 3.16 & 16.1 & 0.481 & 0.234 & 0.718 & 0.43 & 20.60 & 694 \\
 & \textbf{\method} & \textbf{2.91} & \textbf{3.09} & \textbf{16.5} & \textbf{0.427}$^*$ & \textbf{0.172}$^*$ & \textbf{0.780}$^*$ & \textbf{0.91}$^*$ & \textbf{21.17}$^*$ & \textbf{545} \\
\midrule
$\approx$12$\times$ & default & 4 & 4.18 & 12.2 & 0.414 & 0.172 & 0.789 & 0.77 & 21.04 & 538 \\
 & SE + shrink & 4 & 4.18 & 12.2 & 0.382 & 0.161 & 0.795 & 0.77 & 21.01 & 526 \\
 & calibrated & 4 & 4.18 & 12.2 & 0.399 & 0.161 & 0.799 & 0.79 & 21.05 & 518 \\
 & \textbf{\method} & \textbf{3.91} & \textbf{4.11} & \textbf{12.4} & \textbf{0.379} & \textbf{0.144}$^*$ & \textbf{0.821}$^*$ & \textbf{1.06}$^*$ & \textbf{21.56}$^*$ & \textbf{431} \\
\midrule
\multicolumn{11}{@{}l}{\textit{Qwen-Image}, default 50 steps at 29.9\,s, \ir 1.22, \ps 22.00} \\
\midrule
$\approx$22$\times$ & default & 2 & 1.31 & 22.8 & 0.716 & 0.597 & 0.227 & -1.64 & 18.41 & 1817 \\
 & calibrated & 2 & 1.31 & 22.8 & 0.651 & 0.523 & 0.237 & -1.34 & 18.45 & 1827 \\
 & \textbf{\method} & \textbf{2.01} & \textbf{1.35} & \textbf{22.2} & \textbf{0.587}$^*$ & \textbf{0.328}$^*$ & \textbf{0.536}$^*$ & \textbf{0.02}$^*$ & \textbf{19.56}$^*$ & \textbf{1051} \\
\midrule
$\approx$15$\times$ & default & 3 & 1.91 & 15.7 & 0.674 & 0.483 & 0.427 & -0.90 & 19.23 & 1248 \\
 & calibrated & 3 & 1.91 & 15.7 & 0.543 & 0.285 & 0.603 & 0.17 & 20.01 & 907 \\
 & MrFlow 4+1 & 2.35 & 1.66 & 18.0 & 0.643 & 0.449 & 0.467 & -0.56 & 19.51 & 1144 \\
 & MrFlow 6+1 & 3.02 & 2.06 & 14.5 & 0.581 & 0.319 & 0.640 & 0.35 & 20.47 & 781 \\
 & \textbf{\method} & \textbf{3.01} & \textbf{1.94} & \textbf{15.4} & \textbf{0.500}$^*$ & \textbf{0.195}$^*$ & \textbf{0.754}$^*$ & \textbf{0.81}$^*$ & \textbf{20.77}$^*$ & \textbf{574} \\
\midrule
$\approx$12$\times$ & default & 4 & 2.50 & 12.0 & 0.627 & 0.399 & 0.552 & -0.28 & 19.80 & 973 \\
 & SE + shrink & 4 & 2.50 & 12.0 & 0.529 & 0.275 & 0.679 & 0.43 & 20.54 & 726 \\
 & calibrated & 4 & 2.50 & 12.0 & 0.456 & 0.188 & 0.743 & 0.73 & 20.68 & 610 \\
 & \textbf{\method} & \textbf{4.01} & \textbf{2.54} & \textbf{11.8} & \textbf{0.449} & \textbf{0.159}$^*$ & \textbf{0.803}$^*$ & \textbf{0.98}$^*$ & \textbf{21.32}$^*$ & \textbf{457} \\
\midrule
\multicolumn{11}{@{}l}{\textit{SD3.5-Large}, default 50 steps at 14.3\,s, \ir 1.05, \ps 21.95} \\
\midrule
$\approx$20$\times$ & default & 2 & 0.67 & 21.3 & 0.727 & 0.516 & 0.331 & -1.30 & 18.85 & 1494 \\
 & calibrated & 2 & 0.67 & 21.3 & 0.693 & 0.477 & 0.384 & -1.15 & 18.94 & 1363 \\
 & \textbf{\method} & \textbf{2.08} & \textbf{0.70} & \textbf{20.4} & \textbf{0.661}$^*$ & \textbf{0.371}$^*$ & \textbf{0.490}$^*$ & \textbf{-0.66}$^*$ & \textbf{19.26}$^*$ & \textbf{1123} \\
\midrule
$\approx$15$\times$ & default & 3 & 0.95 & 15.0 & 0.647 & 0.384 & 0.527 & -0.32 & 19.90 & 1007 \\
 & calibrated & 3 & 0.95 & 15.0 & 0.607 & 0.329 & 0.586 & -0.16 & 19.96 & 908 \\
 & \textbf{\method} & \textbf{3.08} & \textbf{0.98} & \textbf{14.5} & \textbf{0.576}$^*$ & \textbf{0.270}$^*$ & \textbf{0.675}$^*$ & \textbf{0.20}$^*$ & \textbf{20.46}$^*$ & \textbf{722} \\
\midrule
$\approx$11$\times$ & default & 4 & 1.24 & 11.5 & 0.604 & 0.327 & 0.612 & 0.11 & 20.45 & 826 \\
 & SE + shrink & 4 & 1.24 & 11.5 & 0.556 & 0.281 & 0.676 & 0.29 & 20.69 & 711 \\
 & calibrated & 4 & 1.24 & 11.5 & 0.553 & 0.261 & 0.677 & 0.35 & 20.62 & 715 \\
 & CZP ($512^2$) & 6.57 & 1.89 & 7.6 & 0.663 & 0.273 & 0.666 & 0.49 & 20.76 & 713 \\
 & \textbf{\method} & \textbf{4.08} & \textbf{1.27} & \textbf{11.3} & \textbf{0.543}$^*$ & \textbf{0.238}$^*$ & \textbf{0.716}$^*$ & \textbf{0.51} & \textbf{20.92}$^*$ & \textbf{624} \\
\bottomrule
\end{tabular}
}
\end{table*}
\begin{table*}[t]
\centering\footnotesize
\caption{\textbf{\method minus every other sampler of its group on FLUX.1-dev and FLUX.2-dev.} Paired differences with 95\% prompt-cluster intervals. $^*$ marks those excluding zero. Negative \lpips and \ds and positive DINO, \ir and \ps favor \method.}
\label{tab:diff_a}
{\setlength{\tabcolsep}{2.2pt}\begin{tabular}{@{}l l l l l l@{}}
\toprule
\method minus & LPIPS$\downarrow$ & DreamSim$\downarrow$ & DINO$\uparrow$ & ImageReward$\uparrow$ & PickScore$\uparrow$ \\
\midrule
\multicolumn{6}{@{}l}{\textit{FLUX.1-dev}} \\
$\approx$21$\times$: default & $-0.170^*$ {\scriptsize[$-$0.177, $-$0.164]} & $-0.310^*$ {\scriptsize[$-$0.318, $-$0.302]} & $+0.369^*$ {\scriptsize[+0.356, +0.382]} & $+1.66^*$ {\scriptsize[+1.60, +1.73]} & $+1.55^*$ {\scriptsize[+1.48, +1.61]} \\
\phantom{$\approx$21$\times$: }calibrated & $-0.099^*$ {\scriptsize[$-$0.104, $-$0.094]} & $-0.189^*$ {\scriptsize[$-$0.195, $-$0.182]} & $+0.242^*$ {\scriptsize[+0.232, +0.252]} & $+1.12^*$ {\scriptsize[+1.07, +1.17]} & $+0.78^*$ {\scriptsize[+0.73, +0.83]} \\
$\approx$15$\times$: default & $-0.209^*$ {\scriptsize[$-$0.216, $-$0.202]} & $-0.348^*$ {\scriptsize[$-$0.356, $-$0.339]} & $+0.369^*$ {\scriptsize[+0.356, +0.382]} & $+1.71^*$ {\scriptsize[+1.65, +1.78]} & $+1.68^*$ {\scriptsize[+1.61, +1.75]} \\
\phantom{$\approx$15$\times$: }calibrated & $-0.084^*$ {\scriptsize[$-$0.090, $-$0.079]} & $-0.107^*$ {\scriptsize[$-$0.113, $-$0.101]} & $+0.138^*$ {\scriptsize[+0.129, +0.147]} & $+0.43^*$ {\scriptsize[+0.39, +0.47]} & $+0.51^*$ {\scriptsize[+0.47, +0.56]} \\
\phantom{$\approx$15$\times$: }SPEED-4 & $-0.135^*$ {\scriptsize[$-$0.140, $-$0.129]} & $-0.141^*$ {\scriptsize[$-$0.148, $-$0.134]} & $+0.126^*$ {\scriptsize[+0.117, +0.135]} & $+0.43^*$ {\scriptsize[+0.38, +0.47]} & $+0.44^*$ {\scriptsize[+0.39, +0.49]} \\
\phantom{$\approx$15$\times$: }MrFlow 4+1 & $-0.131^*$ {\scriptsize[$-$0.136, $-$0.125]} & $-0.184^*$ {\scriptsize[$-$0.191, $-$0.176]} & $+0.201^*$ {\scriptsize[+0.190, +0.213]} & $+0.53^*$ {\scriptsize[+0.48, +0.59]} & $+0.58^*$ {\scriptsize[+0.51, +0.64]} \\
\phantom{$\approx$15$\times$: }MrFlow 6+1 & $-0.052^*$ {\scriptsize[$-$0.057, $-$0.047]} & $-0.078^*$ {\scriptsize[$-$0.084, $-$0.072]} & $+0.069^*$ {\scriptsize[+0.060, +0.078]} & $+0.06^*$ {\scriptsize[+0.01, +0.10]} & $-0.26^*$ {\scriptsize[$-$0.31, $-$0.21]} \\
$\approx$12$\times$: default & $-0.171^*$ {\scriptsize[$-$0.178, $-$0.164]} & $-0.215^*$ {\scriptsize[$-$0.223, $-$0.206]} & $+0.234^*$ {\scriptsize[+0.221, +0.246]} & $+0.91^*$ {\scriptsize[+0.84, +0.98]} & $+1.15^*$ {\scriptsize[+1.09, +1.22]} \\
\phantom{$\approx$12$\times$: }SE + shrink & $-0.083^*$ {\scriptsize[$-$0.089, $-$0.078]} & $-0.096^*$ {\scriptsize[$-$0.102, $-$0.090]} & $+0.113^*$ {\scriptsize[+0.104, +0.122]} & $+0.29^*$ {\scriptsize[+0.26, +0.33]} & $+0.45^*$ {\scriptsize[+0.41, +0.50]} \\
\phantom{$\approx$12$\times$: }calibrated & $-0.034^*$ {\scriptsize[$-$0.040, $-$0.029]} & $-0.041^*$ {\scriptsize[$-$0.046, $-$0.036]} & $+0.058^*$ {\scriptsize[+0.050, +0.066]} & $+0.37^*$ {\scriptsize[+0.34, +0.41]} & $+0.67^*$ {\scriptsize[+0.64, +0.71]} \\
\phantom{$\approx$12$\times$: }SPEED-6 & $-0.064^*$ {\scriptsize[$-$0.069, $-$0.059]} & $-0.059^*$ {\scriptsize[$-$0.064, $-$0.054]} & $+0.058^*$ {\scriptsize[+0.051, +0.065]} & $+0.15^*$ {\scriptsize[+0.11, +0.18]} & $+0.15^*$ {\scriptsize[+0.11, +0.18]} \\
\phantom{$\approx$12$\times$: }CZP ($512^2$) & $-0.102^*$ {\scriptsize[$-$0.108, $-$0.096]} & $-0.087^*$ {\scriptsize[$-$0.094, $-$0.080]} & $+0.086^*$ {\scriptsize[+0.076, +0.096]} & $+0.24^*$ {\scriptsize[+0.19, +0.28]} & $+0.35^*$ {\scriptsize[+0.30, +0.41]} \\
$\approx$8$\times$: default & $-0.156^*$ {\scriptsize[$-$0.163, $-$0.148]} & $-0.142^*$ {\scriptsize[$-$0.151, $-$0.133]} & $+0.141^*$ {\scriptsize[+0.129, +0.153]} & $+0.40^*$ {\scriptsize[+0.35, +0.46]} & $+0.69^*$ {\scriptsize[+0.63, +0.75]} \\
\phantom{$\approx$8$\times$: }SE + shrink & $-0.015^*$ {\scriptsize[$-$0.021, $-$0.009]} & $-0.010^*$ {\scriptsize[$-$0.014, $-$0.005]} & $+0.011^*$ {\scriptsize[+0.004, +0.018]} & $0.00$ {\scriptsize[$-$0.03, +0.02]} & $+0.08^*$ {\scriptsize[+0.05, +0.11]} \\
\phantom{$\approx$8$\times$: }MrFlow 12+1 & $-0.071^*$ {\scriptsize[$-$0.076, $-$0.066]} & $-0.050^*$ {\scriptsize[$-$0.054, $-$0.045]} & $+0.051^*$ {\scriptsize[+0.043, +0.058]} & $+0.09^*$ {\scriptsize[+0.06, +0.12]} & $+0.09^*$ {\scriptsize[+0.05, +0.13]} \\
\phantom{$\approx$8$\times$: }RALU & $-0.215^*$ {\scriptsize[$-$0.221, $-$0.209]} & $-0.132^*$ {\scriptsize[$-$0.139, $-$0.126]} & $+0.146^*$ {\scriptsize[+0.136, +0.156]} & $+0.20^*$ {\scriptsize[+0.15, +0.24]} & $+0.41^*$ {\scriptsize[+0.37, +0.46]} \\
\phantom{$\approx$8$\times$: }DPCache & $-0.050^*$ {\scriptsize[$-$0.057, $-$0.042]} & $-0.046^*$ {\scriptsize[$-$0.051, $-$0.041]} & $+0.050^*$ {\scriptsize[+0.042, +0.058]} & $+0.03$ {\scriptsize[$-$0.00, +0.06]} & $+0.70^*$ {\scriptsize[+0.66, +0.75]} \\
\phantom{$\approx$8$\times$: }TaylorSeer & $-0.232^*$ {\scriptsize[$-$0.242, $-$0.221]} & $-0.205^*$ {\scriptsize[$-$0.214, $-$0.195]} & $+0.193^*$ {\scriptsize[+0.181, +0.205]} & $+0.49^*$ {\scriptsize[+0.44, +0.54]} & $+1.38^*$ {\scriptsize[+1.30, +1.45]} \\
\midrule
\multicolumn{6}{@{}l}{\textit{FLUX.2-dev}} \\
$\approx$24$\times$: default & $-0.099^*$ {\scriptsize[$-$0.106, $-$0.092]} & $-0.151^*$ {\scriptsize[$-$0.160, $-$0.142]} & $+0.169^*$ {\scriptsize[+0.155, +0.184]} & $+0.96^*$ {\scriptsize[+0.89, +1.04]} & $+0.84^*$ {\scriptsize[+0.76, +0.92]} \\
\phantom{$\approx$24$\times$: }calibrated & $-0.089^*$ {\scriptsize[$-$0.096, $-$0.083]} & $-0.137^*$ {\scriptsize[$-$0.144, $-$0.130]} & $+0.161^*$ {\scriptsize[+0.149, +0.174]} & $+1.05^*$ {\scriptsize[+0.98, +1.11]} & $+0.96^*$ {\scriptsize[+0.90, +1.03]} \\
$\approx$16$\times$: default & $-0.075^*$ {\scriptsize[$-$0.083, $-$0.068]} & $-0.070^*$ {\scriptsize[$-$0.077, $-$0.064]} & $+0.074^*$ {\scriptsize[+0.064, +0.084]} & $+0.51^*$ {\scriptsize[+0.46, +0.57]} & $+0.61^*$ {\scriptsize[+0.55, +0.67]} \\
\phantom{$\approx$16$\times$: }calibrated & $-0.054^*$ {\scriptsize[$-$0.061, $-$0.047]} & $-0.062^*$ {\scriptsize[$-$0.068, $-$0.055]} & $+0.062^*$ {\scriptsize[+0.053, +0.072]} & $+0.48^*$ {\scriptsize[+0.42, +0.53]} & $+0.57^*$ {\scriptsize[+0.51, +0.63]} \\
$\approx$12$\times$: default & $-0.035^*$ {\scriptsize[$-$0.044, $-$0.026]} & $-0.028^*$ {\scriptsize[$-$0.034, $-$0.023]} & $+0.033^*$ {\scriptsize[+0.024, +0.041]} & $+0.28^*$ {\scriptsize[+0.24, +0.33]} & $+0.52^*$ {\scriptsize[+0.47, +0.57]} \\
\phantom{$\approx$12$\times$: }SE + shrink & $-0.003$ {\scriptsize[$-$0.012, +0.006]} & $-0.017^*$ {\scriptsize[$-$0.022, $-$0.012]} & $+0.026^*$ {\scriptsize[+0.017, +0.035]} & $+0.29^*$ {\scriptsize[+0.25, +0.34]} & $+0.55^*$ {\scriptsize[+0.49, +0.60]} \\
\phantom{$\approx$12$\times$: }calibrated & $-0.020^*$ {\scriptsize[$-$0.028, $-$0.012]} & $-0.018^*$ {\scriptsize[$-$0.023, $-$0.012]} & $+0.022^*$ {\scriptsize[+0.013, +0.031]} & $+0.27^*$ {\scriptsize[+0.23, +0.31]} & $+0.51^*$ {\scriptsize[+0.46, +0.56]} \\
\bottomrule
\end{tabular}
}
\end{table*}
\begin{table*}[t]
\centering\footnotesize
\caption{\textbf{\method minus every other sampler of its group on Qwen-Image and SD3.5-Large.} Paired differences with 95\% prompt-cluster intervals. $^*$ marks those excluding zero. Negative \lpips and \ds and positive DINO, \ir and \ps favor \method.}
\label{tab:diff_b}
{\setlength{\tabcolsep}{2.2pt}\begin{tabular}{@{}l l l l l l@{}}
\toprule
\method minus & LPIPS$\downarrow$ & DreamSim$\downarrow$ & DINO$\uparrow$ & ImageReward$\uparrow$ & PickScore$\uparrow$ \\
\midrule
\multicolumn{6}{@{}l}{\textit{Qwen-Image}} \\
$\approx$22$\times$: default & $-0.129^*$ {\scriptsize[$-$0.137, $-$0.120]} & $-0.269^*$ {\scriptsize[$-$0.281, $-$0.257]} & $+0.309^*$ {\scriptsize[+0.290, +0.328]} & $+1.65^*$ {\scriptsize[+1.56, +1.75]} & $+1.14^*$ {\scriptsize[+1.05, +1.24]} \\
\phantom{$\approx$22$\times$: }calibrated & $-0.064^*$ {\scriptsize[$-$0.071, $-$0.057]} & $-0.195^*$ {\scriptsize[$-$0.205, $-$0.185]} & $+0.299^*$ {\scriptsize[+0.282, +0.315]} & $+1.36^*$ {\scriptsize[+1.27, +1.44]} & $+1.11^*$ {\scriptsize[+1.03, +1.19]} \\
$\approx$15$\times$: default & $-0.174^*$ {\scriptsize[$-$0.184, $-$0.164]} & $-0.287^*$ {\scriptsize[$-$0.299, $-$0.275]} & $+0.327^*$ {\scriptsize[+0.309, +0.346]} & $+1.71^*$ {\scriptsize[+1.62, +1.80]} & $+1.53^*$ {\scriptsize[+1.43, +1.63]} \\
\phantom{$\approx$15$\times$: }calibrated & $-0.043^*$ {\scriptsize[$-$0.051, $-$0.035]} & $-0.090^*$ {\scriptsize[$-$0.099, $-$0.081]} & $+0.150^*$ {\scriptsize[+0.136, +0.164]} & $+0.64^*$ {\scriptsize[+0.57, +0.70]} & $+0.76^*$ {\scriptsize[+0.69, +0.83]} \\
\phantom{$\approx$15$\times$: }MrFlow 4+1 & $-0.143^*$ {\scriptsize[$-$0.153, $-$0.134]} & $-0.253^*$ {\scriptsize[$-$0.265, $-$0.242]} & $+0.287^*$ {\scriptsize[+0.269, +0.304]} & $+1.37^*$ {\scriptsize[+1.27, +1.46]} & $+1.26^*$ {\scriptsize[+1.17, +1.36]} \\
\phantom{$\approx$15$\times$: }MrFlow 6+1 & $-0.081^*$ {\scriptsize[$-$0.090, $-$0.073]} & $-0.124^*$ {\scriptsize[$-$0.135, $-$0.113]} & $+0.113^*$ {\scriptsize[+0.098, +0.129]} & $+0.46^*$ {\scriptsize[+0.38, +0.54]} & $+0.30^*$ {\scriptsize[+0.22, +0.39]} \\
$\approx$12$\times$: default & $-0.178^*$ {\scriptsize[$-$0.189, $-$0.167]} & $-0.239^*$ {\scriptsize[$-$0.251, $-$0.228]} & $+0.251^*$ {\scriptsize[+0.234, +0.269]} & $+1.26^*$ {\scriptsize[+1.17, +1.34]} & $+1.52^*$ {\scriptsize[+1.42, +1.61]} \\
\phantom{$\approx$12$\times$: }SE + shrink & $-0.080^*$ {\scriptsize[$-$0.089, $-$0.072]} & $-0.116^*$ {\scriptsize[$-$0.125, $-$0.107]} & $+0.124^*$ {\scriptsize[+0.111, +0.138]} & $+0.55^*$ {\scriptsize[+0.49, +0.61]} & $+0.78^*$ {\scriptsize[+0.70, +0.86]} \\
\phantom{$\approx$12$\times$: }calibrated & $-0.007$ {\scriptsize[$-$0.016, +0.001]} & $-0.029^*$ {\scriptsize[$-$0.036, $-$0.022]} & $+0.060^*$ {\scriptsize[+0.050, +0.071]} & $+0.25^*$ {\scriptsize[+0.20, +0.29]} & $+0.64^*$ {\scriptsize[+0.58, +0.70]} \\
\midrule
\multicolumn{6}{@{}l}{\textit{SD3.5-Large}} \\
$\approx$20$\times$: default & $-0.066^*$ {\scriptsize[$-$0.072, $-$0.059]} & $-0.146^*$ {\scriptsize[$-$0.156, $-$0.135]} & $+0.159^*$ {\scriptsize[+0.143, +0.175]} & $+0.64^*$ {\scriptsize[+0.56, +0.71]} & $+0.40^*$ {\scriptsize[+0.33, +0.48]} \\
\phantom{$\approx$20$\times$: }calibrated & $-0.032^*$ {\scriptsize[$-$0.039, $-$0.025]} & $-0.106^*$ {\scriptsize[$-$0.115, $-$0.096]} & $+0.106^*$ {\scriptsize[+0.092, +0.121]} & $+0.48^*$ {\scriptsize[+0.41, +0.56]} & $+0.31^*$ {\scriptsize[+0.24, +0.38]} \\
$\approx$15$\times$: default & $-0.071^*$ {\scriptsize[$-$0.079, $-$0.064]} & $-0.114^*$ {\scriptsize[$-$0.124, $-$0.105]} & $+0.148^*$ {\scriptsize[+0.132, +0.164]} & $+0.52^*$ {\scriptsize[+0.45, +0.60]} & $+0.56^*$ {\scriptsize[+0.48, +0.64]} \\
\phantom{$\approx$15$\times$: }calibrated & $-0.031^*$ {\scriptsize[$-$0.037, $-$0.025]} & $-0.059^*$ {\scriptsize[$-$0.067, $-$0.051]} & $+0.089^*$ {\scriptsize[+0.077, +0.101]} & $+0.36^*$ {\scriptsize[+0.30, +0.43]} & $+0.49^*$ {\scriptsize[+0.43, +0.56]} \\
$\approx$11$\times$: default & $-0.061^*$ {\scriptsize[$-$0.069, $-$0.053]} & $-0.089^*$ {\scriptsize[$-$0.098, $-$0.080]} & $+0.104^*$ {\scriptsize[+0.090, +0.119]} & $+0.40^*$ {\scriptsize[+0.33, +0.47]} & $+0.47^*$ {\scriptsize[+0.39, +0.55]} \\
\phantom{$\approx$11$\times$: }SE + shrink & $-0.013^*$ {\scriptsize[$-$0.020, $-$0.006]} & $-0.043^*$ {\scriptsize[$-$0.050, $-$0.036]} & $+0.040^*$ {\scriptsize[+0.030, +0.050]} & $+0.22^*$ {\scriptsize[+0.17, +0.27]} & $+0.23^*$ {\scriptsize[+0.17, +0.29]} \\
\phantom{$\approx$11$\times$: }calibrated & $-0.010^*$ {\scriptsize[$-$0.017, $-$0.004]} & $-0.023^*$ {\scriptsize[$-$0.030, $-$0.015]} & $+0.039^*$ {\scriptsize[+0.029, +0.050]} & $+0.16^*$ {\scriptsize[+0.11, +0.21]} & $+0.29^*$ {\scriptsize[+0.24, +0.35]} \\
\phantom{$\approx$11$\times$: }CZP ($512^2$) & $-0.120^*$ {\scriptsize[$-$0.129, $-$0.111]} & $-0.035^*$ {\scriptsize[$-$0.043, $-$0.027]} & $+0.050^*$ {\scriptsize[+0.037, +0.063]} & $+0.01$ {\scriptsize[$-$0.05, +0.08]} & $+0.16^*$ {\scriptsize[+0.09, +0.23]} \\
\bottomrule
\end{tabular}
}
\end{table*}
Against the default sampler, every paired interval favors \method on every metric, model and budget (\cref{tab:mainfull,tab:diff_a,tab:diff_b}). Against the calibrated schedule, every interval favors \method at 2 and 3 NFEs. At 4 NFEs every interval does so except \lpips on Qwen-Image, where the two are on par. Against SE + shrink at 4 NFEs, every interval favors \method except \lpips on FLUX.2-dev, where the two are on par. Against the multi-resolution samplers, every \lpips, \ds and DINO interval favors \method. On Qwen-Image, MrFlow with 12 low-resolution steps is 0.010 \lpips farther from the reference than \method at 4 NFEs, while taking 3.24\,s against 2.54\,s. In the $8\times$ group of FLUX.1-dev, \method is on par in \ir with SE + shrink and DPCache and ahead on every other metric.

Per prompt, \method is closer to the reference than the calibrated schedule on 92\%, 87\% and 69\% of the 1000 FLUX.1-dev test prompts at 2, 3 and 4 NFEs. By MJHQ category, it leads in all 10 at 2 and 3 NFEs and in 9 at 4 NFEs. Against SE + shrink, which at 2 and 3 NFEs is the default grid with the shrink (\cref{app:schedules}), the shares are 94\%, 97\% and 85\%, and \method leads in all 10 categories at every budget. On Qwen-Image and SD3.5-Large, \method is closer than the calibrated schedule on 57--80\% of the prompts. At equal \lpips, interpolated in measured seconds along the full-resolution baseline of \cref{fig:frontier}, \method is 1.03--1.35$\times$ faster on FLUX.1-dev, Qwen-Image and SD3.5-Large at 2 to 4 NFEs.

\subsection{Text alignment}
\label{app:geneval}
\begin{table}[t]
\centering\footnotesize
\caption{\textbf{GenEval}, official protocol, 553 prompts with 4 seeds each, and the first seed for the default 50-step sampler. The full-resolution schedule is the calibrated one, or SE + shrink where marked SE. Paired 95\% intervals.}
\label{tab:geneval}
{\setlength{\tabcolsep}{2.6pt}\begin{tabular}{@{}l c c c c@{}}
\toprule
model & NFE & full-res.\ schedule & \textbf{\method} & $\Delta$ (95\% CI) \\
\midrule
FLUX.1-dev & 2 & 0.253 & \textbf{0.507} & $+0.254^*$ {\scriptsize[+0.229, +0.279]} \\
 & 3 & 0.459 & \textbf{0.615} & $+0.156^*$ {\scriptsize[+0.134, +0.178]} \\
 & 4 & 0.609 & \textbf{0.642} & $+0.033^*$ {\scriptsize[+0.014, +0.051]} \\
 & 50 & \multicolumn{3}{l}{default sampler: 0.656} \\
FLUX.2-dev & 2 & 0.451 & \textbf{0.744} & $+0.293^*$ {\scriptsize[+0.267, +0.320]} \\
 & 3 & 0.690 & \textbf{0.832} & $+0.142^*$ {\scriptsize[+0.121, +0.164]} \\
 & 4 & 0.800 & \textbf{0.848} & $+0.049^*$ {\scriptsize[+0.032, +0.064]} \\
 & 50 & \multicolumn{3}{l}{default sampler: 0.840} \\
Qwen-Image & 2 & 0.339 & \textbf{0.686} & $+0.347^*$ {\scriptsize[+0.319, +0.375]} \\
 & 3 & 0.752 & \textbf{0.829} & $+0.076^*$ {\scriptsize[+0.057, +0.096]} \\
 & 4 & 0.830 & \textbf{0.845} & $+0.015$ {\scriptsize[$-$0.002, +0.032]} \\
 & 50 & \multicolumn{3}{l}{default sampler: 0.863} \\
SD3.5-Large & 2 & 0.306 & \textbf{0.368} & $+0.062^*$ {\scriptsize[+0.041, +0.083]} \\
 & 3 & 0.535 & \textbf{0.594} & $+0.058^*$ {\scriptsize[+0.038, +0.079]} \\
 & 4 & 0.612$^{\mathrm{SE}}$ & \textbf{0.660} & $+0.048^*$ {\scriptsize[+0.029, +0.068]} \\
 & 50 & \multicolumn{3}{l}{default sampler: 0.697} \\
\bottomrule
\end{tabular}
}
\end{table}
\method scores higher on GenEval than the full-resolution schedule of the same budget at 2 and 3 NFEs on all 4 models, by 0.06--0.35, and at 4 NFEs on FLUX.1-dev, FLUX.2-dev and SD3.5-Large (\cref{tab:geneval}). On FLUX.2-dev its 4-NFE samples score 0.060 above SE + shrink. On FLUX.1-dev, \method at 4 NFEs scores 0.047 above SPEED with 6 steps and 0.211 above the default sampler with 4 steps, and at 3 NFEs it scores 0.148 above SPEED with 4 steps.

At 4 NFEs \method is on par with the default 50-step sampler on FLUX.1-dev, FLUX.2-dev and Qwen-Image while running 11--12$\times$ faster, and on SD3.5-Large it is within 0.045 of it. On FLUX.2-dev its 3-NFE samples are on par with the 50-step sampler as well.

\subsection{The seed's layout is kept}
\label{app:layout}
\begin{table}[t]
\centering\footnotesize
\caption{\textbf{Layout without blur.} FLUX.1-dev, 1000 test prompts, same noise as the reference. \lpips on both images downsampled to $128^2$, and the mean DINOv2 patch-token cosine at corresponding positions. The anchors below use the first 200 test prompts, and another seed is a second seed's 50-step sample with its 95\% interval. The blurred references use a Gaussian blur with a standard deviation of 8 or 16 pixels at $1024^2$. Best per budget in bold. All paired differences of \method exclude zero.}
\label{tab:blur}
{\setlength{\tabcolsep}{3pt}\begin{tabular}{@{}l c c c c c c@{}}
\toprule
& \multicolumn{3}{c}{\lpips at $128^2$$\downarrow$} & \multicolumn{3}{c}{DINOv2 patch tokens$\uparrow$} \\
\cmidrule(lr){2-4}\cmidrule(l){5-7}
NFE & 2 & 3 & 4 & 2 & 3 & 4 \\
\midrule
default & 0.658 & 0.554 & 0.464 & 0.244 & 0.336 & 0.430 \\
SE + shrink & -- & -- & 0.341 & -- & -- & 0.521 \\
calibrated & 0.509 & 0.356 & 0.285 & 0.339 & 0.485 & 0.580 \\
\method & \textbf{0.376} & \textbf{0.284} & \textbf{0.262} & \textbf{0.479} & \textbf{0.578} & \textbf{0.601} \\
\midrule
\multicolumn{7}{@{}l}{\textit{anchors}} \\
same noise, 100 steps & \multicolumn{3}{c}{0.104} & \multicolumn{3}{c}{0.80} \\
reference, 8-pixel blur & \multicolumn{3}{c}{0.196} & \multicolumn{3}{c}{0.86} \\
reference, 16-pixel blur & \multicolumn{3}{c}{0.427} & \multicolumn{3}{c}{0.67} \\
another seed & \multicolumn{3}{c}{0.398 [0.386, 0.410]} & \multicolumn{3}{c}{0.445 [0.430, 0.461]} \\
\bottomrule
\end{tabular}
}
\end{table}
\begin{table}[t]
\centering\footnotesize
\caption{\textbf{Object layout on the 553 GenEval prompts, FLUX.1-dev.} Agreement of the boxes that the GenEval detector finds in a sample with those in its 50-step reference, as the Jaccard index of the two sets weighted by IoU, where identical boxes give 1. The middle rows give the seed-specific agreement, the agreement minus the mean agreement of samples from 3 other seeds of the same prompt. The default sampler did not run these seeds. Another seed compares the 50-step image of a second seed with the reference, with its 95\% interval, and the last row is a 28-step sample of the default sampler from the reference's noise. Bold is the highest value at each budget, and $^*$ marks a paired interval of \method against the calibrated schedule that excludes zero.}
\label{tab:boxes}
{\setlength{\tabcolsep}{6pt}\begin{tabular}{@{}l c c c@{}}
\toprule
NFE & 2 & 3 & 4 \\
\midrule
\multicolumn{4}{@{}l}{\textit{agreement with the reference}} \\
default & 0.111 & 0.163 & 0.286 \\
calibrated & 0.188 & 0.315 & 0.445 \\
\method & \textbf{0.370}$^*$ & \textbf{0.441}$^*$ & \textbf{0.447} \\
\midrule
\multicolumn{4}{@{}l}{\textit{seed-specific agreement}} \\
calibrated & 0.032 & 0.062 & \textbf{0.153} \\
\method & \textbf{0.104}$^*$ & \textbf{0.140}$^*$ & 0.140 \\
\midrule
another seed, 50 steps & \multicolumn{3}{c}{0.293 [0.273, 0.314]} \\
default, 28 steps & \multicolumn{3}{c}{0.621} \\
\bottomrule
\end{tabular}
}
\end{table}
Few-step samples are blurred, so we also compare layouts directly. At 2 NFEs, the sample of \method is closer to its reference than another seed's 50-step image is, on both layout measures of \cref{tab:blur}. The two measures are \lpips on both images downsampled to $128^2$ and the mean cosine of DINOv2 patch tokens at corresponding positions of a $16\times16$ grid at $224^2$. The 2-step samples of the calibrated schedule and the default sampler are farther from the reference than the reference under a 16-pixel Gaussian blur, so they differ in layout and not only in sharpness.

The objects that a detector finds give the same order on the 553 GenEval prompts (\cref{tab:boxes}). We pair the boxes of a sample with those of its reference class by class with the Hungarian algorithm and score them by the IoU-weighted Jaccard index, which is 1 for identical boxes. Subtracting the mean agreement of the reference with samples from 3 other seeds of the same prompt gives the seed-specific agreement, which removes what the prompt alone determines. At 2 and 3 NFEs \method keeps more of the reference's layout than the calibrated schedule and has a higher seed-specific agreement. Even its 2-NFE samples agree with their references more than the 3-NFE samples of the calibrated schedule do, by 0.054. They also agree with their references more than two 50-step images of the same prompt from different seeds agree with each other. In each of the 6 GenEval tasks, the seed-specific agreement of \method is above zero at every budget.

\subsection{Robustness}
\label{app:robust}
The comparisons hold with a 100-step reference, at other aspect ratios and on a second seed.

\paragraph{100-step reference} On the first 200 test prompts of each model, we replace the 50-step reference by the 100-step sample from the same noise. This moves the 2- to 4-NFE paired differences of \method to the other samplers of \cref{tab:main} by at most 0.013 \lpips, 0.008 \ds and 0.013 DINO. Every difference keeps its sign, and every lead of \method keeps its significance. On FLUX.2-dev at 4 NFEs the \lpips difference to SE + shrink, on par with the 50-step reference, becomes a lead of \method.

\paragraph{Aspect ratio} On FLUX.1-dev at $1344\times768$ and $768\times1344$ with a half-resolution stage, \method at 4 NFEs is 0.136 and 0.137 \lpips closer to the reference than SharpEuler. At 6 NFEs it is 0.055 and 0.034 closer.

\paragraph{Second seed} On a second validation seed, every 2- to 4-NFE \lpips and \ds comparison with SE + shrink replicates in sign and significance, and the \ir differences replicate in sign.

\subsection{Refining a preview keeps the image}
\label{app:consistency}
\begin{table}[t]
\centering\footnotesize
\caption{\textbf{How much a seed's sample changes from 3 to 4 NFEs.} Same-seed \lpips between the 3- and 4-NFE samples on the test prompts. \method uses the configurations of \cref{sec:method} with the seed's noise at the switch, so its 2 budgets share the low-resolution stage and the switch state and differ only in the full-resolution steps.}
\label{tab:consistency}
{\setlength{\tabcolsep}{6pt}\begin{tabular}{@{}l c c c@{}}
\toprule
model & \method & calibrated & default \\
\midrule
FLUX.1-dev & \textbf{0.106} & 0.288 & 0.298 \\
FLUX.2-dev & \textbf{0.070} & 0.205 & 0.162 \\
Qwen-Image & \textbf{0.107} & 0.295 & 0.144 \\
SD3.5-Large & \textbf{0.110} & 0.259 & 0.193 \\
\bottomrule
\end{tabular}
}
\end{table}
A larger budget of \method can reuse the low-resolution stage and its clean estimate from a smaller one, so the stage need not be rerun when a preview is refined. \Cref{tab:consistency} measures how much a seed's sample then changes from 3 to 4 NFEs, as the same-seed \lpips between the two samples. With \method the change is 0.07--0.11 on the 4 models, against 0.20--0.30 with the calibrated schedule and 0.14--0.30 with the default sampler.

\subsection{User study on FLUX.1-dev}
\label{app:userstudy}
\begin{figure*}[t]
\centering
\includegraphics[width=\textwidth]{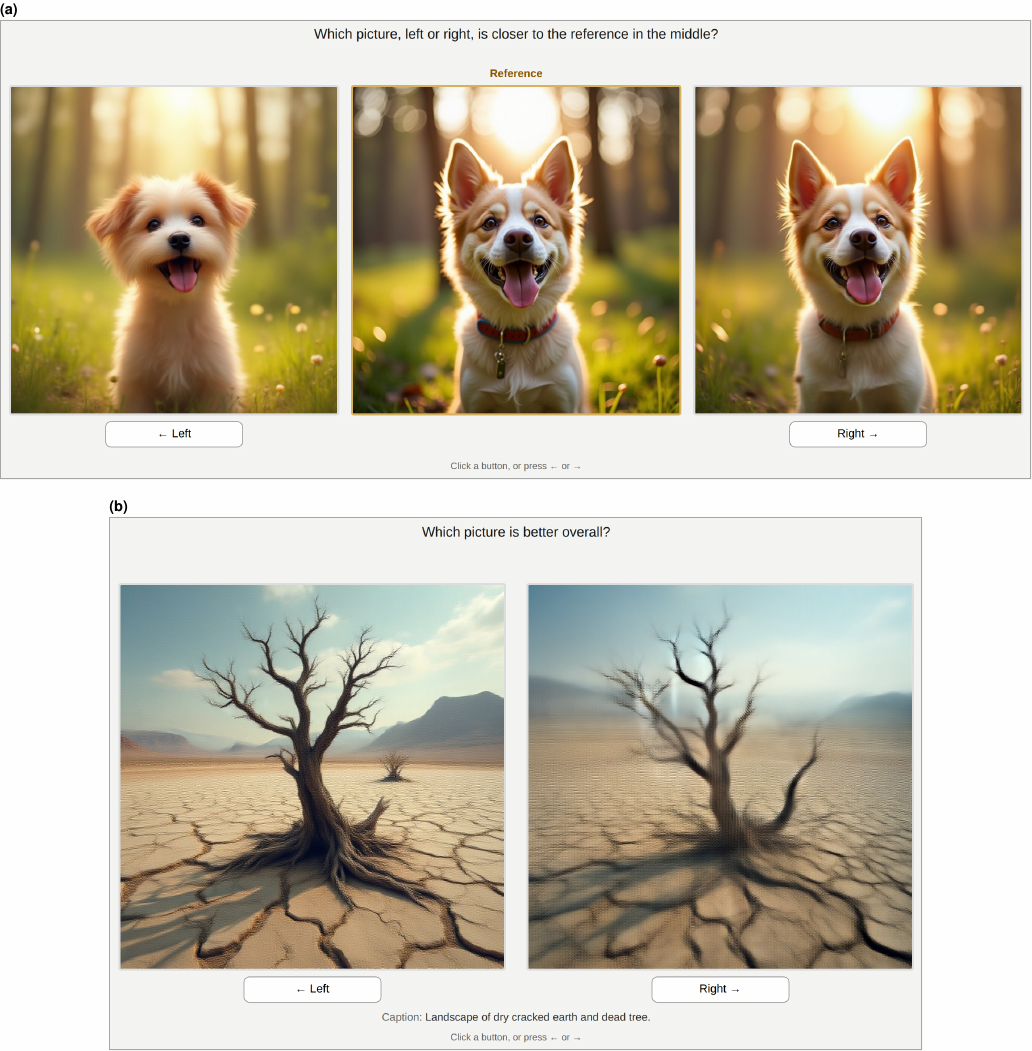}
\caption{\textbf{Sample questions of the two tasks from the user study interface.} (a)~Closeness to the reference. (b)~Overall preference.}
\label{fig:userstudy}
\end{figure*}
\begin{table}[t]
\centering\footnotesize
\caption{\textbf{User study on FLUX.1-dev.} Percentage of trials in which participants chose the sample of \method over that of the other sampler, with 95\% intervals. The last row pools all trials. Chance is 50\%.}
\label{tab:user}
{\setlength{\tabcolsep}{4pt}\begin{tabular}{@{}c l c c@{}}
\toprule
 & other sampler & closer to reference & better overall \\
\midrule
$\approx$21$\times$ & calibrated & 92 {\scriptsize[78, 100]} & 100 {\scriptsize[100, 100]} \\
\midrule
$\approx$15$\times$ & calibrated & 89 {\scriptsize[78, 97]} & 98 {\scriptsize[90, 100]} \\
 & SPEED-4 & 89 {\scriptsize[78, 97]} & 92 {\scriptsize[79, 100]} \\
 & MrFlow 4+1 & 97 {\scriptsize[90, 100]} & 78 {\scriptsize[58, 94]} \\
\midrule
$\approx$12$\times$ & SE + shrink & 90 {\scriptsize[79, 98]} & 91 {\scriptsize[81, 99]} \\
 & calibrated & 76 {\scriptsize[60, 89]} & 97 {\scriptsize[88, 100]} \\
 & SPEED-6 & 79 {\scriptsize[68, 90]} & 68 {\scriptsize[49, 84]} \\
 & CZP ($512^2$) & 83 {\scriptsize[70, 93]} & 71 {\scriptsize[51, 88]} \\
\midrule
\multicolumn{2}{@{}l}{all} & 86 {\scriptsize[81, 91]} & 86 {\scriptsize[80, 91]} \\
\bottomrule
\end{tabular}
}
\par\vspace{-1.5em}
\end{table}

In accordance with our institution's IRB policy, we are conducting a user study that involves only survey procedures. Because responses are recorded without identifiers, the study qualifies for Exempt Category 2 under 45 CFR 46.104(d)(2)(i).

\paragraph{Tasks} The first task shows the 50-step reference between the two samples and asks ``Which picture, left or right, is closer to the reference in the middle?'' (\cref{fig:userstudy}a). Its instructions read ``Judge each picture as a whole: what it shows, where things are, and their shapes and colors. Choose the picture that is closer to the reference, not the one that looks better on its own.'' The second task shows the two samples with the prompt and asks ``Which picture is better overall?'' (\cref{fig:userstudy}b). All participants take the first task first, and each task starts with 3 practice trials that are not counted.

\paragraph{Images} The samples are the FLUX.1-dev test samples of \cref{tab:main}, and CZP's $512^2$ previews are shown at the size of the other images. The prompts are drawn at random from the test set, stratified by MJHQ-30K category, 208 for the first task and 120 for the second. Prompts and images were screened for violent, sexual or otherwise disturbing content.

\paragraph{Participants} The study was completed by 11 participants in a web browser on a desktop or laptop computer. Each answered 104 trials of the first task and 60 of the second, each on a different prompt, and 8 trials with an obvious answer as attention checks. A rule set before data collection excludes participants who fail 2 or more of these checks. It excludes 1 participant and leaves 1040 and 600 trials.

\paragraph{Results} The intervals of \cref{tab:user} are 95\% bootstraps over participants and prompts. At $21\times$ and $15\times$ a row has 120 trials of the first task and 60 of the second, and at $12\times$ it has 140 and 90. With the excluded participant, both pooled shares change by less than 1 point.

\paragraph{Relation to the metrics} In the first task, the share that chose \method rises with the \lpips advantage of \method, from 61\% in the fifth of the trials where this advantage is smallest to 97\% in the fifth where it is largest. Both \lpips and \ds rank the other sample closer in 73 of these trials, and participants chose that sample in 77\% of them.

\subsection{Choosing seeds from previews}
\label{app:select}
\begin{table}[t]
\centering\footnotesize
\caption{\textbf{Choosing among 4 seeds by the few-step samples.} Share in percent of the oracle gain that the pick recovers. The oracle picks by the 50-step images and recovers 100\%, and a random pick recovers 0\%. FLUX.1-dev, 200 prompts with 8 seeds each, and the same score picks and judges. IR and PS are \ir and \ps, and bold is the best.}
\label{tab:select}
{\setlength{\tabcolsep}{8pt}\begin{tabular}{@{}l c c c c@{}}
\toprule
& \multicolumn{2}{c}{IR} & \multicolumn{2}{c@{}}{PS} \\
\cmidrule(lr){2-3}\cmidrule(l){4-5}
NFE & 2 & 3 & 2 & 3 \\
\midrule
default & 0 & 2 & 11 & 4 \\
calibrated & 4 & 20 & 7 & 19 \\
\method & \textbf{19} & \textbf{32} & \textbf{16} & \textbf{31} \\
\bottomrule
\end{tabular}
}
\end{table}
\begin{table}[t]
\centering\footnotesize
\caption{\textbf{Choosing among 4 seeds by the few-step samples of other samplers.} Share in percent of the gain of picking by the 50-step images that a pick by the few-step samples recovers, on FLUX.1-dev as in \cref{tab:select}. NFEs and seconds per image are those of \cref{tab:main} and of this supplement. \mbox{\method + schnell} is \method rendered by FLUX.1-schnell, and Hyper-FLUX runs its 8 steps. \method in bold.}
\label{tab:select_all}
{\setlength{\tabcolsep}{5pt}\begin{tabular}{@{}l r r r r@{}}
\toprule
 & & & \multicolumn{2}{c@{}}{gain recovered (\%)} \\
\cmidrule(l){4-5}
sampler & NFE & sec & IR & PS \\
\midrule
FLUX.1-schnell & 2 & 0.64 & 10 & 16 \\
\textbf{\method + schnell} & \textbf{1.95} & \textbf{0.68} & \textbf{27} & \textbf{23} \\
default & 2 & 0.71 & 0 & 11 \\
calibrated & 2 & 0.71 & 4 & 7 \\
\textbf{\method} & \textbf{1.95} & \textbf{0.71} & \textbf{19} & \textbf{16} \\
SPEED-4 & 2.64 & 0.91 & 20 & 19 \\
default & 3 & 1.02 & 2 & 4 \\
calibrated & 3 & 1.01 & 20 & 19 \\
\textbf{\method} & \textbf{2.95} & \textbf{1.01} & \textbf{32} & \textbf{31} \\
MrFlow 6+1 & 2.91 & 1.14 & 13 & 16 \\
SPEED-6 & 3.95 & 1.31 & 28 & 26 \\
MrFlow 12+1 & 4.82 & 1.71 & 25 & 26 \\
RALU & -- & 1.98 & 0 & 9 \\
CZP, $N{=}8$ & 6.86 & 2.14 & 23 & 26 \\
CZP, $N{=}12$ & 8.82 & 2.73 & 26 & 26 \\
Hyper-FLUX & 8 & 2.78 & 15 & 21 \\
\bottomrule
\end{tabular}
}
\end{table}
At 2 and 3 NFEs, the few-step samples of \method pick better seeds than those of the calibrated schedule and the default sampler (\cref{tab:select}). On 200 other MJHQ-30K prompts with 8 seeds each, we generate the 50-step image of each seed and the few-step samples of each sampler from the same noise, and score all images by \ir and \ps. For every subset of 4 of the 8 seeds, the seed whose few-step sample scores highest is picked. The shortfall of a pick is the score of the best 50-step image among the 4 minus the score of the picked seed's 50-step image, and the same score picks and judges. The share of the gain that the pick recovers is $1-S/S_{\mathrm{rand}}$, where $S$ is the shortfall averaged over all 70 subsets and all prompts and $S_{\mathrm{rand}}$ is the same average for a random pick. \Cref{tab:select_all} adds the multi-resolution and distilled samplers that take up to 3 seconds per image. Every one of them that is slower than \method at 3 NFEs recovers less by both scores. By \ir and \ps, the 2 steps of FLUX.1-schnell recover 10\% and 16\%. Rendered by FLUX.1-schnell, \method recovers 27\% and 23\%.

\FloatBarrier
\section{Larger budgets, caching and previews}
\label{app:budgets}
\noindent Two measured costs explain why the lead of \method over full-resolution schedules is largest at 2 to 4 NFEs (\cref{app:beyond,app:lead}). At these budgets the caching accelerators have no usable operating point (\cref{app:caching}). The previews of CZP that take twice the time of \method are farther from the reference in \lpips (\cref{app:previews}).

\subsection{Beyond 4 NFEs}
\label{app:beyond}
\begin{table}[t]
\centering\footnotesize
\caption{\textbf{Beyond 4 NFEs.} Test-set means and seconds per image on one GPU, for the configurations of \cref{tab:config_more}. Qwen-Image and SD3.5-Large take the coarse switch noise from the stage's own noise estimate. Their seconds are those of the configurations with the seed's noise, which run the same evaluations.}
\label{tab:regime}
{\setlength{\tabcolsep}{2.5pt}\begin{tabular}{@{}l l r r r r r@{}}
\toprule
NFE & sampler & s/img & LPIPS$\downarrow$ & DS$\downarrow$ & IR$\uparrow$ & PS$\uparrow$ \\
\midrule
\multicolumn{7}{@{}l}{\textit{FLUX.1-dev}} \\
5 & SE + shrink & 1.62 & 0.466 & 0.194 & 0.79 & 21.42 \\
 & \method & 1.62 & 0.416 & 0.148 & 0.85 & 21.62 \\
6 & SE + shrink & 1.92 & 0.403 & 0.147 & 0.89 & 21.65 \\
 & \method & 1.92 & 0.389 & 0.137 & 0.89 & 21.73 \\
8 & SE + shrink & 2.53 & 0.319 & 0.102 & 0.93 & 21.77 \\
 & \method & 2.52 & 0.361 & 0.122 & 0.92 & 21.79 \\
\midrule
\multicolumn{7}{@{}l}{\textit{Qwen-Image}} \\
5 & SE + shrink & 3.10 & 0.445 & 0.196 & 0.83 & 21.13 \\
 & \method & 3.13 & 0.429 & 0.144 & 1.05 & 21.50 \\
6 & SE + shrink & 3.70 & 0.387 & 0.148 & 1.01 & 21.44 \\
 & \method & 3.73 & 0.379 & 0.121 & 1.10 & 21.62 \\
8 & SE + shrink & 4.89 & 0.313 & 0.102 & 1.14 & 21.68 \\
 & \method & 4.92 & 0.363 & 0.111 & 1.15 & 21.76 \\
\midrule
\multicolumn{7}{@{}l}{\textit{SD3.5-Large}} \\
5 & SE + shrink & 1.52 & 0.508 & 0.229 & 0.52 & 21.04 \\
 & \method & 1.55 & 0.513 & 0.221 & 0.62 & 21.19 \\
6 & SE + shrink & 1.81 & 0.468 & 0.197 & 0.67 & 21.28 \\
 & \method & 1.86 & 0.458 & 0.181 & 0.64 & 21.27 \\
8 & SE + shrink & 2.37 & 0.401 & 0.153 & 0.80 & 21.48 \\
 & \method & 2.43 & 0.432 & 0.160 & 0.77 & 21.47 \\
\bottomrule
\end{tabular}
}
\end{table}
\begin{figure*}[t]
\centering
\includegraphics[width=\textwidth]{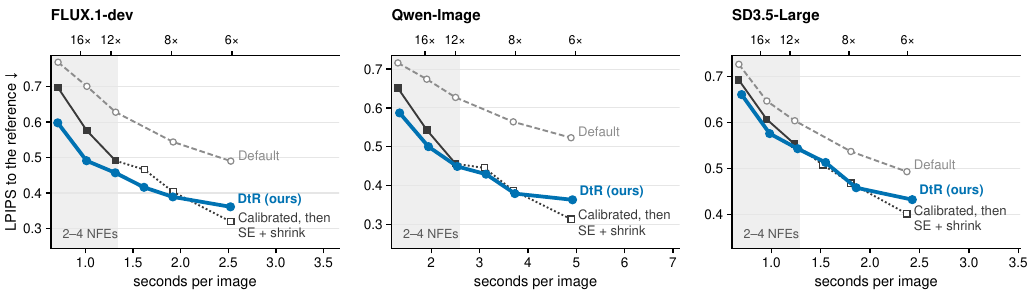}
\caption{\textbf{Fidelity against measured seconds on FLUX.1-dev, Qwen-Image and SD3.5-Large.} Test sets, one timing run per model. The top axis gives the speed-up over the 50-step sampler, and the gray band marks 2 to 4 NFEs. Gray is the default sampler. Dark gray is the calibrated schedule at 2 to 4 NFEs, with filled squares, and SE + shrink from 5 NFEs on, with open squares on a dotted line, because no time shift beats SE + shrink there. Blue is \method.}
\label{fig:frontier}
\end{figure*}
We recommend \method up to 6 NFEs on FLUX.1-dev, Qwen-Image and SD3.5-Large. At 8 NFEs, SE + shrink, which then covers the noise levels above 0.9 with full-resolution steps, is closer to the reference (\cref{tab:regime,fig:frontier}). \Cref{app:lead} explains this change. The configurations follow the rule of \cref{tab:config_more}, and on FLUX.1-dev and Qwen-Image 3 half-resolution steps remain the right choice beyond 4 NFEs. On FLUX.1-dev they are 0.023--0.032 \lpips more faithful at 5 to 8 NFEs than 6-step low-resolution stages of equal cost, and on Qwen-Image 0.043 more faithful at 8 NFEs. With a 100-step reference on 200 FLUX.1-dev test prompts, the lead of \method at 6 NFEs doubles, from 0.018 to 0.037, and at 8 NFEs the two are on par, because the noise levels of SE + shrink fit the 50-step discretization.

\subsection{Why the lead is largest at 2 to 4 NFEs}
\label{app:lead}
\begin{figure*}[t]
\centering
\includegraphics[width=\textwidth]{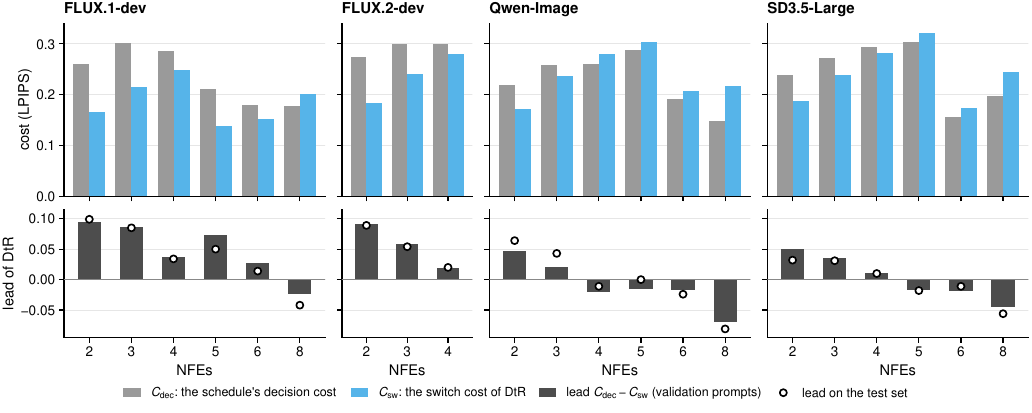}
\caption{\textbf{The lead of \method is the difference of two costs.} The switch takes the seed's noise on every model, as in \cref{tab:budget}. Top, the decision cost $C_{\mathrm{dec}}$ of the full-resolution schedule, the calibrated one up to 4 NFEs and SE + shrink beyond, and the switch cost $C_{\mathrm{sw}}$ of \method. Bottom, the lead $C_{\mathrm{dec}}-C_{\mathrm{sw}}$ on the validation prompts as bars, and the lead measured on the test set as circles. Above zero \method is closer to the reference. FLUX.2-dev was measured at 2 to 4 NFEs. \Cref{tab:budget} gives the numbers.}
\label{fig:lead}
\end{figure*}
\begin{table*}[t]
\centering\footnotesize
\caption{\textbf{The lead decomposed.} \lpips with one seed on at least 30 validation prompts per model, for the configurations of \cref{sec:method} with the seed's noise at the switch, at $\sh=0.90$ up to 4 NFEs, or 5 on Qwen-Image and SD3.5-Large, and at 0.95 beyond. $C_{\mathrm{dec}}$ is the decision cost of the calibrated schedule up to 4 NFEs and of SE + shrink beyond. $C'_{\mathrm{dec}}$ is $C_{\mathrm{dec}}$ with the schedule re-entered at its own noise level nearest $\sh$, which is 0.86--0.95 except 0.83 for SD3.5-Large at 2 NFEs, and every interval of $C'_{\mathrm{dec}}$ excludes zero. $C_{\mathrm{sw}}$ is the switch cost of \method. The lead on the validation prompts is $C_{\mathrm{dec}}-C_{\mathrm{sw}}$, and the lead on the test set is measured directly.}
\label{tab:budget}
{\setlength{\tabcolsep}{2.4pt}\begin{tabular}{@{}l r r r r r r r r r r r r r r r r r r r r @{}}
\toprule
NFE & \multicolumn{5}{c}{FLUX.1-dev} & \multicolumn{5}{c}{FLUX.2-dev} & \multicolumn{5}{c}{Qwen-Image} & \multicolumn{5}{c@{}}{SD3.5-Large} \\
\cmidrule(lr){2-6}\cmidrule(lr){7-11}\cmidrule(lr){12-16}\cmidrule(l){17-21}
 & $C_{\mathrm{dec}}$ & $C'_{\mathrm{dec}}$ & $C_{\mathrm{sw}}$ & \multicolumn{2}{c}{lead} & $C_{\mathrm{dec}}$ & $C'_{\mathrm{dec}}$ & $C_{\mathrm{sw}}$ & \multicolumn{2}{c}{lead} & $C_{\mathrm{dec}}$ & $C'_{\mathrm{dec}}$ & $C_{\mathrm{sw}}$ & \multicolumn{2}{c}{lead} & $C_{\mathrm{dec}}$ & $C'_{\mathrm{dec}}$ & $C_{\mathrm{sw}}$ & \multicolumn{2}{c}{lead} \\
\cmidrule(lr){5-6}\cmidrule(lr){10-11}\cmidrule(lr){15-16}\cmidrule(lr){20-21}
 & & & & val. & test & & & & val. & test & & & & val. & test & & & & val. & test \\
\midrule
2 & 0.26 & 0.22 & 0.17 & $+0.09$ & $+0.10$ & 0.27 & 0.16 & 0.18 & $+0.09$ & $+0.09$ & 0.22 & 0.16 & 0.17 & $+0.05$ & $+0.06$ & 0.24 & 0.35 & 0.19 & $+0.05$ & $+0.03$ \\
3 & 0.30 & 0.23 & 0.21 & $+0.09$ & $+0.08$ & 0.30 & 0.22 & 0.24 & $+0.06$ & $+0.05$ & 0.26 & 0.18 & 0.24 & $+0.02$ & $+0.04$ & 0.27 & 0.24 & 0.24 & $+0.04$ & $+0.03$ \\
4 & 0.29 & 0.18 & 0.25 & $+0.04$ & $+0.03$ & 0.30 & 0.18 & 0.28 & $+0.02$ & $+0.02$ & 0.26 & 0.20 & 0.28 & $-0.02$ & $-0.01$ & 0.29 & 0.28 & 0.28 & $+0.01$ & $+0.01$ \\
5 & 0.21 & -- & 0.14 & $+0.07$ & $+0.05$ & -- & -- & -- & -- & -- & 0.29 & -- & 0.30 & $-0.02$ & $+0.00$ & 0.30 & -- & 0.32 & $-0.02$ & $-0.02$ \\
6 & 0.18 & -- & 0.15 & $+0.03$ & $+0.01$ & -- & -- & -- & -- & -- & 0.19 & -- & 0.21 & $-0.02$ & $-0.02$ & 0.16 & -- & 0.17 & $-0.02$ & $-0.01$ \\
8 & 0.18 & -- & 0.20 & $-0.02$ & $-0.04$ & -- & -- & -- & -- & -- & 0.15 & -- & 0.22 & $-0.07$ & $-0.08$ & 0.20 & -- & 0.24 & $-0.05$ & $-0.06$ \\
\bottomrule
\end{tabular}
}
\end{table*}
The lead of \method over a full-resolution schedule is the difference of two costs in \lpips. The schedule's cost of deciding in 1 or 2 large steps falls with the budget, and the cost of the switch of \method does not. Let $E_{\mathrm{sch}}$ be the error of the schedule, the calibrated one up to 4 NFEs and SE + shrink beyond, and let $E_{\mathrm{DtR}}$, $E_{\mathrm{re}}$ and the switch cost $C_{\mathrm{sw}}$ be as in \cref{app:switch}. The decision cost $C_{\mathrm{dec}}=E_{\mathrm{sch}}-E_{\mathrm{re}}$ is what the schedule loses by deciding the image in 1 or 2 large full-resolution steps. At every budget the lead is
\begin{equation}
\underbrace{E_{\mathrm{sch}}-E_{\mathrm{DtR}}}_{\text{lead}}=\underbrace{(E_{\mathrm{sch}}-E_{\mathrm{re}})}_{C_{\mathrm{dec}}}-\underbrace{(E_{\mathrm{DtR}}-E_{\mathrm{re}})}_{C_{\mathrm{sw}}}.
\label{eq:lead}
\end{equation}

At 2 to 4 NFEs, 1 or 2 full-resolution steps must cross the whole range of noise levels where the composition is decided, and the decision cost of FLUX.1-dev is 0.26--0.30 (\cref{fig:lead,tab:budget}). At 6 and 8 NFEs the first step of a schedule that is dense at high noise is shorter, and the decision cost falls to 0.18. The switch cost does not fall with the budget. The difference of the two matches the lead on the test set within 0.03 at every budget and changes sign between 6 and 8 NFEs. A full-resolution stage whose first step closely follows the switch lowers $E_{\mathrm{re}}$ more than $E_{\mathrm{DtR}}$, so the switch cost grows there, to 0.20 for FLUX.1-dev at 8 NFEs with $\sh=0.95$. This is the resolution-specific error of \cref{prop:floor}, which a denser full-resolution stage does not remove.

Part of the decision cost comes from the full-resolution stage of \method having more steps below $\sh$ than the schedule. $C'_{\mathrm{dec}}$ removes this part, as it re-enters the schedule itself at its own noise level nearest $\sh$. On all 4 models this re-entry still removes 25--52\% of the schedule's error, and re-entering the schedule after its first step removes 23--37\% at 4 NFEs. On the CFG models, part of the decision differs between resolutions (\cref{app:guidance}), so from 3 NFEs on, their switch cost is 0.02--0.06 larger than on FLUX.1-dev at the same switch point, and the lead closes earlier. Their configurations in \cref{tab:main,tab:regime}, from 4 NFEs on Qwen-Image and from 5 on SD3.5-Large, take the coarse switch noise from the stage's own noise estimate, which adds 0.012--0.032 \lpips to the lead (\cref{tab:noise}). On FLUX.2-dev the switch cost already nearly equals the decision cost at 4 NFEs. Caching accelerators pay neither cost once their key steps are dense enough, which is why they take over from 8 key evaluations (\cref{app:caching}).

The diamonds of \cref{fig:decision}a start the 1 to 3 full-resolution steps of \method from the reference trajectory's state at $\sigma=0.9$. At 3 and 4 NFEs they end even closer to the reference than the calibrated schedule re-entered at its own noise level nearest 0.9, so the low-resolution stage only needs to provide a nearly correct state at the switch, as \cref{sec:lowres} shows.

\subsection{Caching accelerators}
\label{app:caching}
\begin{table}[t]
\centering\footnotesize
\caption{\textbf{Caching accelerators at matched budget.} FLUX.1-dev, 1000 prompts, same seed. The budget is the NFE count of \method and the number of full evaluations of a caching accelerator. Seconds per image on one GPU include the 50 outer steps, and \method is in bold.}
\label{tab:cache}
{\scriptsize\setlength{\tabcolsep}{1.8pt}\begin{tabular}{@{}l r r r r r r@{}}
\toprule
sampler & budget & s/img & LPIPS$\downarrow$ & DS$\downarrow$ & IR$\uparrow$ & PS$\uparrow$ \\
\midrule
\multicolumn{7}{@{}l}{\textit{budget 4--5}} \\
TaylorSeer, interval 24 & 4.0 & 2.19 & 0.825 & 0.667 & -1.56 & 17.77 \\
DPCache ($K{=}5$, infeasible) & 5.0 & 2.03 & 0.888 & 0.749 & -2.14 & 16.94 \\
\textbf{\method, 4 NFE} & \textbf{4.0} & \textbf{1.31} & \textbf{0.457} & \textbf{0.166} & \textbf{0.82} & \textbf{21.44} \\
\textbf{\method, 5 NFE} & \textbf{5.0} & \textbf{1.62} & \textbf{0.416} & \textbf{0.148} & \textbf{0.85} & \textbf{21.62} \\
\midrule
\multicolumn{7}{@{}l}{\textit{budget 6}} \\
TaylorSeer, interval 12 & 6.0 & 2.80 & 0.620 & 0.342 & 0.40 & 20.35 \\
DPCache ($K{=}6$) & 6.0 & 2.32 & 0.439 & 0.183 & 0.86 & 21.03 \\
\textbf{\method, 6 NFE} & \textbf{6.0} & \textbf{1.92} & \textbf{0.389} & \textbf{0.137} & \textbf{0.89} & \textbf{21.73} \\
\midrule
\multicolumn{7}{@{}l}{\textit{budget 7--8}} \\
DPCache ($K{=}7$) & 7.0 & 2.61 & 0.323 & 0.113 & 0.92 & 21.24 \\
TaylorSeer, interval 8 & 8.0 & 3.39 & 0.430 & 0.171 & 0.91 & 21.60 \\
DPCache ($K{=}8$) & 8.0 & 2.91 & 0.270 & 0.080 & 0.97 & 21.58 \\
SE + shrink & 8.0 & 2.53 & 0.319 & 0.102 & 0.93 & 21.77 \\
\textbf{\method, 8 NFE} & \textbf{8.0} & \textbf{2.52} & \textbf{0.361} & \textbf{0.122} & \textbf{0.92} & \textbf{21.79} \\
\midrule
\multicolumn{7}{@{}l}{\textit{budget 9--13}} \\
DPCache ($K{=}9$) & 9.0 & 3.20 & 0.235 & 0.066 & 0.96 & 21.60 \\
TaylorSeer, interval 5 & 12.0 & 4.56 & 0.290 & 0.092 & 0.96 & 21.92 \\
DPCache ($K{=}13$) & 13.0 & 4.37 & 0.130 & 0.031 & 0.96 & 21.87 \\
\bottomrule
\end{tabular}
}
\end{table}
\begin{table}[t]
\centering\scriptsize
\caption{\textbf{Low-resolution previews of CZP}, our reimplementation of \cite{lowrespreview2026} (\cref{app:czp}). $N=8$, 12 and 30 run without the correction, and the published $N=30$ runs with it. $512^2$ outputs against the reference downsampled to $512^2$. Same prompts, seeds and references as \method. Every metric, and below, \method minus CZP on the same prompts, where $^*$ marks a 95\% interval that excludes zero. CZP with $N=8$ is the CZP ($512^2$) of \cref{tab:main,tab:mainfull}.}
\label{tab:czp}
{\setlength{\tabcolsep}{2.2pt}\begin{tabular}{@{}l r r r r r r r r@{}}
\toprule
sampler & NFE & sec & $\times$ & LPIPS$\downarrow$ & DS$\downarrow$ & DINO$\uparrow$ & IR$\uparrow$ & PS$\uparrow$ \\
\midrule
\multicolumn{9}{@{}l}{\textit{FLUX.1-dev, 1000 prompts}} \\
\midrule
default, 50 steps & 50 & 15.2 & 1 & -- & -- & -- & 0.94 & 21.81 \\
\method, 3 NFEs & 2.95 & 1.01 & 15.0 & 0.491 & 0.192 & 0.726 & 0.63 & 21.01 \\
\method, 4 NFEs & 3.95 & 1.31 & 11.6 & 0.457 & 0.166 & 0.764 & 0.82 & 21.44 \\
CZP, $N{=}8$ & 6.86 & 2.14 & 7.1 & 0.559 & 0.253 & 0.678 & 0.59 & 21.09 \\
CZP, $N{=}12$ & 8.82 & 2.73 & 5.6 & 0.506 & 0.201 & 0.738 & 0.76 & 21.38 \\
CZP, $N{=}30$ & 18.63 & 5.72 & 2.7 & 0.381 & 0.111 & 0.833 & 0.88 & 21.64 \\
CZP, published & 20.22 & 6.21 & 2.5 & 0.359 & 0.104 & 0.846 & 0.85 & 21.51 \\
\midrule
\multicolumn{9}{@{}l}{\textit{SD3.5-Large, 500 prompts}} \\
\midrule
default, 50 steps & 50 & 14.3 & 1 & -- & -- & -- & 1.05 & 21.95 \\
\method, 3 NFEs & 3.08 & 0.98 & 14.5 & 0.576 & 0.270 & 0.675 & 0.20 & 20.46 \\
\method, 4 NFEs & 4.08 & 1.27 & 11.3 & 0.543 & 0.238 & 0.716 & 0.51 & 20.92 \\
CZP, $N{=}8$ & 6.57 & 1.89 & 7.6 & 0.663 & 0.273 & 0.666 & 0.49 & 20.76 \\
CZP, $N{=}12$ & 8.43 & 2.42 & 5.9 & 0.596 & 0.232 & 0.712 & 0.66 & 21.08 \\
CZP, $N{=}30$ & 17.85 & 5.08 & 2.8 & 0.473 & 0.167 & 0.773 & 0.85 & 21.32 \\
CZP, published & 19.28 & 5.48 & 2.6 & 0.448 & 0.150 & 0.790 & 0.86 & 21.35 \\
\bottomrule
\end{tabular}
}\\[4pt]
{\setlength{\tabcolsep}{2.2pt}\begin{tabular}{@{}l r r r r r r@{}}
\toprule
\method $-$ CZP & time ratio & $\Delta$LPIPS & $\Delta$DS & $\Delta$DINO & $\Delta$IR & $\Delta$PS \\
\midrule
\multicolumn{7}{@{}l}{\textit{FLUX.1-dev}} \\
$N{=}8$ vs.\ 3 NFEs & 2.1$\times$ & $-$0.068$^*$ & $-$0.062$^*$ & +0.049$^*$ & +0.04 & $-$0.08$^*$ \\
$N{=}12$ vs.\ 4 NFEs & 2.1$\times$ & $-$0.049$^*$ & $-$0.034$^*$ & +0.026$^*$ & +0.06$^*$ & +0.06$^*$ \\
published vs.\ 4 NFEs & 4.7$\times$ & +0.098$^*$ & +0.062$^*$ & $-$0.082$^*$ & $-$0.03 & $-$0.07$^*$ \\
\midrule
\multicolumn{7}{@{}l}{\textit{SD3.5-Large}} \\
$N{=}8$ vs.\ 3 NFEs & 1.9$\times$ & $-$0.087$^*$ & $-$0.003 & +0.009 & $-$0.29$^*$ & $-$0.31$^*$ \\
$N{=}12$ vs.\ 4 NFEs & 1.9$\times$ & $-$0.053$^*$ & +0.006 & +0.004 & $-$0.16$^*$ & $-$0.16$^*$ \\
published vs.\ 4 NFEs & 4.3$\times$ & +0.094$^*$ & +0.088$^*$ & $-$0.074$^*$ & $-$0.36$^*$ & $-$0.43$^*$ \\
\bottomrule
\end{tabular}
}
\end{table}
Below 6 full evaluations the caching accelerators have no usable operating point, and at 6 \method is more faithful and faster (\cref{tab:cache}). DPCache's dynamic program may not skip more than 30\% of the schedule at once, so it has no feasible schedule below 6 key steps. Its released code then falls back to computing the first $K$ steps and forecasting the rest, which gives \lpips 0.888 at $K=5$. TaylorSeer's forecast fails at intervals of 12 steps or more and leaves a dotted texture over the image, with \lpips 0.825 and 0.620 at 4 and 6 full evaluations. At 6 full evaluations \method is closer to the reference than both on every fidelity metric and rated higher by \ps. From 8 key evaluations and 2.9\,s on, DPCache is closer to the reference than every few-step sampler, and with 13 it reaches 0.130 in 4.4\,s. TaylorSeer stays behind every other sampler of \cref{fig:cache} at equal time. Caching takes over at the budget where the lead of \method over full-resolution schedules changes sign (\cref{app:lead}).

\subsection{Low-resolution previews of CZP}
\label{app:previews}
At twice the time of \method, the previews of CZP with $N=8$ and 12 are less faithful than \method on FLUX.1-dev at 3 and 4 NFEs on all 3 fidelity metrics (\cref{tab:czp}). On SD3.5-Large, at twice the time of \method, they are less faithful in \lpips and on par in \ds and DINO. On both models the published setting is closer to the reference than \method at 4 NFEs, at 4.7$\times$ and 4.3$\times$ its time. On FLUX.1-dev, SE + shrink at 8 NFEs is closer still in 2.5\,s (\cref{fig:cache}).

\FloatBarrier
\section{Other models and video}
\label{app:transfer}
\noindent \method carries over to a held-out model whose settings were predicted before its test run (\cref{app:heldout}), works with distilled models as renderers (\cref{app:distill}) and transfers to video on Wan2.1 (\cref{app:video}).

\subsection{A held-out model, FLUX.2-klein-base}
\label{app:heldout}
\begin{figure*}[t]
\centering
\includegraphics[width=\textwidth]{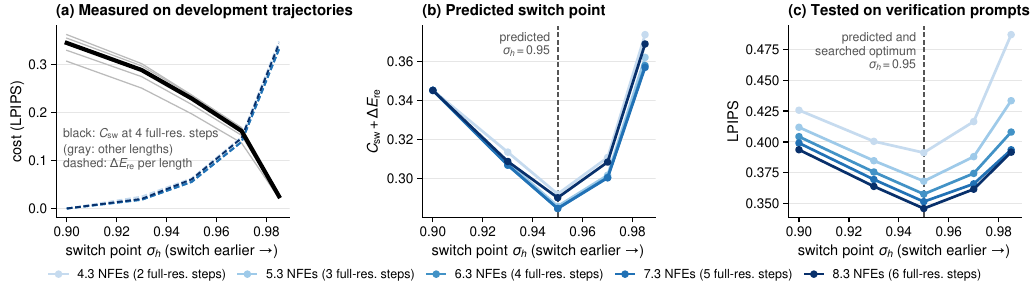}
\caption{\textbf{The switch point predicted on a held-out model.} FLUX.2-klein-base with a $768^2$ low-resolution stage. (a)~On the 36 development trajectories, the switch cost $C_{\mathrm{sw}}$ for the 4-step full-resolution stage in black, with other lengths in gray, and the cost of switching earlier $\Delta E_{\mathrm{re}}$ for stages of 2 to 6 steps, dashed. (b)~Their sum per stage. The dashed line marks its minimum, $\sh=0.95$ at all 5 stages, which is the switch point predicted before the test run. (c)~On the 36 verification prompts, the error of \method per stage. The searched optimum is $\sh=0.95$ at all 5 stages as well.}
\label{fig:transfer}
\end{figure*}
\begin{table*}[t]
\centering\scriptsize
\caption{\textbf{FLUX.2-klein-base on the 500 test prompts.} Same seeds as the other models. Seconds per image from its own interleaved run on the same GPU. Every metric, \method minus each baseline with 95\% prompt-cluster intervals, where $^*$ marks those excluding zero, and GenEval with the official protocol.}
\label{tab:klein}
{\setlength{\tabcolsep}{3.6pt}\begin{tabular}{@{}c l r r r r r r r r r@{}}
\toprule
& sampler & NFE & s/img & speed-up ($\times$) & LPIPS$\downarrow$ & DreamSim$\downarrow$ & DINO$\uparrow$ & ImageReward$\uparrow$ & PickScore$\uparrow$ & FD-DINO$\downarrow$ \\
\midrule
\multicolumn{2}{@{}l}{default, 50 steps} & 50 & 14.0 & 1 & -- & -- & -- & 0.94 & 21.23 & -- \\
\midrule
$\approx$20$\times$ & default & 2 & 0.66 & 21.2 & 0.664 & 0.484 & 0.395 & -1.27 & 18.74 & 1353 \\
 & \textbf{\method} & \textbf{2.14} & \textbf{0.70} & \textbf{19.9} & \textbf{0.584}$^*$ & \textbf{0.361}$^*$ & \textbf{0.548}$^*$ & \textbf{-0.72}$^*$ & \textbf{19.12}$^*$ & 1054 \\
\midrule
$\approx$14$\times$ & default & 3 & 0.94 & 14.9 & 0.527 & 0.302 & 0.630 & -0.11 & 19.78 & 843 \\
 & \textbf{\method} & \textbf{3.14} & \textbf{0.98} & \textbf{14.3} & \textbf{0.440}$^*$ & \textbf{0.214}$^*$ & \textbf{0.739}$^*$ & \textbf{0.32}$^*$ & \textbf{20.20}$^*$ & 618 \\
\midrule
$\approx$11$\times$ & default & 4 & 1.22 & 11.5 & 0.434 & 0.210 & 0.744 & 0.38 & 20.36 & 598 \\
 & SharpEuler & 4 & 1.22 & 11.5 & 0.421 & 0.203 & 0.751 & 0.41 & 20.36 & 590 \\
 & \textbf{\method} & \textbf{4.27} & \textbf{1.30} & \textbf{10.8} & \textbf{0.393}$^*$ & \textbf{0.172}$^*$ & \textbf{0.788}$^*$ & \textbf{0.50}$^*$ & \textbf{20.39} & 517 \\
\bottomrule
\end{tabular}
}\\[4pt]
{\setlength{\tabcolsep}{2.2pt}\begin{tabular}{@{}l l l l l l@{}}
\toprule
group: \method\ minus & LPIPS$\downarrow$ & DreamSim$\downarrow$ & DINO$\uparrow$ & ImageReward$\uparrow$ & PickScore$\uparrow$ \\
\midrule
$\approx$20$\times$: default & $-0.079^*$ {\scriptsize[$-$0.087, $-$0.072]} & $-0.123^*$ {\scriptsize[$-$0.132, $-$0.114]} & $+0.152^*$ {\scriptsize[+0.137, +0.167]} & $+0.55^*$ {\scriptsize[+0.48, +0.62]} & $+0.38^*$ {\scriptsize[+0.31, +0.45]} \\
$\approx$14$\times$: default & $-0.087^*$ {\scriptsize[$-$0.093, $-$0.080]} & $-0.088^*$ {\scriptsize[$-$0.095, $-$0.082]} & $+0.109^*$ {\scriptsize[+0.098, +0.120]} & $+0.44^*$ {\scriptsize[+0.38, +0.50]} & $+0.43^*$ {\scriptsize[+0.38, +0.48]} \\
$\approx$11$\times$: default & $-0.041^*$ {\scriptsize[$-$0.049, $-$0.033]} & $-0.038^*$ {\scriptsize[$-$0.045, $-$0.032]} & $+0.044^*$ {\scriptsize[+0.033, +0.055]} & $+0.11^*$ {\scriptsize[+0.06, +0.17]} & $+0.03$ {\scriptsize[$-$0.02, +0.07]} \\
\phantom{$\approx$11$\times$: }SharpEuler & $-0.028^*$ {\scriptsize[$-$0.036, $-$0.021]} & $-0.031^*$ {\scriptsize[$-$0.037, $-$0.025]} & $+0.037^*$ {\scriptsize[+0.027, +0.048]} & $+0.09^*$ {\scriptsize[+0.04, +0.14]} & $+0.03$ {\scriptsize[$-$0.01, +0.08]} \\
\bottomrule
\end{tabular}
}\\[4pt]
{\setlength{\tabcolsep}{6pt}\begin{tabular}{@{}l c c c@{}}
\toprule
GenEval & 2 NFEs & 3 NFEs & 4 NFEs \\
\midrule
default, SharpEuler at 4 NFEs & 0.226 & 0.536 & 0.654 \\
\textbf{\method} & \textbf{0.462} & \textbf{0.661} & \textbf{0.713} \\
default, 50 steps & \multicolumn{3}{c}{0.791} \\
\bottomrule
\end{tabular}
}
\end{table*}
\begin{figure*}[t]
\centering
\input{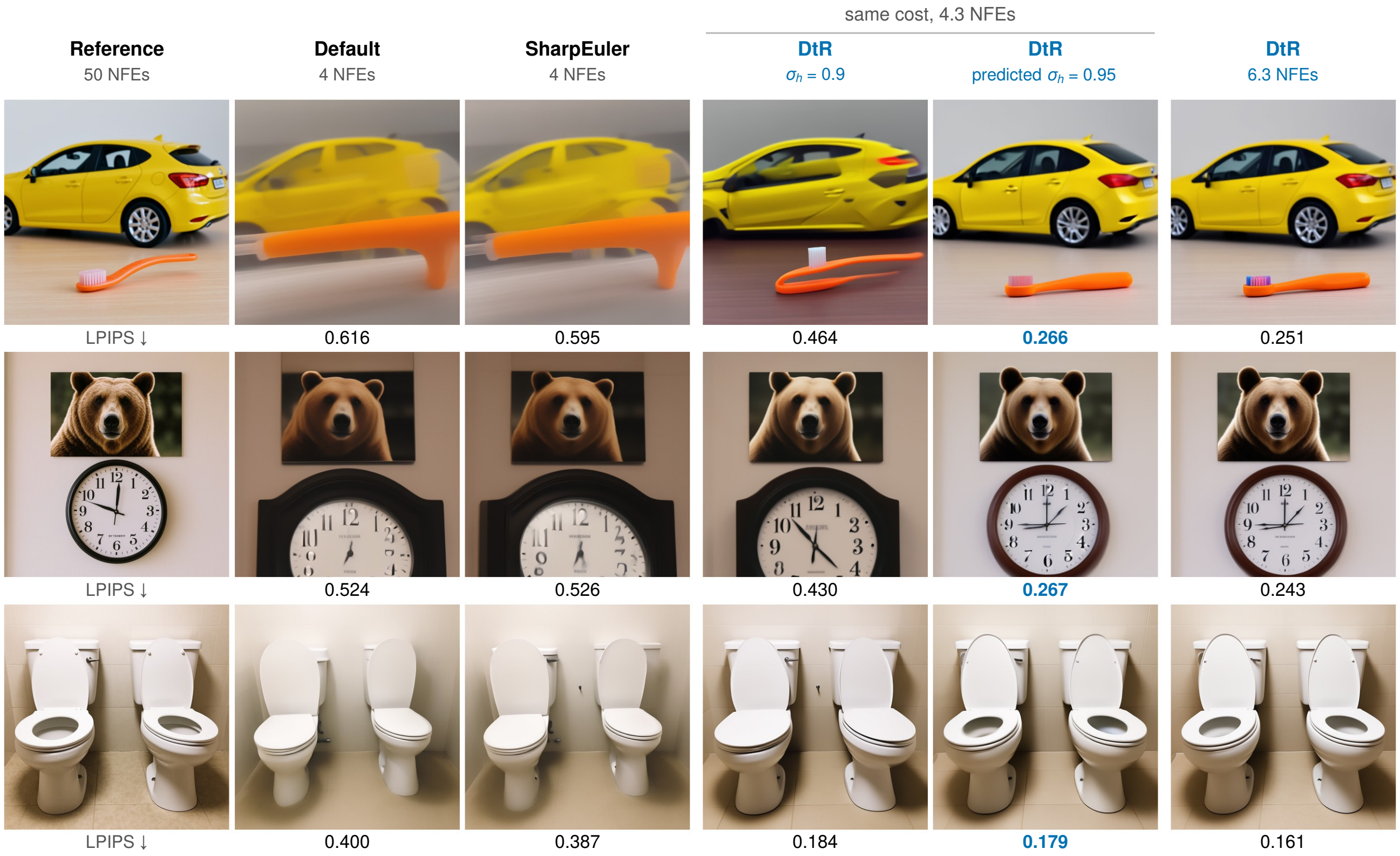}
\caption{\textbf{FLUX.2-klein-base, held-out model.} Test prompts with the same noise, same-seed \lpips to the model's 50-step sample under each image, and the lowest value up to 4.3 NFEs in bold. The two \method columns at 4.3 NFEs have the same cost and differ only in the switch point. The default 4-step sampler and SharpEuler blur or lose the layout. \method with the seen models' switch point $\sh=0.9$ keeps it partly. The predicted switch point $\sh=0.95$ keeps it at 4.3 NFEs, and 6.3 NFEs add detail.}
\label{fig:klein}
\end{figure*}
FLUX.2-klein-base played no part in designing the sampler. Its low resolution, switch point and switch noise were predicted before its test run, and the predictions held. The model is the undistilled 4B FLUX.2-klein-base at classifier-free guidance 4, and its development and verification prompts are those of \cref{tab:prompts}.

\paragraph{Low resolution} Before running the model, we fixed the rule of \cref{alg:newmodel}. A low resolution qualifies when an evaluation there is cheap enough to buy 3 low-resolution steps and the cross-resolution match of \cref{app:crossres} holds over the whole stage, down to $\sigma=0.9$. FLUX.2-klein-base qualifies at $768^2$, with a mean best-match cosine of 0.74, and not at $512^2$, where the match is lost below $\sigma=0.91$. The losses predicted for a $512^2$ stage and for fresh noise at the switch appeared as well. A $512^2$ stage loses the fine bands it has already decided and is 0.143 \lpips farther from the reference than full-resolution steps at the same noise levels.

\paragraph{Switch point} On 36 development trajectories we measured the cost of switching earlier for full-resolution stages of 2 to 6 evaluations, and the switch cost, which the prediction takes from the 4-evaluation stage (\cref{fig:transfer}a). Their sum predicted $\sh=0.95$ at every budget, including the shortest stage, where FLUX.1-dev and Qwen-Image switch at 0.90--0.93 (\cref{fig:transfer}b). We wrote this prediction down before the test run. A grid search over 5 switch points on the 36 verification prompts found 0.95 at all 5 budgets (\cref{fig:transfer}c). At the 2 shortest stages, the predicted switch point is 0.035 and 0.044 \lpips closer to the reference than $\sh=0.9$, the choice of the seen models for short stages. On this model the cost of switching earlier does not fall with the length of the full-resolution stage, which is why the shortest stage switches as early as the longest.

\paragraph{Switch noise and steps} The guidance direction of FLUX.2-klein-base keeps a cosine of 0.64 under the noise replacement, close to FLUX.1-dev's 0.70, so the seed's noise was predicted. The measured preference has the predicted sign and grows with the guidance scale as predicted (\cref{fig:noise}). The rule for the number of steps carries over as well, as the stage improves up to 4 steps at $768^2$ and not beyond (\cref{fig:supp_k3}).

\paragraph{Results} On the first 500 test prompts, with the seeds of the other models, \method is closer to the reference than the default sampler at 2 and 3 NFEs, and than the default sampler and SharpEuler at 4 NFEs, on every fidelity metric (\cref{tab:klein}). It is rated higher by \ir at every budget and by \ps at 2 and 3 NFEs. On GenEval it scores 0.236 and 0.126 higher than the default sampler at 2 and 3 NFEs and 0.060 higher than SharpEuler at 4 NFEs. \Cref{fig:klein} shows examples.

\subsection{Distilled models as renderers}
\label{app:distill}
\begin{table*}[t]
\centering\footnotesize
\caption{\textbf{Distilled models as renderers on FLUX.1-dev.} 1000 test prompts. Same noise and 50-step references as \method. Seconds per $1024^2$ image on one GPU. Rendered rows run FLUX.1-dev's low-resolution stage and switch, then the renderer's full-resolution steps. Another seed compares the distilled model's image from a different seed with the same reference. GenEval uses the official protocol on 553 prompts, 4 seeds per prompt for \method as in \cref{tab:geneval} and the first for the other rows. The default 50-step sampler scores 0.656. A dash marks values not measured.}
\label{tab:distill}
{\setlength{\tabcolsep}{4pt}\begin{tabular}{@{}l r r r r r r r r@{}}
\toprule
sampler & NFE & sec & \lpips & \ds & DINO & \ir & \ps & GenEval \\
\midrule
\multicolumn{9}{@{}l}{\textit{FLUX.1-dev and FLUX.1-schnell}} \\
FLUX.1-dev, default, 4 steps & 4 & 1.32 & 0.628 & 0.381 & 0.530 & $-$0.08 & 20.29 & 0.424 \\
\method, 2 NFEs & 1.95 & 0.71 & 0.598 & 0.325 & 0.572 & $-$0.24 & 19.95 & 0.507 \\
\method, 4 NFEs & 3.95 & 1.31 & 0.457 & 0.166 & 0.764 & 0.82 & 21.44 & 0.642 \\
FLUX.1-schnell, 2 steps & 2 & 0.64 & 0.589 & 0.283 & 0.661 & 0.97 & 21.67 & 0.654 \\
FLUX.1-schnell, 4 steps (default) & 4 & 1.18 & 0.585 & 0.286 & 0.656 & 0.96 & 21.59 & 0.648 \\
FLUX.1-schnell, 4 steps, another seed & 4 & 1.18 & 0.656 & 0.317 & 0.634 & 0.94 & 21.60 & -- \\
\method rendered by FLUX.1-schnell, 2 NFEs & 1.95 & 0.68 & 0.465 & 0.194 & 0.721 & 0.90 & 21.86 & 0.636 \\
\method rendered by FLUX.1-schnell, 4 NFEs & 3.95 & 1.22 & 0.448 & 0.190 & 0.731 & 0.90 & 21.71 & 0.624 \\
\midrule
\multicolumn{9}{@{}l}{\textit{FLUX.1-dev + Hyper-FLUX (8-step LoRA)}} \\
Hyper-FLUX, 8 steps (default) & 8 & 2.78 & 0.466 & 0.171 & 0.768 & 0.96 & 21.89 & 0.671 \\
Hyper-FLUX, 8 steps, another seed & 8 & 2.78 & 0.588 & 0.221 & 0.712 & 0.94 & 21.87 & -- \\
\method rendered by Hyper-FLUX, 2 NFEs & 1.95 & 0.79 & 0.542 & 0.244 & 0.682 & 0.35 & 20.48 & 0.570 \\
\method rendered by Hyper-FLUX, 4 NFEs & 3.95 & 1.44 & 0.431 & 0.152 & 0.781 & 0.89 & 21.64 & 0.654 \\
\bottomrule
\end{tabular}
}
\vspace*{-0.2em}
\end{table*}
\begin{figure*}[t]
\centering
\input{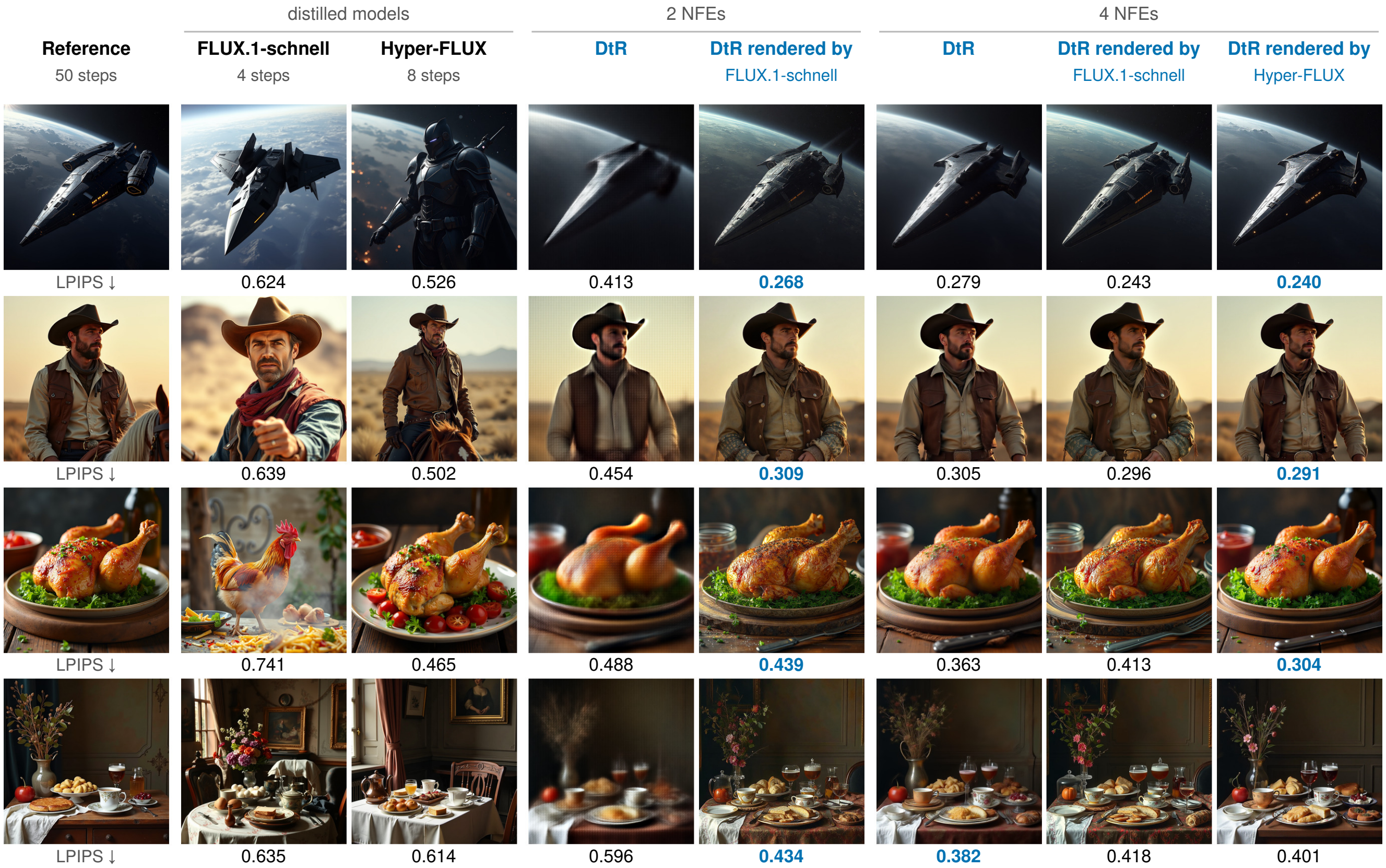}
\caption{\textbf{\method with distilled models on FLUX.1-dev.} Examples share the noise of the 50-step reference. Columns show the 50-step reference, the distilled models' own samples with FLUX.1-schnell's 4 steps and Hyper-FLUX's 8 steps, and \method at 2 and at 4 NFEs, alone and with its full-resolution steps rendered by a distilled model. Both distilled models change the image of the seed. With them as renderers, \method keeps it. Same-seed \lpips to the reference under each image, and the lowest value of each NFE group in bold.}
\label{fig:distill_qual}
\end{figure*}
A distilled model changes the seed's image. With such a model as the renderer, \method keeps the image at the distilled model's speed (\cref{tab:distill,fig:distill_qual}). FLUX.1-schnell \cite{ladd2024} is a step-distilled model of the FLUX.1 family that shares FLUX.1-dev's latent space, and Hyper-FLUX \cite{hypersd2024} is an 8-step LoRA of FLUX.1-dev trained to keep its trajectory. Each runs its default schedule from the noise of the 50-step reference. To render the decision of \method, FLUX.1-dev runs the low-resolution stage and the switch, and the distilled model takes the full-resolution steps at its own noise levels.

\paragraph{The distilled model's own image} FLUX.1-schnell's 4 steps keep only part of FLUX.1-dev's image. They are 0.13 \lpips farther from the reference than \method at 4 NFEs, and a schnell sample from another seed is only 0.07 farther still.

\paragraph{Rendered by FLUX.1-schnell} At 2 NFEs, \method rendered by FLUX.1-schnell is 0.12 \lpips closer to the reference than schnell's own 2 steps at about the same time. It is rated nearly as high as schnell and is on par with it on GenEval. It also comes within 0.01 \lpips of \method at 4 NFEs.

\paragraph{Rendered by Hyper-FLUX and FLUX.2-klein-4B} At 4 NFEs, \method rendered by Hyper-FLUX is 0.036 \lpips closer to the reference than Hyper-FLUX's own 8 steps in about half their time, and on par with them on GenEval. Within the FLUX.2-klein family, FLUX.2-klein-base decides with 2 steps at $768^2$ until $\sh=0.95$, and the step-distilled FLUX.2-klein-4B renders with a single full-resolution step. On the first 200 test prompts, this 2.14-NFE sample is 0.077 \lpips closer to the 50-step image of FLUX.2-klein-base than the 4 steps of FLUX.2-klein-4B. It takes 0.82$\times$ their sampling time without text encoding.

\subsection{Video}
\label{app:video}
\begin{figure*}[t]
\centering
\includegraphics[width=\textwidth]{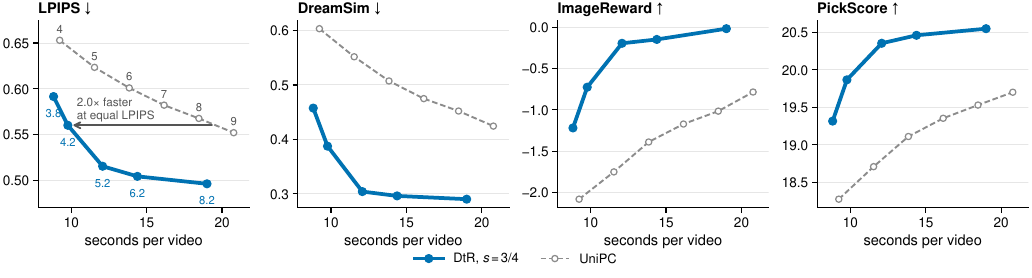}
\caption{\textbf{\method on Wan2.1-T2V-1.3B against seconds per video.} 100 held-out prompts, frame-level fidelity to the 50-step UniPC video and ratings. \method runs its stage at ${s=3/4}$ with 3.8 to 8.2 NFEs, and UniPC takes 4 to 9 steps. The arrow marks the time that UniPC needs to reach the \lpips of the 4.2-NFE configuration. Seconds per video are from the GPU of this subsection.}
\label{fig:video_curve}
\end{figure*}
\begin{table*}[t]
\centering\footnotesize
\caption{\textbf{Wan2.1-T2V-1.3B, 100 held-out prompts.} Frame-level, same-seed fidelity to the 50-step UniPC video, and seconds per video on one GPU. \method rows give $k$ low-resolution steps at ${s=3/4}$, the switch point $\sh$, and the full-resolution stage of $T$ steps on the default grid below $\sh$. On the right are paired differences of \method minus UniPC at the same and the next step count, with $^*$ if the 95\% interval excludes zero.}
\label{tab:video}
{\scriptsize\setlength{\tabcolsep}{2.4pt}\begin{tabular}{@{}l r r r r r r@{}}
\toprule
sampler & NFE & s/video & LPIPS$\downarrow$ & DS$\downarrow$ & IR$\uparrow$ & PS$\uparrow$ \\
\midrule
UniPC-4 & 4.00 & 9.2 & 0.653 & 0.603 & $-$2.08 & 18.27 \\
UniPC-5 & 5.00 & 11.5 & 0.623 & 0.552 & $-$1.75 & 18.71 \\
UniPC-6 & 6.00 & 13.8 & 0.601 & 0.507 & $-$1.39 & 19.11 \\
UniPC-7 & 7.00 & 16.2 & 0.582 & 0.475 & $-$1.17 & 19.35 \\
UniPC-8 & 8.00 & 18.5 & 0.568 & 0.452 & $-$1.02 & 19.53 \\
UniPC-9 & 9.00 & 20.8 & 0.552 & 0.424 & $-$0.78 & 19.70 \\
\midrule
\textbf{\method, $k{=}2$, $\sh{=}0.90$, $T{=}3$} & \textbf{3.82} & \textbf{8.8} & \textbf{0.592} & \textbf{0.458} & \textbf{$-$1.22} & \textbf{19.32} \\
\textbf{\method, $k{=}3$, $\sh{=}0.90$, $T{=}3$} & \textbf{4.23} & \textbf{9.8} & \textbf{0.560} & \textbf{0.388} & \textbf{$-$0.73} & \textbf{19.87} \\
\textbf{\method, $k{=}3$, $\sh{=}0.95$, $T{=}4$} & \textbf{5.24} & \textbf{12.1} & \textbf{0.515} & \textbf{0.304} & \textbf{$-$0.20} & \textbf{20.35} \\
\textbf{\method, $k{=}3$, $\sh{=}0.95$, $T{=}5$} & \textbf{6.24} & \textbf{14.4} & \textbf{0.504} & \textbf{0.296} & \textbf{$-$0.15} & \textbf{20.46} \\
\textbf{\method, $k{=}3$, $\sh{=}0.95$, $T{=}7$} & \textbf{8.24} & \textbf{19.0} & \textbf{0.496} & \textbf{0.290} & \textbf{$-$0.02} & \textbf{20.55} \\
\bottomrule
\end{tabular}
}\hfill
{\scriptsize\setlength{\tabcolsep}{3pt}\begin{tabular}{@{}l r r r r@{}}
\toprule
pair & $\Delta$LPIPS & $\Delta$DS & $\Delta$IR & $\Delta$PS \\
\midrule
4.2 NFEs $-$ UniPC-4 & $-$0.093$^*$ & $-$0.216$^*$ & +1.35$^*$ & +1.59$^*$ \\
4.2 NFEs $-$ UniPC-5 & $-$0.063$^*$ & $-$0.164$^*$ & +1.02$^*$ & +1.16$^*$ \\
3.8 NFEs $-$ UniPC-4 & $-$0.061$^*$ & $-$0.146$^*$ & +0.86$^*$ & +1.04$^*$ \\
5.2 NFEs $-$ UniPC-5 & $-$0.108$^*$ & $-$0.248$^*$ & +1.55$^*$ & +1.65$^*$ \\
5.2 NFEs $-$ UniPC-6 & $-$0.086$^*$ & $-$0.203$^*$ & +1.19$^*$ & +1.24$^*$ \\
6.2 NFEs $-$ UniPC-6 & $-$0.097$^*$ & $-$0.211$^*$ & +1.24$^*$ & +1.35$^*$ \\
6.2 NFEs $-$ UniPC-7 & $-$0.078$^*$ & $-$0.179$^*$ & +1.02$^*$ & +1.11$^*$ \\
8.2 NFEs $-$ UniPC-8 & $-$0.071$^*$ & $-$0.162$^*$ & +1.00$^*$ & +1.02$^*$ \\
8.2 NFEs $-$ UniPC-9 & $-$0.056$^*$ & $-$0.134$^*$ & +0.77$^*$ & +0.85$^*$ \\
\bottomrule
\end{tabular}
}
\end{table*}
\begin{table*}[t]
\centering\footnotesize
\caption{\textbf{VBench quality dimensions on 200 VBench prompts.} VBench 0.1.5 in custom-input mode, SPEED's 6 video-evaluation dimensions, means over videos. Dynamic degree is the fraction of videos judged dynamic from their optical flow. UniPC-25 is SPEED's 25-step baseline. SPEED runs its first stage at $s=2/3$ or $s=1/2$.}
\label{tab:video_vbench}
{\setlength{\tabcolsep}{6pt}\begin{tabular}{@{}l r r r r r r r@{}}
\toprule
sampler & NFE & Subj.\ cons.$\uparrow$ & Bg.\ cons.$\uparrow$ & Motion sm.$\uparrow$ & Dynamic$\uparrow$ & Aesthetic$\uparrow$ & Imaging$\uparrow$ \\
\midrule
UniPC, 50 steps & 50 & 0.962 & 0.957 & 0.989 & 0.390 & 0.570 & 0.688 \\
UniPC-25 & 25 & 0.958 & 0.953 & 0.987 & 0.435 & 0.560 & 0.658 \\
\midrule
UniPC-4 & 4.00 & 0.938 & 0.953 & 0.985 & 0.160 & 0.375 & 0.390 \\
UniPC-5 & 5.00 & 0.941 & 0.951 & 0.984 & 0.180 & 0.419 & 0.436 \\
UniPC-6 & 6.00 & 0.942 & 0.949 & 0.984 & 0.210 & 0.451 & 0.473 \\
UniPC-7 & 7.00 & 0.945 & 0.948 & 0.983 & 0.245 & 0.474 & 0.504 \\
UniPC-8 & 8.00 & 0.945 & 0.949 & 0.983 & 0.275 & 0.492 & 0.528 \\
UniPC-9 & 9.00 & 0.945 & 0.949 & 0.983 & 0.310 & 0.503 & 0.551 \\
\midrule
\textbf{\method, $k{=}2$, $\sh{=}0.90$, $T{=}3$} & \textbf{3.82} & \textbf{0.917} & \textbf{0.936} & \textbf{0.982} & \textbf{0.355} & \textbf{0.438} & \textbf{0.459} \\
\textbf{\method, $k{=}3$, $\sh{=}0.90$, $T{=}3$} & \textbf{4.23} & \textbf{0.922} & \textbf{0.933} & \textbf{0.981} & \textbf{0.515} & \textbf{0.489} & \textbf{0.548} \\
\textbf{\method, $k{=}3$, $\sh{=}0.95$, $T{=}4$} & \textbf{5.24} & \textbf{0.930} & \textbf{0.934} & \textbf{0.982} & \textbf{0.570} & \textbf{0.514} & \textbf{0.611} \\
\textbf{\method, $k{=}3$, $\sh{=}0.95$, $T{=}5$} & \textbf{6.24} & \textbf{0.930} & \textbf{0.937} & \textbf{0.983} & \textbf{0.595} & \textbf{0.518} & \textbf{0.614} \\
\textbf{\method, $k{=}3$, $\sh{=}0.95$, $T{=}7$} & \textbf{8.24} & \textbf{0.932} & \textbf{0.936} & \textbf{0.982} & \textbf{0.605} & \textbf{0.528} & \textbf{0.636} \\
\midrule
SPEED $s{=}2/3$, 6 steps & 3.92 & 0.917 & 0.937 & 0.982 & 0.355 & 0.427 & 0.451 \\
SPEED $s{=}2/3$, 8 steps & 5.23 & 0.920 & 0.937 & 0.981 & 0.430 & 0.463 & 0.510 \\
SPEED $s{=}2/3$, 10 steps & 6.53 & 0.922 & 0.935 & 0.981 & 0.485 & 0.485 & 0.553 \\
SPEED $s{=}1/2$, 8 steps & 5.43 & 0.906 & 0.938 & 0.975 & 0.410 & 0.449 & 0.503 \\
SPEED $s{=}1/2$, 10 steps & 6.58 & 0.907 & 0.937 & 0.973 & 0.445 & 0.467 & 0.538 \\
\bottomrule
\end{tabular}
}
\end{table*}
\begin{figure*}[t]
\centering
\input{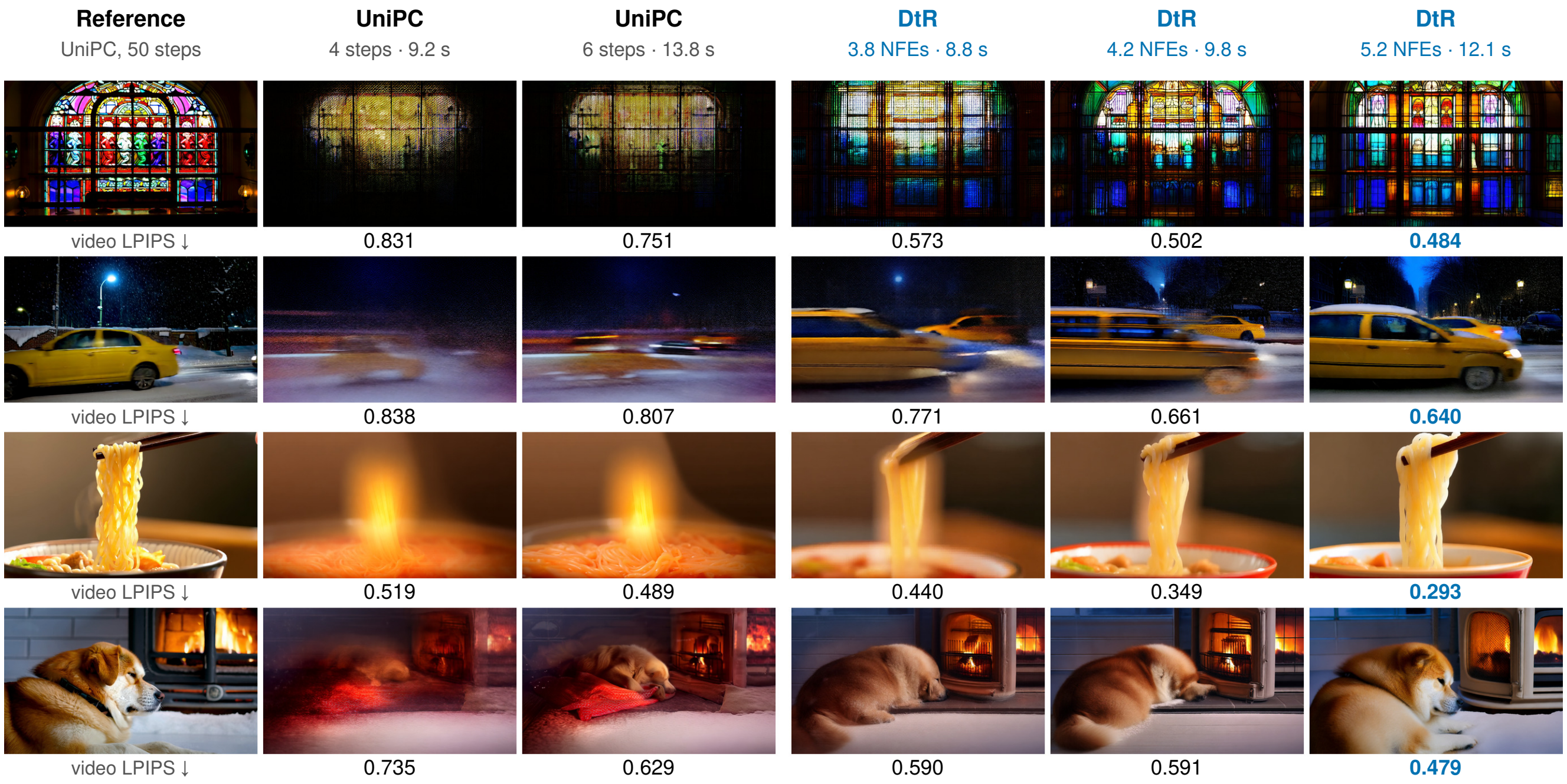}
\caption{\textbf{Frame 40 of 4 held-out prompts.} Wan2.1-T2V-1.3B with the same noise as the 50-step reference. Under each frame is the video's \lpips to the reference, the mean over 9 frames, best per row in bold. Seconds per video are from the GPU of this subsection. UniPC at 4 and 6 steps leaves the frames dark or blurred, while \method recovers the reference's stained-glass window, taxi in the snow, lifted noodles and dog at the fireplace, and is closer to it at all 3 budgets, each faster than 6 UniPC steps.}
\label{fig:qual_video}
\end{figure*}
The same sampler transfers to video. On Wan2.1-T2V-1.3B it is ahead of the default sampler at every budget, and from 4.2 NFEs on even when the default sampler takes an extra step (\cref{tab:video}).

\paragraph{Setup} We run Wan2.1-T2V-1.3B \cite{wan2025} with Diffusers weights at $480\times832$ and 81 frames. It uses classifier-free guidance 5.0 and the official flow shift 3.0. The default UniPC sampler \cite{unipc2023} with 50 steps gives the reference video of each seed, and \method takes Euler steps in both stages, as in \cref{alg:method}. Fidelity is measured per frame. \lpips and \ds compare frames 0, 10, \dots, 80 of a few-step video with those of its reference and are averaged per video, and \ir and \ps are averaged over the same frames. The configuration is set on 16 prompts, and 100 held-out prompts confirm it. A further 200 prompts, drawn evenly from the prompt lists of VBench's 6 quality dimensions, give the VBench evaluation at the scale of SPEED's. Seconds per video are measured on an RTX PRO 6000 Blackwell Server Edition GPU, a different GPU from that of the image experiments.

\paragraph{Configuration} The algorithm transfers unchanged, and its settings follow \cref{alg:newmodel}. On Wan the lowest resolution at which the stage keeps the reference's composition is ${s=3/4}$, rounded to $352\times624$ on the model's 16-pixel grid, with Fourier band-limited noise. It is 0.049--0.057 \lpips more faithful than a half-resolution stage on the 16 tuning prompts. The switch point follows the two-cost rule of \cref{app:switch}, whose switch cost does not depend on the budget here either. The minimum of the summed curve matches the searched optimum in 5 of 6 settings, and it gives $\sh=0.90$ with the 3-step full-resolution stage and 0.95 with longer stages. The stage's own noise estimate does not help on Wan, so the switch keeps the seed's noise.

\paragraph{Fidelity and ratings} On the 100 held-out prompts every paired comparison favors \method on \lpips, \ds, \ir and \ps, against UniPC with the nearest step count and, from 4.2 NFEs on, with one step more (\cref{tab:video}). At equal \lpips, UniPC needs about twice the time of the 4.2-NFE configuration, 20\,s against 9.8\,s (\cref{fig:video_curve}). \Cref{fig:qual_video} shows frames.

\paragraph{Motion and VBench} UniPC's few-step videos barely move, with a dynamic degree of 0.16--0.31 against 0.39 for the 50-step sampler, whereas the videos of \method move (\cref{tab:video_vbench}). With 2 low-resolution steps the motion of \method is close to the reference's, with a dynamic degree of 0.36 against 0.39. Against UniPC with the same and with an extra step, \method is 0.07--0.18 higher in imaging quality and 0.03--0.11 higher in aesthetic quality. On subject and background consistency, which favor static videos, it stays within 0.022 of UniPC, and within 0.004 on motion smoothness. At 8.2 NFEs \method takes 19\,s, and UniPC with 25 steps, the baseline of SPEED's video evaluation, takes 58\,s. \method comes within 0.022 of it in imaging quality and within 0.032 in aesthetic quality, where UniPC with 9 steps is 0.107 and 0.057 below it.

\paragraph{SPEED on its own video model} We run SPEED's official video schedule with its published Wan constants in our implementation, whose expansion and transition functions reproduce the official code bit for bit. As in the image comparison of \cref{app:multires}, the reference's noise is coupled into both of its stages. On a subset of 32 held-out prompts, SPEED is less faithful than \method at equal time, 0.13--0.15 \lpips behind with its half-resolution first stage and 0.03--0.09 behind with the two-thirds stage that its README recommends. On VBench, from 5 NFEs on, \method also scores 0.03--0.11 higher than SPEED at similar NFEs in imaging and aesthetic quality (\cref{tab:video_vbench}).

\FloatBarrier
\section{Ablations}
\label{app:ablation}
\noindent This section gives the protocols of \cref{tab:ablation} and further ablations. The switch-point sweep is in \cref{app:switch}, the number and resolution of the low-resolution steps are in \cref{app:steps}, and the controlled comparison of where and how SPEED switches is in \cref{app:multires}.

\subsection{Protocols of the ablations}
\label{app:protocols}
\begin{table}[t]
\centering\footnotesize
\caption{\textbf{Protocols of the rows of \cref{tab:ablation,tab:ablation_lowres}.} Prompt sets as in \cref{tab:prompts}.}
\label{tab:protocols}
\setlength{\tabcolsep}{3.5pt}
\begin{tabular}{@{}l l l@{}}
\toprule
rows & prompts & budget \\
\midrule
noise at the switch & development & 4 NFEs \\
evolved state, late switch & test, FLUX.1-dev & 4 NFEs \\
more steps, second order & development & 4 NFEs \\
$s=3/8$ & ablation & 4 NFEs \\
noise-level map & 30 validation & 4 NFEs \\
switch at 0.95 & validation & 5 full-res.\ steps \\
schedule split & development & 4.95 NFEs \\
\bottomrule
\end{tabular}

\end{table}
Each group of rows in \cref{tab:ablation,tab:ablation_lowres} uses its own prompts and budget (\cref{tab:protocols}). The late switch and step placement at $\sigma\approx0.76$ come from the controlled comparison of \cref{tab:speed2x2} on the 1000 test prompts, where they cost 0.051 \lpips.

\subsection{Inside the low-resolution stage}
\label{app:inside}
\begin{figure*}[t]
\centering
\includegraphics[width=\textwidth]{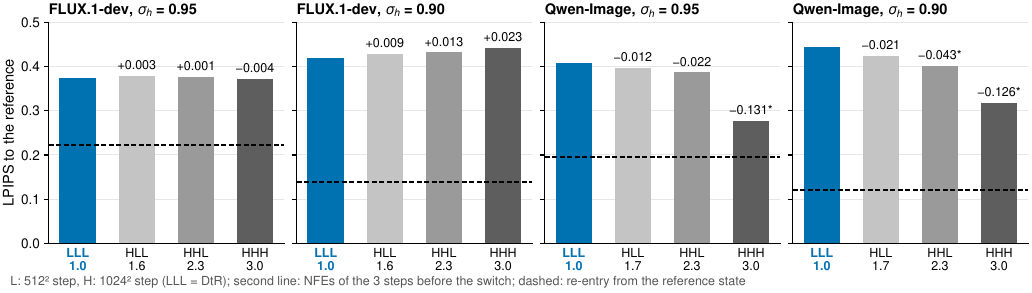}
\caption{\textbf{Resolution of the 3 steps of the low-resolution stage.} 30 validation prompts, the full-resolution stage of the 6-NFE configuration, the seed's noise at the switch. L and H are $512^2$ and $1024^2$ steps, and LLL is \method. Each schedule's NFEs are under it. Labels are paired \lpips differences, each schedule minus LLL, with $^*$ if the interval excludes zero. Dashed is the re-entry from the reference state with the same full-resolution stage (\cref{app:switch}).}
\label{fig:supp_schedule}
\end{figure*}
\begin{table}[t]
\centering\footnotesize
\caption{\textbf{Ablations of the low-resolution stage.} As in \cref{tab:ablation}, each row changes one part of \method, and $\Delta$\lpips is positive when the variant is less faithful. The swapped noise-level map runs the low-resolution stage of FLUX.1-dev on SNR-shifted noise levels and that of Qwen-Image on unshifted ones. Dashes mark variants not run on Qwen-Image.}
\label{tab:ablation_lowres}
{\setlength{\tabcolsep}{2.4pt}\begin{tabular}{@{}l r r@{}}
\toprule
variant & FLUX.1 & Qwen \\
\midrule
24 / 48 steps (vs.\ 3) & $+0.007$ / $+0.014$ & -- \\
$s=3/8$, equal cost (vs.\ $1/2$) & $+0.030^*$ & -- \\
second-order integrator & $+0.001$ & -- \\
swapped noise-level map & $+0.039^*$ & $+0.001$ \\
\bottomrule
\end{tabular}
}
\end{table}
\Cref{tab:ablation_lowres} gives the ablations of the low-resolution stage that \cref{sec:ablation} summarizes. More steps and a second-order integrator do not help, and a lower resolution at equal cost is less faithful. Swapping the noise-level maps of FLUX.1-dev and Qwen-Image hurts on FLUX.1-dev and changes little on Qwen-Image.

At 3 steps, running each step of the stage at half or at full resolution makes no difference on FLUX.1-dev (\cref{fig:supp_schedule}). Stages of 3 half-resolution steps, of mixed steps and of 3 full-resolution steps are on par with each other on 30 validation prompts at both switch points, within 0.023 \lpips. A full-resolution first step, as in Bottleneck Sampling \cite{bottleneck2025}, therefore does not change the fidelity of \method. On Qwen-Image a single full-resolution step does not help either. Uniform placement of the 3 steps is as faithful as front-loaded placement and more faithful than back-loaded placement.

Scaling the guidance inside the stage to 0.5, 0.75 or 1.25 times the native scale does not improve fidelity at 3 and 4 NFEs on validation prompts with one seed. On Qwen-Image every such scale is on par or worse and on SD3.5-Large worse, so a scale does not correct the deflection of the guidance direction (\cref{app:guidance}).

\subsection{At the switch}
\label{app:atswitch}
Fresh noise at the switch is worse in all 50 setting--budget pairs, by 0.013 to 0.200 \lpips, in the order of how much of the spectrum is still undecided at the switch (\cref{prop:handover}). The stage needs only the coarse noise, because the bands above it are decided after the switch. Replacing the low-resolution noise by the spectral crop of $\epsilon$, which agrees with $\epsilon$ coefficient by coefficient up to the half-resolution Nyquist frequency, leaves \lpips unchanged on FLUX.1-dev and Qwen-Image. The $2\times2$ block mean already carries this coarse noise at a cosine of 0.996 below 4 cycles per image.

Defining the low-resolution stage by $\sh$ rather than by the steps of a grid makes it independent of the grid. A plain split of SharpEuler's grid at $\sh$, with its steps above $\sh$ at half resolution and the rest at full resolution, matches \method at equal cost on FLUX.1-dev, 0.404 against 0.405 \lpips at 4.95 NFEs. On Qwen-Image the split reaches parity only at 1.5--1.75$\times$ the cost, because Qwen-Image's grid is sparse near $\sigma=1$ and the split switches late.

\subsection{Without the first-step shrink}
\label{app:shrink}
Without its first-step shrink, \method is still closer to the reference than SE + shrink, which keeps the shrink, at 2 to 4 NFEs on Qwen-Image and at 2 and 3 NFEs on SD3.5-Large, by 0.026--0.112 \lpips. At 4 NFEs on SD3.5-Large it is then on par in \lpips and ahead in \ds and \ir.

\FloatBarrier
\section{Qualitative results}
\label{app:qual}
\begin{figure*}[t]
\centering
\input{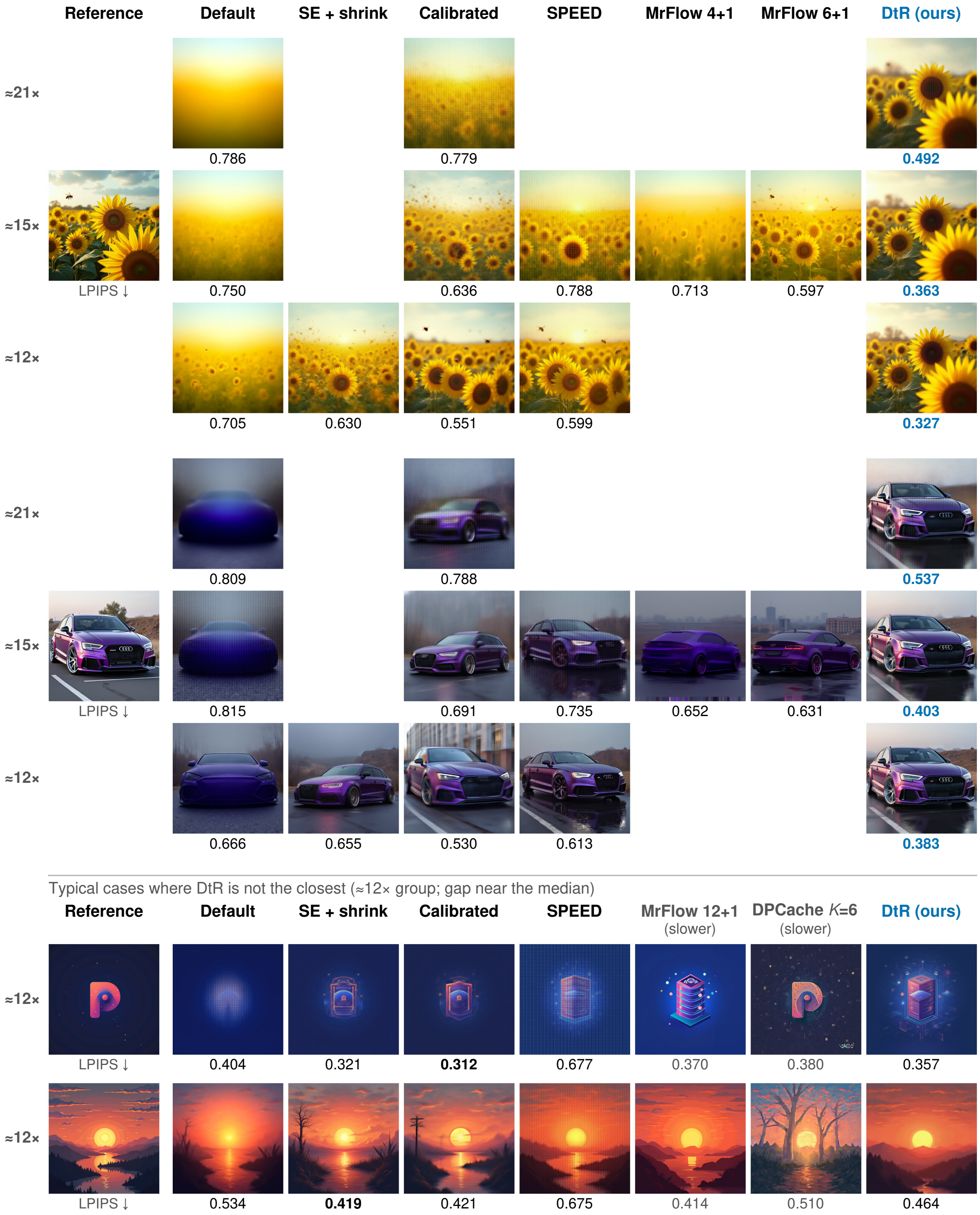}
\caption{\textbf{FLUX.1-dev in the speed groups of \cref{tab:main}, and where \method falls behind.} Test prompts, same noise as the reference. Rows are groups at about $21\times$, $15\times$ and $12\times$ with 2, 3 and 4 NFEs, empty if not in the group. SPEED has 4 steps at $15\times$ and 6 at $12\times$, and CZP is in \cref{fig:teaser}. At the bottom are 2 prompts where another $12\times$ sampler is closer, by about the median of such gaps, 0.044 \lpips. Gray marks the $8\times$ group. \lpips to the reference under each image, best of each group in bold.}
\label{fig:supp_more}
\end{figure*}
\begin{figure*}[t]
\centering
\input{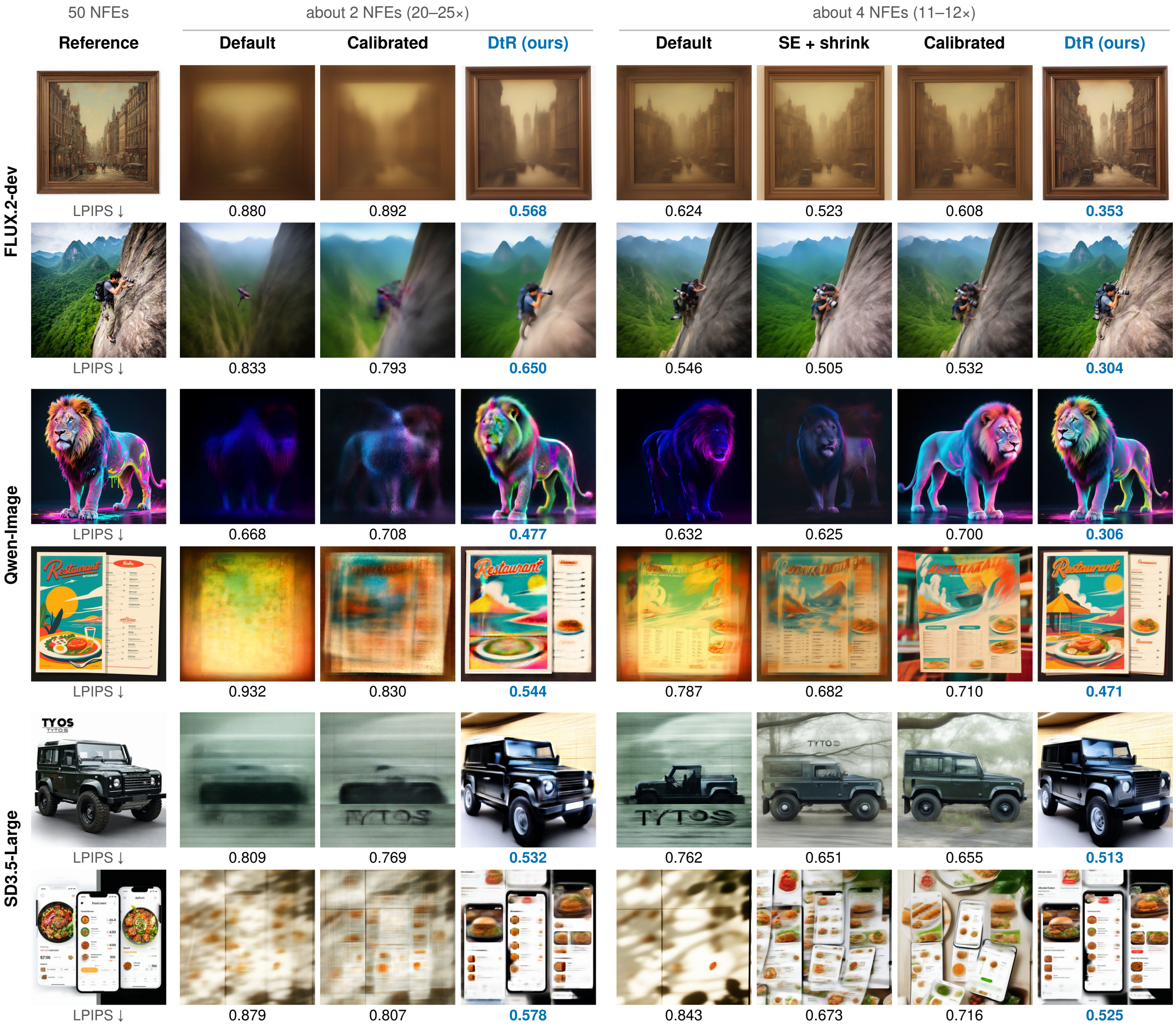}
\caption{\textbf{More FLUX.2-dev, Qwen-Image and SD3.5-Large comparisons.} Test prompts, same noise as the 50-step reference, at about 2 and about 4 NFEs with the samplers of \cref{tab:main}. Same-seed \lpips to the reference under each image, best at each budget in bold.}
\label{fig:supp_models}
\end{figure*}
\noindent In each figure, every sample shares its noise with the reference in the first column, and the number under an image is its \lpips to the reference. \Cref{fig:supp_more} shows FLUX.1-dev with the samplers of the speed groups of \cref{tab:main}, with 2 cases at the bottom where another sampler is closer. \Cref{fig:supp_models} adds 2 cases per model to \cref{fig:models_main}, and \cref{fig:klein} in \cref{app:heldout} shows the held-out model.

\paragraph{Where \method falls behind} \method is closer to the reference than the calibrated schedule on most prompts, 69--92\% on FLUX.1-dev at 2 to 4 NFEs and 57--80\% on Qwen-Image and SD3.5-Large (\cref{app:full}). Where it falls behind, the half-resolution stage has usually decided a small detail differently. On the sunset of \cref{fig:supp_more}, \method keeps the sun, the mountains and the valley, but its half-resolution stage misses the sun's reflection in the river. On the logo, only DPCache, which keeps the 50-step trajectory, draws the letter. Prompts that mention text or logos fall behind twice as often at 4 NFEs. Switching the prompts that mention text or logos to the calibrated schedule would lower the test-set mean \lpips by at most 0.002.

\end{document}